\documentclass[letterpaper]{article} 
\usepackage{aaai2026}  
\usepackage{times}  
\usepackage{helvet}  
\usepackage{courier}  
\usepackage[hyphens]{url}  
\usepackage{graphicx} 
\usepackage{natbib}  
\usepackage{caption} 
\usepackage{algorithm}
\usepackage{algorithmic}
\usepackage{xcolor}

\usepackage{newfloat}
\usepackage{listings}
\DeclareCaptionStyle{ruled}{labelfont=normalfont,labelsep=colon,strut=off} 
\floatstyle{ruled}
\newfloat{listing}{tb}{lst}{}
\floatname{listing}{Listing}
\usepackage{graphicx}

\usepackage{amsmath}
\usepackage{subfigure}
\usepackage{booktabs} 
\usepackage{multirow}
\usepackage{placeins}
\usepackage{afterpage}
\usepackage{caption}

\title{Importance-Aware Data Selection for Efficient LLM Instruction Tuning}
\author{
    Tingyu Jiang\textsuperscript{\rm 1},
    Shen Li\textsuperscript{\rm 1}, 
    Yiyao Song\textsuperscript{\rm 1},
    Lan Zhang\textsuperscript{\rm 2},
    Hualei Zhu\textsuperscript{\rm 1},
    Yuan Zhao\textsuperscript{\rm 1},
    Xiaohang Xu\thanks{Corresponding Author.}\textsuperscript{\rm 3}, 
    Kenjiro Taura\textsuperscript{\rm 3}, 
    Hao Henry Wang\textsuperscript{\rm 1}
}
\affiliations{
    \textsuperscript{\rm 1}Alibaba Cloud Computing\\
    \textsuperscript{\rm 2}Independent Researcher\\
    \textsuperscript{\rm 3}Graduate School of Information Science and Technology, The University of Tokyo\\
    \{jiangtingyu.jty, lishen.ls, songyiyao.syy, zhuhualei.zhl, yise.zhaoyuan, qiao.wh\}@alibaba-inc.com\\
    s23221054004@smail.cczu.edu.cn\\ 
    \{xhxu, tau\}@eidos.ic.i.u-tokyo.ac.jp
}

\usepackage{bibentry}

\begin{document}

\maketitle

\begin{abstract}
Instruction tuning plays a critical role in enhancing the performance and efficiency of Large Language Models (LLMs). Its success depends not only on the quality of the instruction data but also on the inherent capabilities of the LLM itself. Some studies suggest that even a small amount of high-quality data can achieve instruction fine-tuning results that are on par with, or even exceed, those from using a full-scale dataset. However, rather than focusing solely on calculating data quality scores to evaluate instruction data, there is a growing need to select high-quality data that maximally enhances the performance of instruction tuning for a given LLM. In this paper, we propose the Model Instruction Weakness Value (MIWV) as a novel metric to quantify the importance of instruction data in enhancing model's capabilities. The MIWV metric is derived from the discrepancies in the model’s responses when using In-Context Learning (ICL), helping identify the most beneficial data for enhancing instruction tuning performance. Our experimental results demonstrate that selecting only the top 1\% of data based on MIWV can outperform training on the full dataset. Furthermore, this approach extends beyond existing research that focuses on data quality scoring for data selection, offering strong empirical evidence supporting the effectiveness of our proposed method.
\end{abstract}

\section{Introduction}

With the rapid development of natural language processing, LLMs have been able to perform a wide variety of complex language tasks, including writing, translation, conversation, etc~\cite{chiang2023vicuna,schaeffer2024emergent,lin2024wildbench,liu2023emergent,cheng2024towards}. Representative models, such as OpenAI's O1 and GPT-4, have the advanced capabilities to rapidly learn from vast amounts of data and produce responses that closely align with task requirements. However, with the diversification of application scenarios, solely relying on large-scale data pretraining is no longer sufficient to meet all specific needs. One effective and economical solution is instruction tuning~\cite{xu2023wizardlm,yu2024large,shu2023exploitability,tang2024instruction}, which significantly enhances the model's performance on specific tasks by adjusting the model's response to task instructions. This approach allows the LLM to better understand user intent, thereby generating more context-relevant and personalized responses to align with instructions.

Most of the work~\cite{wei2024magicoder,taori2023instruction,wang2023self,mukherjee2023orca} related to instruction tuning has focused on collecting larger, more diverse, and more complex datasets, which costs a great quantity of manpower and material resource, but with low effectiveness~\cite{aghajanyan2021muppet,tang2022mvp}. Moreover, blindly increasing dataset size does not ensure desired results, as it can introduce noise and redundancy, hindering the improvement of instruction tuning capabilities~\cite{zhang2023siren}. Therefore, the key is to adopt an effective data selection strategy when performing instruction tuning, which focus on the quality of data rather than the sheer quantity~\cite{ye2025limoreasoning,li2024quantity}. Recently, the success of ~\cite{liu2024deepseek,xia2024less,xie2023data} have profoundly validated this by selecting samples that enhance the model's reasoning capabilities.

In this paper, we propose a universal data selection strategy, which uses in-context learning (ICL) to find high-quality instruction data from the source dataset that is conducive to enhancing the performance of the model. Given that different pretrained LLMs excel in various task domains and have differing capability scopes, our proposed method is applicable to each model and can select the data that best enhances instruction tuning performance for the given LLM. For each sample in the instruction dataset, we first retrieve the most relevant other sample as its one-shot example. We then assess the LLM's ICL performance on prompts presented with and without one-shot examples. If the LLM's performance with one-shot examples is worse than without, it indicates that the LLM lacks the fundamental abilities to respond to this type of instruction effectively. This type of instruction sample acts as high-quality data that is beneficial to enhancing the model's performance. 

Furthermore, we propose the Model Instruction Weakness Value (MIWV) metric, which evaluates the effect of instruction samples on enhancing the model's capabilities by calculating the loss difference between the LLM responses to the prompt with and without the one-shot example. The higher the MIWV, the more LLM's capabilities are improved by learning the instruction sample. It is noteworthy that we consider the situation where there is no instruction in the source dataset strikingly similar to the given sample, because an irrelevant one-shot example may have negative impact on the response to that instruction. In this case, a high MIWV indicates that the sample is more likely to be chosen, thereby ensuring the diversity of our high-quality data subset. Consequently, a varied instruction-tuning subset can be derived from the source instruction dataset by prioritizing the important samples according to the metrics, which enhances the performance of the LLM.

We conduct instruction tuning experiments on the Alpaca~\cite{taori2023instruction} and WizardLM~\cite{xu2023wizardlm} datasets. Our model outperforms the model trained on the full-scale dataset while utilizing only 1\% of the data. Furthermore, we evaluate our method on widely recognized benchmarks, including the Open LLM Leaderboard~\cite{han2025automatic} and Alpaca Eval~\cite{dubois2024alpacafarm,li2023alpacaeval}. The experimental results demonstrate that our proposed method enhances the performance of the model for instruction tuning. Our contributions are as follows:

\begin{enumerate}
    \item We propose a universal data selection method based on the computation of sample importance, which can be applied to all LLMs. It is simple, efficient, and fully automated, with no need for model training and dependence on external LLMs. 
    \item We propose a novel metric named Model Instruction Weakness Value (MIWV) to quantify the importance of instruction samples in enhancing LLM's capabilities. This metric provides a guide to selecting the most beneficial samples from a vast pool of instruction data for improving model's performance. To the best of our knowledge, our work is the first to design a quantitative evaluation metric by leveraging the inherent contextual learning capabilities within instruction data.
    \item The experimental results show that instruct-tuning the LLM with a small but high-quality dataset from our data selection strategy significantly enhances its capabilities, which demonstrates the effectiveness and superiority of our method. Moreover, our method achieves better outcomes compared to other advanced approaches.

\end{enumerate}

\section{Related Works}
\subsection{Instruction Tuning}
Instruction tuning as a widely utilized training method, allows the model to effectively follow instructions within a specific domain~\cite{DBLP:conf/naacl/LiuWYCSCYY24,longpre2023flan,wang2024survey,qin2024unleashing}, ensuring that its responses align with the expected knowledge within that domain. Recent research has focused on developing technical innovations to improve the performance of LLMs, with a specific focus on their abilities to generalize to unfamiliar instructions. Models such as OpenAI's GPT-4~\cite{achiam2023gpt} and GPT-4o~\cite{hurst2024gpt} are trained through instruction tuning and reinforcement learning to produce responses that are more suitable for specific scenarios and favored by users. Since OpenAI has not yet released the source code of advanced models, most of the current work such as Alpaca~\cite{taori2023instruction} and WizardLM~\cite{li2024quantity,DBLP:conf/emnlp/KungY0CP23,touvron2023llama} adopt LLM from the LLaMA series for instruction tuning. The Qwen2.5 series models~\cite{yang2024qwen2} perform instruction tuning on larger datasets, laying a solid foundation for the application of reasoning capabilities.

\subsection{Instruction Data Selection}
Although instruction tuning is effective, it also has many challenges: (1) high-quality instruction datasets are required, and existing instruction datasets are usually limited in quantity, diversity, and creativity; (2) continuously increasing the number of tasks and the amount of data may enhance the performance of instruction tuning, but it also consumes a lot of time and resources. InstructMining~\cite{cao2023instruction} introduces a set of carefully selected metrics for evaluating the quality of instruction samples and applies a statistical regression model to select the high-quality subset, but does not provide performance comparable to models trained using the full data. INSTAG~\cite{lu2023instag} and Alpagasus ~\cite{DBLP:conf/iclr/ChenLYWGYTS0HJ24} utilize ChatGPT to tag the instruction data to ensure diversity and complexity. QDIT~\cite{bukharin2023data} and Deita~\cite{DBLP:conf/iclr/0131Z00H24} offer the methods to simultaneously control dataset diversity and quality, enabling the study of their effects on instruction tuning. However, these approaches require the use of an additional  model to filter high-quality data. RECOST~\cite{zhang2024recost} selects data by calculating differences in conditional entropy, but it relies on external knowledge data. Additionally, some research filters high-quality data using the fine-tuning model itself. For example, SelectIT~\cite{liu2024selectit} evaluates sample quality via multiple inferences, deriving token probability distributions to compute a quality score. DiverseEvol~\cite{wu2023self} introduces a self-evolving mechanism that enables the model to proactively sample more effective subsets and iteratively enhance the training set to improve performance.

\section{Methodology}

\begin{figure}[!ht]
    \centering
    \includegraphics[width=\columnwidth]{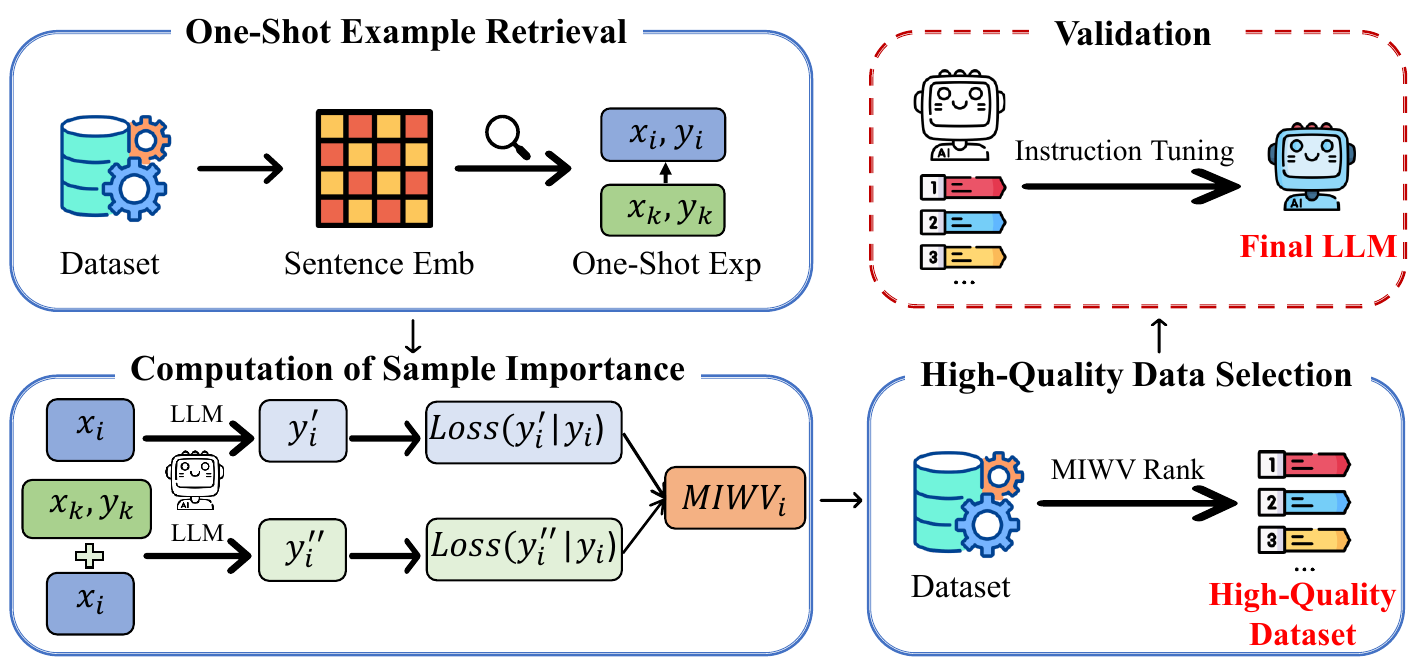}
    \caption{Overview of our proposed method.}
    \label{fig:overview}
\end{figure}

In this section, we propose a universal data selection method based on sample importance, which is calculated by retrieving one-shot example ICL. The framework of our method, as demonstrated in Figure~\ref{fig:overview}, consists of three key steps: one-shot example retrieval, computation of sample importance, and high-quality data selection. The details of each step are introduced as follows.

\subsection{One-Shot Example Retrieval}
\label{sec:One-Shot Example Retrieval}

In this step, we first compute vector embeddings for all samples in the instruction dataset. Then, for each sample, we find the most similar one among the others (excluding itself). Specifically, for an initial instruction dataset $\textit{D}$, we define $x = map\left (Instruction,\left [ Input \right ] \right )$ as the complete instruction input, $\textit{y}$ as the corresponding response, and the map function is aligned with the original target dataset. Thus, $\textit{D}$ consists of
$n$ instruction-response pairs $\{(x_1, y_1), (x_2, y_2), . . . , (x_n, y_n)\}$. With embedding model $E\left ( \cdot \right )$ , each token of sample $x_i$ is represented as:
\begin{equation}
\textbf h_{i}^{1}, \textbf h_{i}^{2}, ..., \textbf h_{i}^{Q} = E(x_{i}^{1}, x_{i}^{2}, ..., x_{i}^{Q}),
\end{equation}
where $Q$ is the number of tokens in the sample. Then, we get the representation $\textbf h_{i}$ as follows: 
\begin{equation}
\textbf h_{i} = \frac{\sum_{q=1}^{Q}\textbf h_{i}^{q}}{Q}.
\end{equation}
For the $x_{i}$ and another sample $x_{j}$ (where $j\neq i$), the cosine similarity $sim(\textbf h_{i},\textbf h_{j})$ between the two can be calculated by their respective corresponding vectors $\textbf h_{i}$ and $\textbf h_{j}$. The specific calculation formula is as follows:
\begin{equation}
sim(\textbf h_{i}, \textbf h_{j}) = \frac{\textbf h_{i} \cdot \textbf h_{j}}{\lVert \textbf h_{i} \rVert \lVert \textbf h_{j} \rVert}.
\label{eq:3}
\end{equation}
Thus, we calculate the similarity between $x_{i}$ and all other samples based on Eq.~\ref{eq:3}, and find the sample $x_{k}$ that is most similar to $x_{i}$ by the maximum similarity value:
\begin{equation}
k = arg\max_{j\neq i} \left ( sim\left (  \textbf h_{i}, \textbf h_{j} \right )  \right ),j \in \{1, 2, ..., n\}.
\end{equation}
Finally, we define $\left ( x_{k}, y_{k}\right )$ as the one-shot example of $\left ( x_{i}, y_{i}\right )$.

\label{sec:Experiments}
\begin{figure*}[!ht]
	\centering	
	\subfigure[The distribution of the number of win and win rates for the Alpaca instruction dataset on different models under the Vicuna test set.]{
		\includegraphics[width=0.45\textwidth]{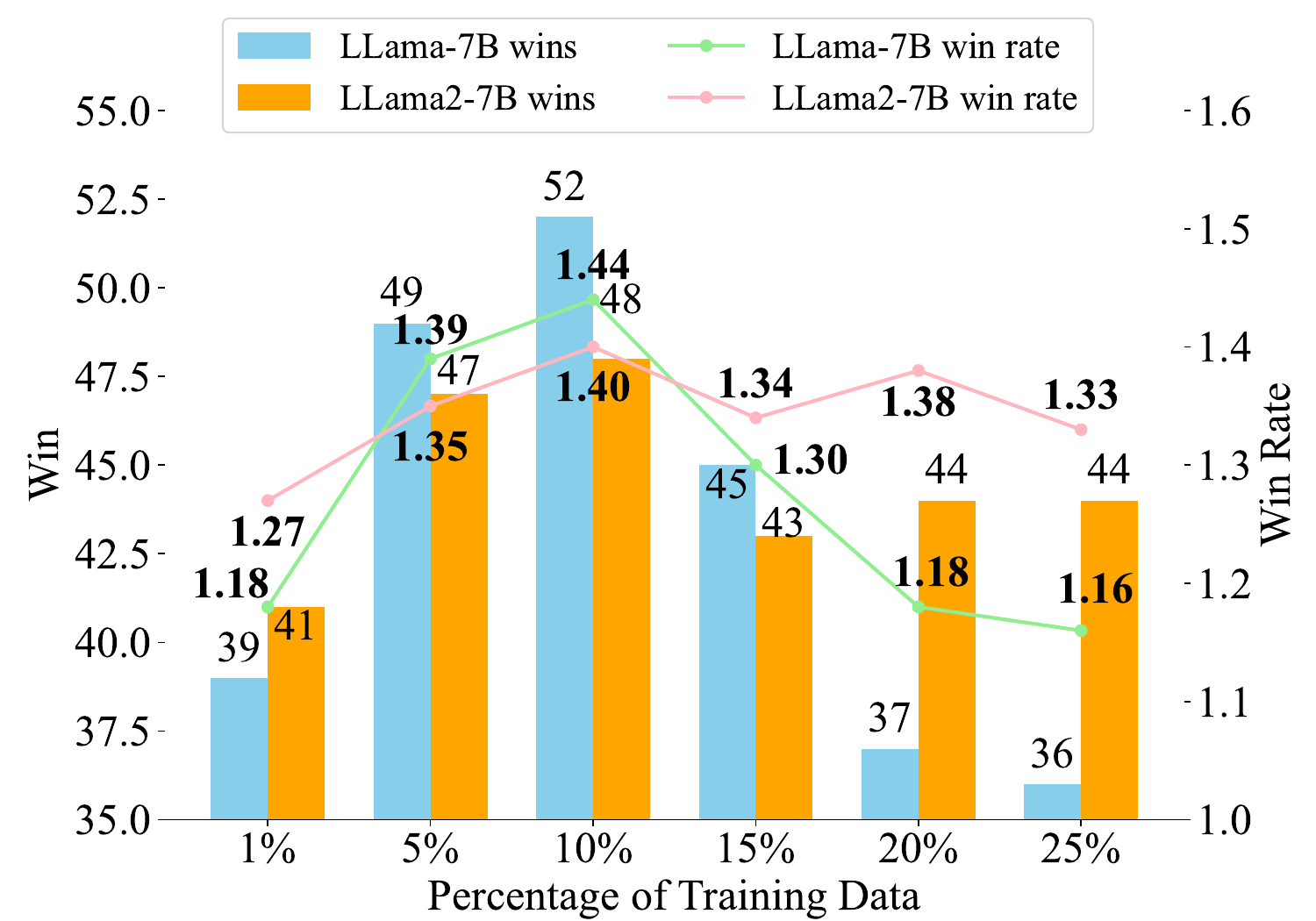}
		\label{fig:2a}
	}\ \ \
	\subfigure[The distribution of the number of win and win rates for the WizardLM instruction dataset on different models under the Sinstruct test set.]{
		\includegraphics[width=0.45\textwidth]{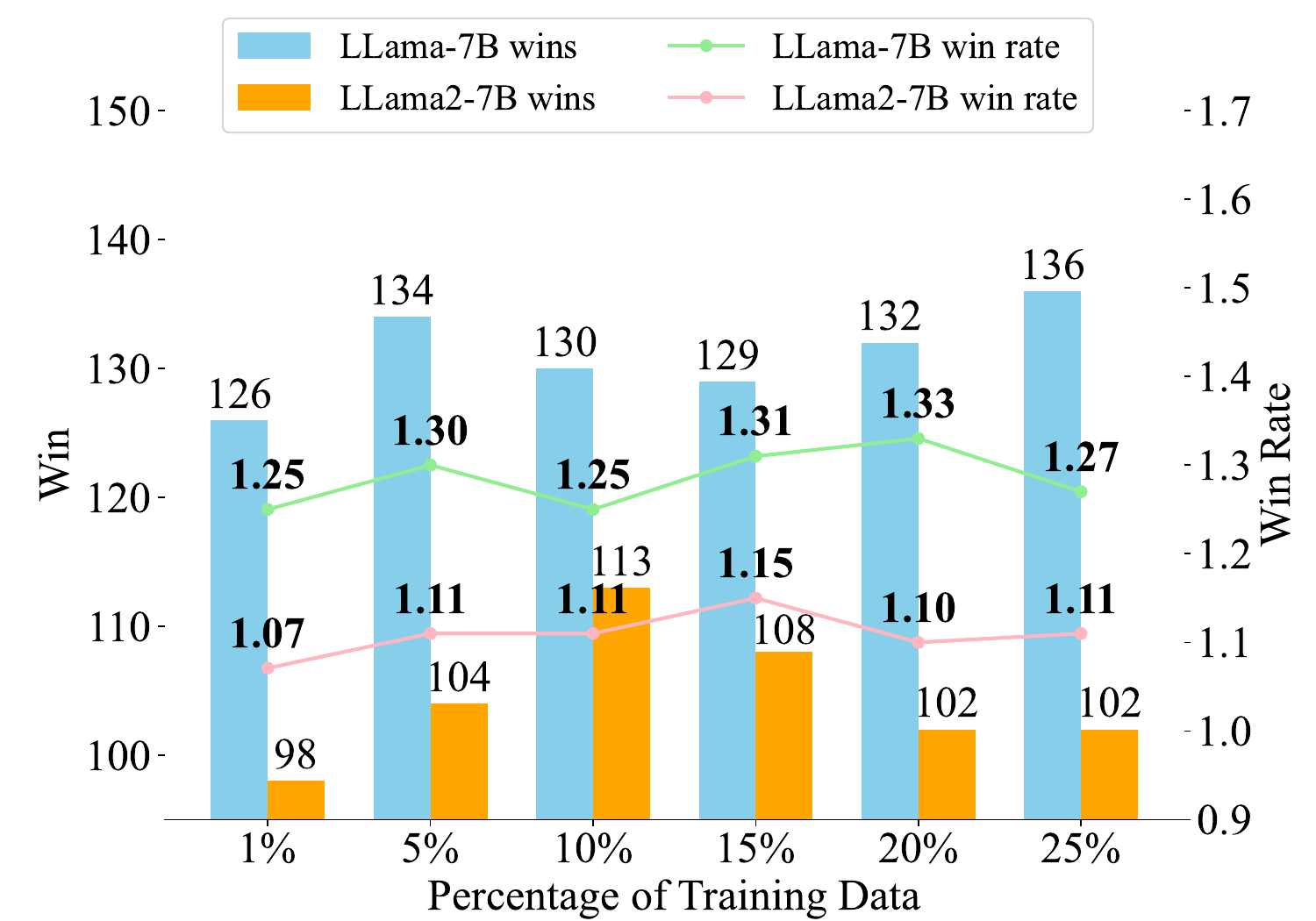}
		\label{fig:2b}
        }\ \ \
	\caption{Comparing our models trained on selected data with full data. Both (a) and (b) use GPT-4 as the judge.}
\label{fig:2}
\end{figure*}

\subsection{Computation of Sample Importance}
\label{sec: Computation of Sample Importance}
It is challenging to assess the quality of an instruction sample without the assistance of human experts, let alone applying it to all LLMs. Therefore, a quantitative metric to measure the importance of a sample based on the LLMs is essential. We boldly introduce the MIWV metric to quantatively evaluate the importance of each instruction sample in enhancing model's performance. The MIWV of each sample is defined as the loss difference between the LLM responses to the prompt with and without one-shot examples. During instruction tuning, the loss of the instruction-response pair is calculated by continuously predicting the next token based on the previous words when the instruction is given.
\begin{equation}
L_\theta\left (  y_{i}|x_{i} \right ) = -\frac{1}{A} \sum_{a=1}^{A} \log{p\left ( y_{i}^{a}|x_{i}, y_{i}^{1},...,y_{i}^{a-1} \right ) },
\end{equation}
where $A$ represents the length of the groundtruth answer $y_{i}$, and $y_{i}^{a}$ denotes its $a$-th token. $L_\theta\left ( y_{i}|x_{i} \right )$ is used to evaluate the degree of challenge encountered by the LLM in generating an answer to a given instruction. However, it does not directly reflect the actual response abilities of LLM to the instruction sample, since the performance of LLM may be affected by the inherent characteristics of the instruction, and its potential capabilities may not be fully stimulated. More accurately, in order to measure the model's capabilities to understand and respond to the instruction, we introduce one-shot examples into the prompt to enrich the input information, which can stimulate LLM's capabilities, as follows:
\begin{equation}
C = \mbox{Prompt}\left (x_{k}, y_{k}\right ),
\end{equation}
\begin{multline}
L_{\theta}(y_i | x_i, C) = -\frac{1}{A} \sum_{a=1}^{A} \log p \Big( y_i^a \,\big|\, x_i, C, \\
y_i^1, \ldots, y_i^{a-1} \Big),
\end{multline}
where $C$ is the one-shot prompt of the instruction sample $\left ( x_{i},  y_{i}\right ))$. We refer to $L_\theta\left (  y_{i}|x_{i}, C  \right )$ as the prompt loss. It measures the capabilities of the model to make the correct response when prompted by the similar sample.
Consequently, in order to reflect the model's weakness in dealing with the instruction sample, MIWV is calculated as
\begin{equation}
\mathrm{MIWV}\left ( x_{i},  y_{i}\right ) = \textit{L}_\theta\left (  y_{i}|x_{i}, C  \right ) - L_\theta\left (  y_{i}|x_{i} \right ).
\end{equation}
A high MIWV value indicates that samples elicit weak responses from the LLM, making them valuable for LLM's enhancement of capabilities.
\subsection{High-Quality Data Selection}
In order to obtain the final LLM that performs better on specific tasks, it's essential to acquire a high-quality dataset for model instruction tuning. The high-quality dataset is made up of important instruction samples with high MIWV value. 

We have used MIWV to evaluate the importance of each sample. In this section, we rank all instruction samples based on the calculated MIWV values and prioritize those with higher MIWV values. Consequently, the samples with the highest MIWV values form a high-quality subset of data. This subset is then used for instruction tuning to validate the model’s abilities improvement.

\section{Experiments}
\subsection{Datasets}
\label{sec:dataset}
\noindent\textbf{Train Datasets.} To verify the effectiveness of our method, we conduct experiments on two datasets: Alpaca~\cite{taori2023instruction}, and WizardLM~\cite{xu2023wizardlm}, which contain 52002 and 63655 instruction samples, respectively. The dataset details are in Appendix.~\ref{sec:appendix A}.

\noindent\textbf{Test Datasets.} We use five distinct test datasets based on existing literature: Vicuna~\cite{chiang2023vicuna}, Koala~\cite{vu2023koala}, WizardLM~\cite{xu2023wizardlm}, Self-instruct~\cite{wang2022self}, and LIMA~\cite{zhou2024lima}. These datasets total 1,030 carefully designed instruction samples.

\subsection{Experimental Setting}
We implement MIWV using PyTorch 2.0.1 on a Linux server equipped with 984GB RAM, Intel Xeon Platinum 8369B CPU @ 2.90GHz, and Nvidia A100 Tensor Core 80GB GPUs. Bge-en-large~\cite{DBLP:conf/sigir/XiaoLZMLN24} serves as the embedding model for one-shot sample retrieval. 
We employ LLaMA-7B and LLaMA2-7B/13B following the training parameters in~\cite{taori2023instruction}, using the Alpaca codebase.
All experiments are repeated three times with arithmetic mean results reported. The detailed training configurations are provided in Appendix.~\ref{sec:appendix A}.

\subsection{Evaluation Metrics}
Considering the disadvantages of manual evaluation, we employ three  metrics~\cite{chang2024survey,DBLP:conf/iclr/ChenLYWGYTS0HJ24,zhou2024lima} to assess the effectiveness of our method, including (1) Pair-wise Comparison, (2) Open LLM Leaderboard, and (3) Alpaca Eval. The details of the evaluation metrics are listed in Appendix.~\ref{sec:appendix B}.

\begin{table*}[!ht]
\centering
\small
\begin{tabular}{ll|c|ccccc|c}
\toprule
\textbf{Dataset/}                           & \multicolumn{1}{c|}{\textbf{MIWV}}        & \textbf{Pairwise} & \multicolumn{5}{c|}{\textbf{Huggingface Open LLM Leaderboard}$\uparrow$} & \textbf{Alpaca} \\
\textbf{Base Model}                         & \multicolumn{1}{c|}{\textbf{Ratio (Size)}} & \textbf{Win Rate$\uparrow$}     & \textbf{Average}  & ARC    & HellaSwag  & MMLU   & TruthfulQA  & \textbf{Eval$\uparrow$}           \\ \midrule
\textbf{Alpaca/}  & 100\%                                     & 1.000             & 55.25             & 54.35  & 78.65      & 47.02  & 40.98       & 27.75               \\
\textbf{LLaMA2-7B}                                            & 1\% (520)                                 & 1.127    &  56.17                 & 57.25  & 78.28      &  47.84      & 41.32       & \textbf{39.50}                   \\& 5\% (2600)                                 & 1.214    &  56.91                 & 58.82  & 79.31      &  48.83      & 40.98       & \textbf{39.87}                    \\
                                            & 10\% (5200)                                & 1.228    &         \textbf{57.36}          & 58.53  & 79.90      & 49.26       & 41.74       & -                   \\
                                            & 15\% (7800)                                & \textbf{1.248}    &        57.08           & 58.93  & 78.73      &     48.40   & 42.27       & -                   \\ \midrule
\textbf{Alpaca/} & 100\%                                     & 1.000             & 58.78             & 57.59  & 81.98      & 54.05  & 41.49       & 35.00               \\
\textbf{LLaMA2-13B}                                            & 1\% (520)                                 & 1.063    &   60.36               & 62.80  & 82.03      &  54.97      & 41.64       & \textbf{41.30}                     \\ & 5\% (2600)                                 & 1.160    &   \textbf{61.48}                & 63.46  & 83.82      &  55.82      & 42.82       & \textbf{48.24}                     \\
                                            & 10\% (5200)                                & 1.200    &    60.74               & 62.69  & 82.73      &      55.06  & 42.47       & -                   \\
                                            & 15\% (7800)                                & \textbf{1.256}    &   61.11                & 62.82  & 82.76      &     55.56   & 43.28       & -                   \\ \midrule
\textbf{WizardLM/}  & 100\%                                     & 1.000             &    55.02               & 58.61  & 73.32      & 41.21       & 46.92       & 59.25                    \\
\textbf{LLaMA2-7B}                                           & 1\% (636)                                 & 1.048    &  55.45                  & 60.14  & 75.54      &  40.48      & 45.37       & \textbf{60.12}                    \\  & 5\% (3182)                                 & 1.096    &  55.58                  & 61.49  & 75.32      &  40.76      & 45.98       & \textbf{61.37}                    \\
                                            & 10\% (6365)                                & 1.114    & 55.98                & 60.47  & 76.47      &  40.25      & 46.72       & -                   \\
                                            & 15\% (9548)                                & \textbf{1.153}    &  \textbf{57.07}                & 60.85  & 76.98      & 41.70       & 48.73       & -                   \\ \midrule
\textbf{WizardLM/} & 100\%                                     & 1.000             &       60.23            & 61.32  & 80.39      &  51.00      & 48.19       & 67.95                    \\
\textbf{LLaMA2-13B}                                             & 1\% (636)                                 & 1.043    &   60.30                & 61.73  & 80.25      & 50.67       & 48.54       &  \textbf{68.32}                   \\ & 5\% (3182)                                 & 1.050    &   60.41                & 61.92  & 80.78      & 50.25       & 48.67       &  \textbf{69.81}                   \\
                                            & 10\% (6365)                                & 1.107    &   60.49                & 61.56  & 80.46      & 50.80       & 49.12       & -                   \\
                                            & 15\% (9548)                                & \textbf{1.118}    &  \textbf{61.00}                 & 62.34  & 80.84      &  51.23      & 49.58       & -                   \\ \bottomrule
\end{tabular}
\caption{Comparison of our method with four data selection ratios (1\%, 5\%, 10\%, 15\%, 100\%) when fine-tuning two LLMs (LLaMA2-7B/13B) on Alpaca and WizardLM datasets. The finetuned models are evaluated by the Pair-wise win rate (compared to the baseline model finetuned on 100\% data), Open LLM Leaderboard,
and AlpacaEval.}
\label{tab:3}
\end{table*}

\subsection{Main Results}

We apply the data selection method on two train datasets and five test datasets, and use the selected subsets to fine-tune the baseline LLMs, including LLaMA-7B and LLaMA2-7B pretrained models. 
As shown in Figure~\ref{fig:2a}, it can be observed that the Alpaca 1\% model (trained on the top 1\% of data selected by MIWV) significantly outperforms the officially trained Alpaca model (trained on the complete dataset) on the Vicuna test set. By comparing the models obtained from different data ratios, the model trained with 10\% data achieves the highest win rate. Figure~\ref{fig:2b} further illustrates that on the Sinstruct test set, the WizardLM 1\% model, which is trained on the top 1\% of data selected by MIWV, also outperforms the WizardLM model trained on the full dataset. In experiments using the LLaMA-7B pretrained model, the WizardLM 20\% model has the best win rate, while in experiments with the LLaMA2-7B pretrained model, the WizardLM 15\% model exhibits the best performance. It is noteworthy that as the proportion of data used for training increases, the overall win rate exhibits a declining trend. It suggests that the data may interfere with each other or contain harmful noise, further validating the effectiveness of our MIWV method for data selection.

Table~\ref{tab:3} presents the results of three automatic evaluation methods. It utilizes the PairWise Win Rate to directly compare our models with the corresponding models trained on the full dataset. These values greater than 1.0 for this metric indicates that our models outperform full-dataset models. Additionally, the performance of our models and the baseline models on the Huggingface Open LLM Leaderboard and the AlpacaEval Leaderboard are also illustrated in the table. It reveals that models trained on 1\%, 5\%, 10\%, and 15\% of the data surpass the models trained on the full dataset in benchmark tests across both the LLaMA2-7B and LLaMA2-13B models. These results validate the effectiveness of our method.

Furthermore, the experimental results of the MIWV method on the classic multi-task natural language processing dataset NIV2~\cite{wang2022super} further validate its effectiveness. For details, we list on Appendix.~\ref{sec:appendix D}.

\subsection{Comparison with Other Methods}

We compare the performance and efficiency of our method with five other widely accepted studies using LLaMA2-7B on the Alpaca dataset, including IFD Score~\cite{li2024quantity}, SelectIT~\cite{liu2024selectit}, Superfiltering~\cite{DBLP:conf/acl/LiZHLZWCZ24}, Alpagasus~\cite{DBLP:conf/iclr/ChenLYWGYTS0HJ24}, Deita~\cite{DBLP:conf/iclr/0131Z00H24}, DiverseEvol~\cite{wu2023self}, Nuggets~\cite{li2023one} and RECOST~\cite{zhang2024recost}. 

As shown in Table~\ref{tab:1}, the win rate of our method on the test set is significantly ahead of the latest works and is second only to Superfiltering in terms of efficiency. For the IFD Score, both the SelectIT and DiverseEvol approaches require model training, resulting in lower efficiency and poorer performance compared to our method. Nuggets and RECOST rely on a specific set of tasks and external knowledge, respectively, which introduce scoring biases. This is the main reason their win rates are lower than those of our method. The performance of Alpagasus and Deita are not only inferior to our method, but their reasoning efficiency is also constrained by ChatGPT API rate limits. As for Superfiltering, although the method is more efficient, inconsistent models used in data selection and instruction fine-tuning can have a negative impact on the win rate.

\begin{figure*}[!ht]
	\centering	
	\subfigure[The performance of various data selection strategies.]{
		\includegraphics[width=0.45\textwidth]{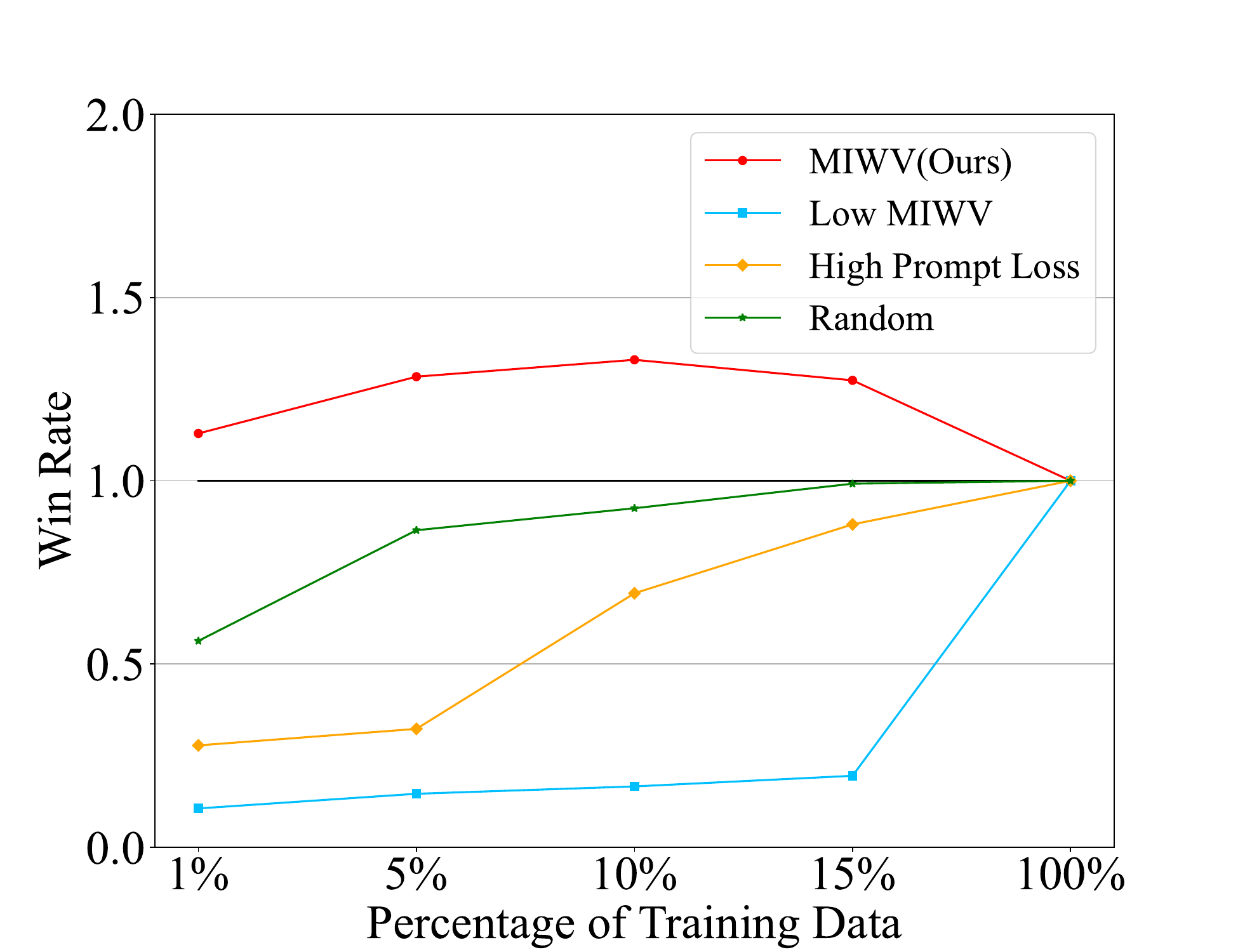}
		\label{fig:444a}
	}\ \ \
	\subfigure[The performance of using various embedding models.]{
		\includegraphics[width=0.45\textwidth]{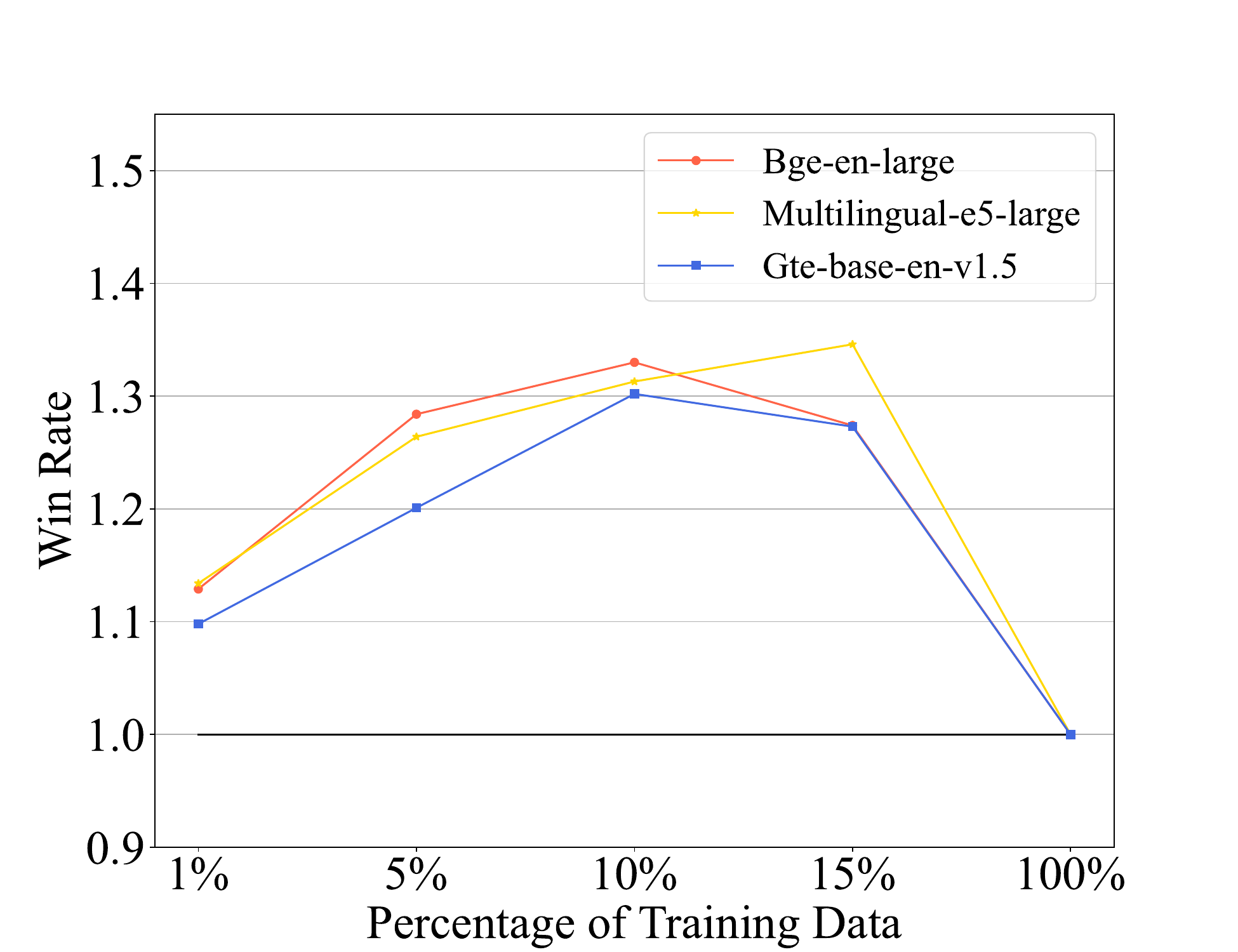}
		\label{fig:444b}
        }\ \ \
	\caption{The comparison of the win rate between models and the official Alpaca model of ablation study.}
\label{fig:444}
\end{figure*}

\subsection{Ablation Study}
We conduct extensive ablation experiments from two aspects to validate the effectiveness of our method: different data selection strategies and embedding models for instruction sample retrieval.

\noindent\textbf{Ablation on Data Selection strategy.}
In addition to our method, we systematically investigate three other data selection strategies: Data Randomly Selected, Data Selected with High Prompt Loss, and Data Selected with Low MIWV. For each data selection strategy, we set four different data ratios: 1\%, 5\%, 10\%, and 15\%. We train the LLaMA-7B models on the Alpaca dataset for each of these ratios and compare their performance against the official Alpaca model. The results are shown in Figure~\ref{fig:444}.

\noindent\textbf{Data Randomly Selected.} We train the models using randomly chosen data. As shown in Figure~\ref{fig:444a}, the models trained with random data always perform worse than the official Alpaca model.

\noindent\textbf{Data Selected with High Prompt Loss.} We select data for instruction tuning based on high Prompt Loss. As shown in Figure~\ref{fig:444a}, the model trained with these data performs worse than the official Alpaca model. This indicates that the high Prompt Loss fails to select truly valuable data.

\noindent\textbf{Data Selected with Low MIWV.} We select data with low MIWV for instruction tuning. As shown in Figure~\ref{fig:444a}, the model trained with low MIWV data has the lowest win rate, which is also worse than the official Alpaca model.

The results above show that the other three data selection strategies all fail to select valuable data to enhance the model's performance, while our method achieves the goal.

\begin{table}[!t]
\centering
\small
\begin{tabular}{l|cccc|c}
\toprule
Metric               & \multicolumn{4}{c|}{Win Rate $\uparrow$}      & Time          \\ \midrule
Selection Scale & \textbf{1\%} & \textbf{5\%}   & \textbf{10\%}  & \textbf{15\%}  & min \\ \midrule
IFD Score    & 0.794        & 0.853          & 0.761          & 0.927       & 161   \\
SelectIT & 0.954   & 1.073    & \underline{1.151}          & 1.183  & 184  \\
Superfiltering   & 0.972    & 1.133    & 1.101          & \underline{1.193}  & \textbf{8}  \\
Alpagasus  & 0.982      & 1.028          & 1.119    & 1.170        & 120  \\
Deita     & 1.009      & 1.092          & 1.032          & 1.013   & 282  \\
DiverseEvol & 1.018   & \underline{1.142}    & 1.137         & 1.165  & 300  \\
Nuggets       & 1.037    & 1.041          & 1.124          & 1.050 & 210 \\ 
RECOST    & \underline{1.092}   & 1.138    & 1.147          & 1.110  & 152  \\
\midrule
\textbf{MIWV (Ours)} & \textbf{1.119}  & \textbf{1.211} & \textbf{1.178} & \textbf{1.234} & \underline{85} \\ \bottomrule
\end{tabular}
\caption{Comparison win rates of other methods. All comparisons are performed by GPT-4 on the WizardLM test set. The time shown represents the time used for data selection.}
\label{tab:1}
\end{table}

\noindent\textbf{Ablation on Embedding Model.}
In order to ensure the reliability and generalization of the experiments, we extract one-shot examples across various embedding models, including Bge-en-large, Multilingual-e5-large~\cite{wang2024multilingual}, and Gte-base-en-v1.5~\cite{zhang2024mgte}. These models, with different scales and multilingual capabilities, increase the experiment's comprehensiveness. We create high-quality instruction subsets with 1\%, 5\%, 10\%, and 15\% ratios from the Alpaca dataset for instruction tuning on LLaMA-7B, as shown in Figure~\ref{fig:444b}. Our findings indicate that our method effectively enhances model's performance regardless of the embedding model. Specifically, models with 5\% and 10\% subsets using Bge-en-large and the 15\% subset with Multilingual-e5-large performed best. Although the subset chosen with Gte-base-en-v1.5 shows a slightly lower win rate, it still significantly outperforms the model trained on the full dataset.

\subsection{Analysis}
\begin{figure}[!t]
    \centering
    \includegraphics[width=0.9\columnwidth]{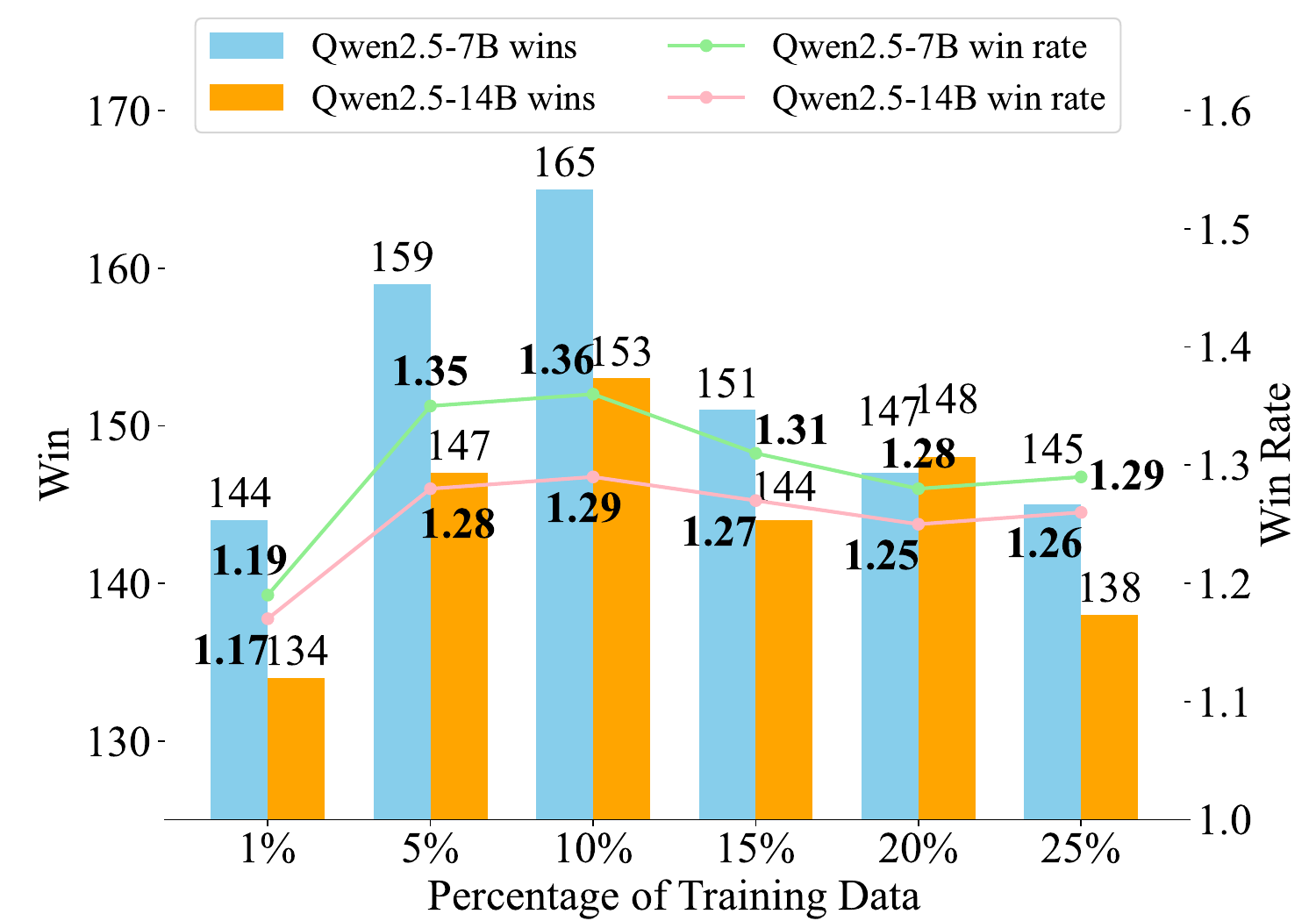}
    \caption{Comparison of Qwen2.5 models trained on selected and full data.}
    \label{fig:qwen}
\end{figure}

\noindent\textbf{Cross-Series Models Effectiveness.}
To further validate the effectiveness and versatility of our method, we apply it to Qwen2.5-7B and Qwen2.5-14B models, which has distinct architectures and capabilities compared to LLaMA/LLaMA2. Figure~\ref{fig:qwen} shows the distribution of win numbers and win rates for the Alpaca instruction dataset under the LIMA test set evaluated by GPT-4. We can find that our method demonstrates exceptional performance on the Qwen2.5 series models. Specifically, models trained with a small amount of high-quality filtered data consistently outperform those trained on the original complete dataset, underscoring the broad applicability of our approach.

\begin{table*}[!ht]
\centering
\small
\begin{tabular}{l|ccc|ccc|ccc|ccc|ccc}
\toprule
Test Datasets              & \multicolumn{3}{|c|}{Vicuna} & \multicolumn{3}{c|}{Koala} & \multicolumn{3}{c|}{WizardLM} & \multicolumn{3}{c|}{Sinstruct} & \multicolumn{3}{c}{LIMA}                                                                                                                \\ \midrule
GPT4 Eval& win$\uparrow$     & tie$\uparrow$    & lose$\downarrow$    & win$\uparrow$   & tie$\uparrow$    & lose$\downarrow$     & win$\uparrow$     & tie$\uparrow$     & lose$\downarrow$      & win$\uparrow$      & tie$\uparrow$     & lose$\downarrow$      & win$\uparrow$    & tie$\uparrow$    & lose$\downarrow$    \\ \midrule
IFD Score& 30      & 12     & 38      & 56     & 58     & 66      & 62     & 49      & 107       & 86      & 67      & 99       &122    & 62     & 116     \\
ICL + IFD Score& 34      & 15     & 31      & 59     & \textbf{61}     & 60      & 74     & 55      & 89       & 92       & 73      & 87       & 129    & 68     & 103     \\ 
MIWV (Ours)& \textbf{39}     & \textbf{16}     & \textbf{25}      & \textbf{76}     & 52     & \textbf{52}      & \textbf{97}     & \textbf{62}      & \textbf{59}       & \textbf{100}       & \textbf{81}      & \textbf{71}       & \textbf{136}    & \textbf{74}     & \textbf{90}     \\ \bottomrule
\end{tabular}
\caption{Comparison of the performance of "IFD Score" with introducing In-Context Learning strategy and the original "IFD Score" on five test sets, using GPT-4 as the judge.}
\label{tab:2}
\end{table*}

\noindent\textbf{ICL-Guided Effectiveness.}
To further demonstrate the role of ICL in the data selection process, we design a comparative experiment based on the "IFD Score". Specifically, we replace the clustering step in the original "IFD Score" method with an ICL approach that extracts embeddings from instruction samples, retrieves one-shot examples for each sample, and computes IFD scores via model inference. We select the 1\% subset of the Alpaca instruction dataset and conduct experiments on the LLaMA-7B. The comparison results of the two methods are demonstrated in Table~\ref{tab:2}. The experimental results show that the model trained with data selected by ICL performs better than the original "IFD Score", with an overall win rate of 1.017 significantly higher than 0.939. Furthermore, the proposed MIWV approach achieves the highest win rate of 1.140, significantly surpassing both methods. These results emphasize the effectiveness and superiority of context learning-assisted data selection and the MIWV metric.

\subsection{Case Study}
\begin{figure}[!ht]
    \centering
    \includegraphics[width=\columnwidth]{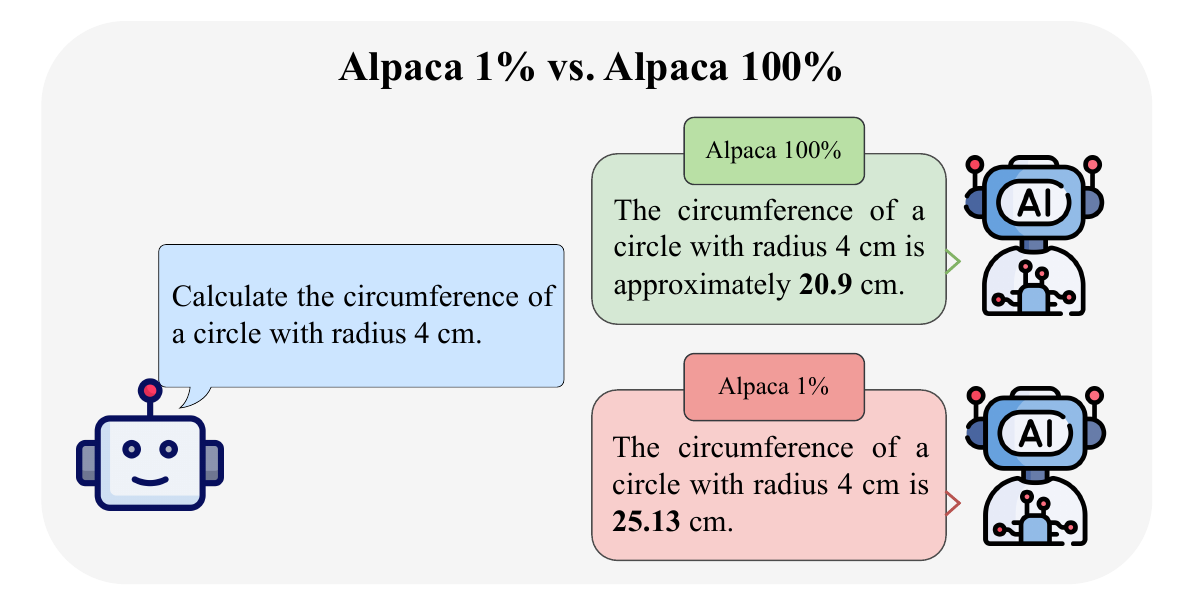}
    \caption{Case Study on Mathematical Capabilities: LLaMA2-7B Models of Alpaca 1\% and Alpaca 100\%.}
    \label{fig:case_study}
\end{figure}
Figure~\ref{fig:case_study} presents a case study focused on mathematical abilities, utilizing the LLaMA2-7B model trained with 1\% of the Alpaca dataset (Alpaca-1\%) and with the full Alpaca dataset (Alpaca-100\%). The results demonstrate that the Alpaca-1\% model successfully provides a correct answer, whereas the Alpaca-100\% model fails to do so. 

\subsection{Data Characteristics}
\begin{figure}[!ht]
    \centering
    \includegraphics[width=0.9\columnwidth]{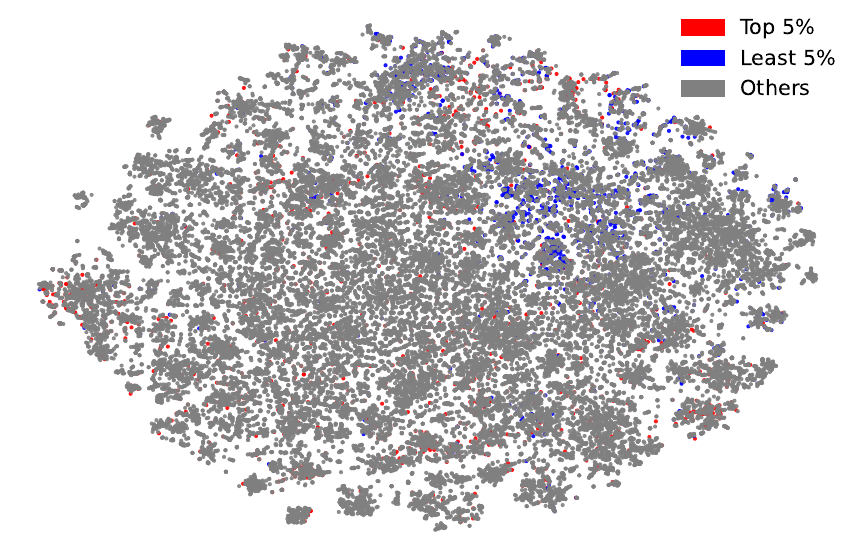}
    \caption{Visualization using t-SNE on instruction embeddings from the Alpaca dataset.}
    \label{fig:tsne}
\end{figure}

\noindent\textbf{Distribution Characteristics.} We visualize the 2D distribution of high-dimensional embeddings from the Alpaca dataset using t-SNE. In Figure~\ref{fig:tsne}, red, blue, and gray points denote the top 5\%, least 5\%, and remaining other samples ranked by MIWV, respectively. The visualization reveals that higher MIWV samples exhibit uniform distribution across the instruction spectrum while lower MIWV samples cluster in specific regions. This pattern indicates that selected data should evenly cover the instruction set distribution to maximize diversity. The clustering of samples in the embedding space indicates that MIWV remains a meaningful discriminative indicator even in the reduced-dimensional representation. Notably, Lower MIWV samples typically involve basic tasks like editing punctuation, words, or simple sentences. In contrast, higher MIWV samples include both basic and more complex tasks, such as storytelling and explaining phenomena.  This diversity is critical for LLM performance, enabling deeper understanding, adaptability, and handling of complex language structures.

\begin{figure}[!t]
    \centering
    \includegraphics[width=0.9\columnwidth]{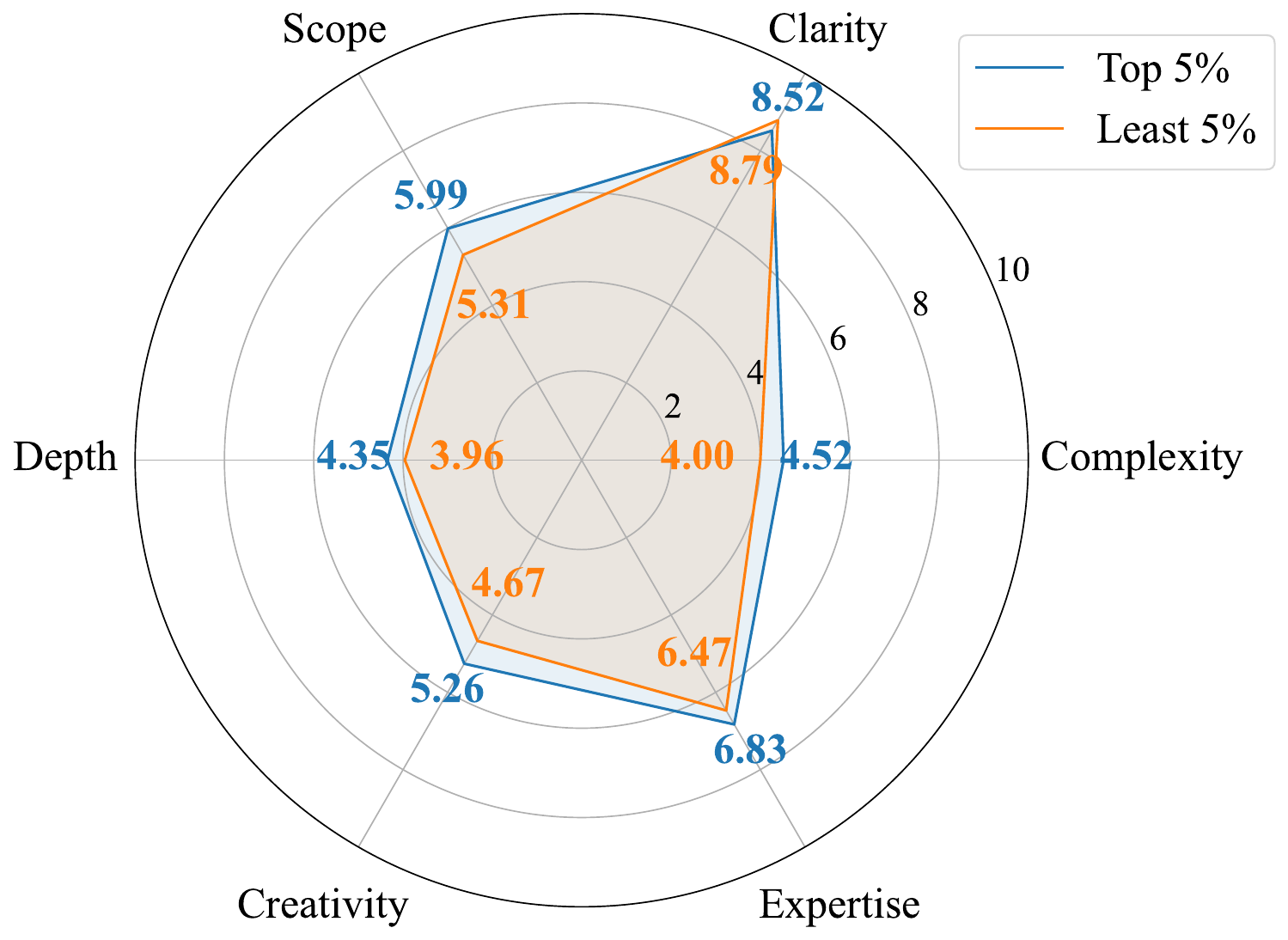}
    \caption{Quality scores for the top and least 5\% of data instances in the Alpaca dataset evaluated by GPT-4.}
    \label{fig:5}
\end{figure}
\noindent\textbf{Quality Characteristics.} To validate MIWV's alignment with data quality, we use GPT-4 to evaluate 100 randomly selected top and the least 5\% samples across six dimensions. As shown in Figure~\ref{fig:5}, higher MIWV samples outperform in Complexity, Scope, Depth, Creativity, and Expertise while maintaining comparable Clarity. This multi-dimensional analysis confirms MIWV's effectiveness in balancing instructional diversity and cognitive challenges, critical for robust LLM training. Integrating spatial distribution and quality metrics highlight MIWV as a robust criterion. Appendix.~\ref{sec:appendix E} shows the details of six dimensions.

\section{Conclusion}

In this paper, we propose a universal high-quality data selection method based on a novel metric called Model Instruction Weakness Value (MIWV), which quantifies instruction sample importance through response discrepancies in In-Context Learning (ICL). Extensive experimental results consistently demonstrate that the LLM instruction-tuned on a small amount of high-quality subset achieves results comparable to or even surpassing those tuned on the full-scale dataset, which proves the effectiveness and superiority of our method. This finding also highlights the potential of our method for the best use of limited resources, offering a new approach for economical and efficient data utilization in model training. Furthermore, the application of ICL for data selection is a relatively novel concept that is expected to inspire future research in data selection methodologies.

\section*{Acknowledgments}
This work was supported by JST CREST Grant Number JPMJCR21M2, including the AIP Challenge Program.

\bibliography{aaai2026}
\clearpage

\appendix

\section{Implementation Details}
\label{sec:appendix A}
The Alpaca dataset is constructed by applying the self-instruction technology~\cite{wang2023self} to the text-DaVinci-003, which contains 52,002 instruction samples and has been widely used in multiple studies. The WizardLM dataset is generated using the Evol-Instruct algorithm, which aims to cover instruction tasks of different complexity levels, with a total of 63,655 instruction samples.

We adopt existing methodologies to filter samples subject to "AI censure" in WizardLM, resulting in a WizardLM dataset comprising 63,655 instances. Experiments are conducted using the Alpaca codebase. For the LLaMA-7B pretrained model, the Adam optimizer is configured with a learning rate of $2 \times 10^{-5}$, a batch size of 128, and training is performed over three epochs. The maximum input length is set to 512 for training on the Alpaca dataset and 1,024 for training on the WizardLM dataset. For the LLaMA2-7B and LLaMA2-13B models, the maximum input length is configured to 2,048 for both the Alpaca and WizardLM datasets. Throughout the model training and evaluation phases, all parameter settings are kept consistent with those of the original work~\cite{li2023alpacaeval,zheng2023judging}.

\clearpage

\section{Evaluation Details} 
\label{sec:appendix B}
\subsection{Pair-wise Comparison}
Referring to work~\cite{li2024quantity}, in the experiment of instruction tuning of Alpaca and WizardLM datasets, we use GPT-4 to score each generated response of the model on the test dataset. The scoring scores range from 1 to 10, reflecting properties such as relevance and accuracy. In addition, in order to eliminate the influence of position bias on GPT4 scoring, the responses are evaluated twice by exchanging the order of responses. We compare the scores of the model in these two evaluations respectively. The specific rules are as follows:

\noindent\textbf{Wins:} outperforms in both or wins one score comparison and ties in the other.

\noindent\textbf{Ties:} ties in both or wins one score comparison and loses in the other.

\noindent\textbf{Loses:} lags in both or ties one score comparison and loses in the other

To compare model performance more clearly, we quantify the overall win rate of the model, which is calculated as $(Num(Win) - Num(Lose))/Num(All) +1$.
\subsection{Open LLM Leaderboard}
The Huggingface Open LLM Leaderboard is a dedicated platform for evaluating the performance of generative LLMs. This leaderboard utilizes the Eval Harness evaluation methodology proposed by~\cite{han2025automatic}, a comprehensive framework designed to thoroughly assess language models across a wide range of tasks. It primarily focuses on four key benchmark tests: ARC~\cite{clark2018think}, HellaSwag~\cite{zellers2019hellaswag}, MMLU~\cite{hendrycks2020measuring}, and TruthfulQA~\cite{lin2021truthfulqa}. These benchmark tests comprehensively cover essential aspects of the models, including reasoning abilities, commonsense understanding, and factual accuracy, ensuring a holistic evaluation of the models' performance across diverse and challenging tasks.
\subsection{Alpaca Eval}
The AlpacaEval Leaderboard is an automated evaluation platform for large language models based on the AlpacaFarm evaluation dataset~\cite{dubois2024alpacafarm,li2023alpacaeval}. As an efficient and reliable tool, AlpacaEval benchmarks the performance of various language models in adhering to general user instructions by comparing their outputs to those of Davinci003. The results demonstrate that its evaluations are highly consistent with annotations provided by human experts. Due to budget constraints, evaluations are conducted on only 5\% of the dataset.

\clearpage

\section{Detailed Main Results}
\subsection{Results on the Alpaca Dataset}
\begin{figure*}[!ht]
	\centering	
  	\subfigure[The distribution of the number of win and win rates for the Alpaca instruction dataset on different models under the Koala test set.]{
		\includegraphics[width=0.47\textwidth]{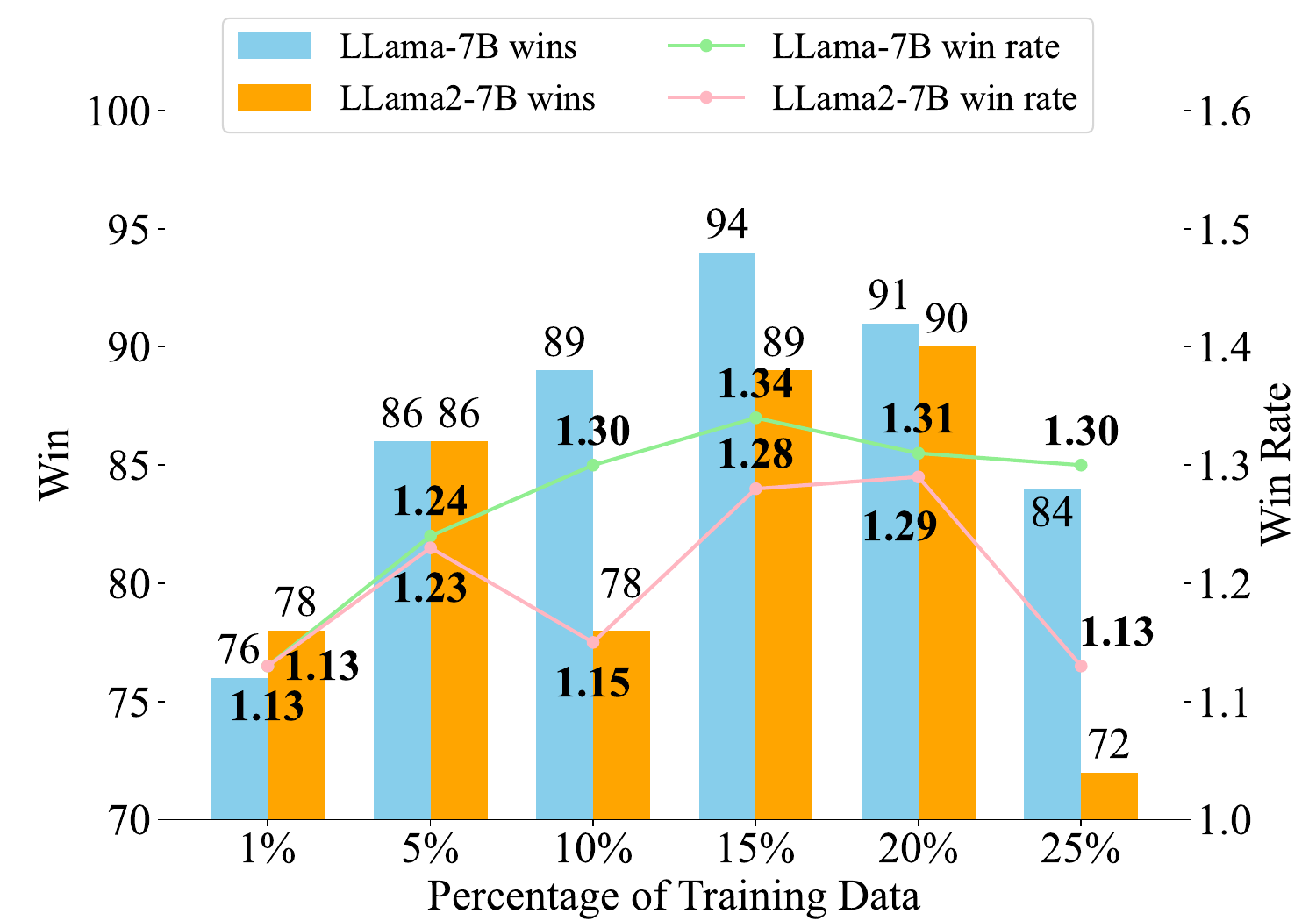}
		\label{fig:5a}
	}\ \ \
        \subfigure[The distribution of the number of win and win rates for the Alpaca instruction dataset on different models under the WizardLM test set.]{
		\includegraphics[width=0.47\textwidth]{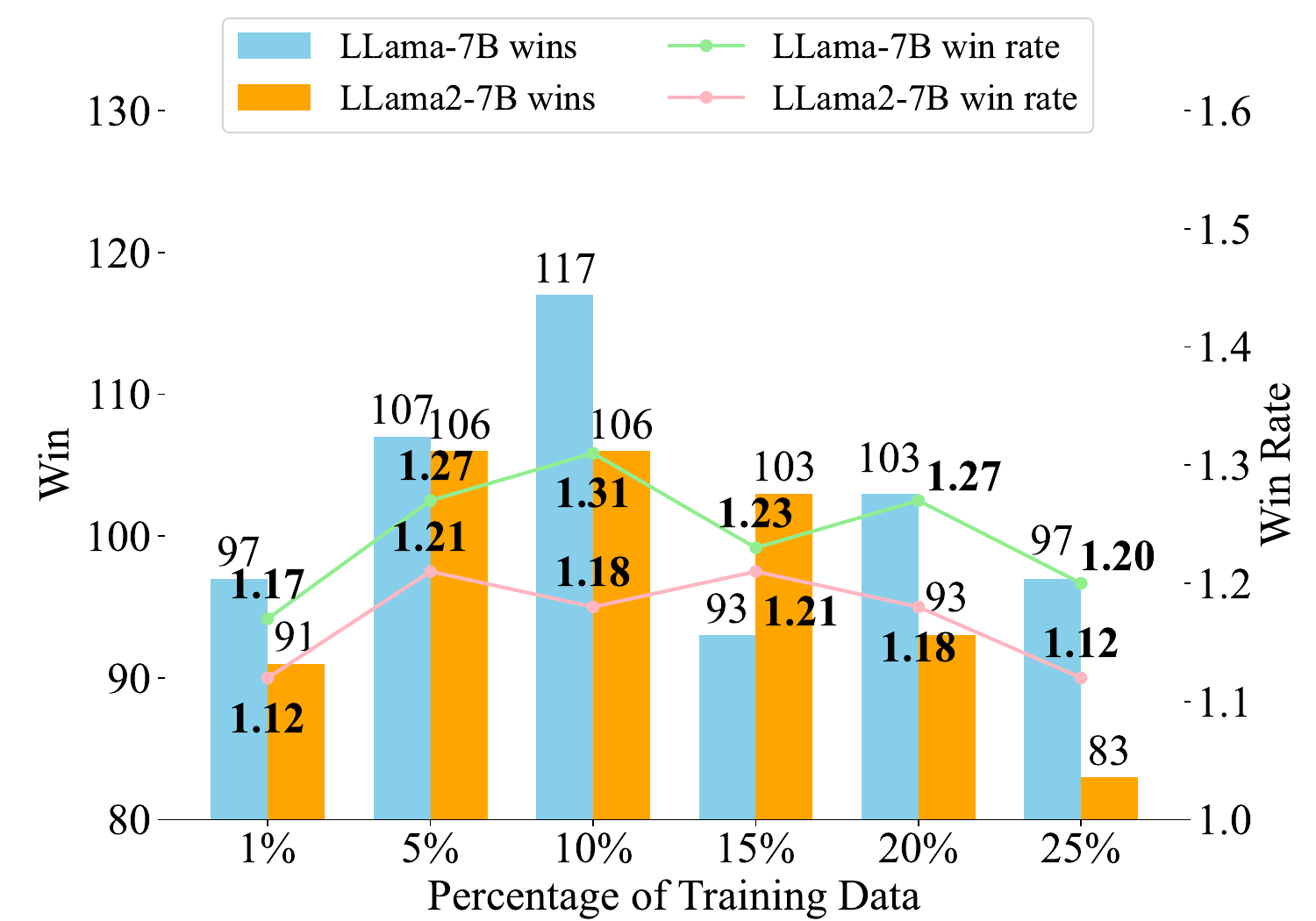}
		\label{fig:5b}
	}\ \ \
        \subfigure[The distribution of the number of win and win rates for the Alpaca instruction dataset on different models under the Sinstruct test set.]{
		\includegraphics[width=0.47\textwidth]{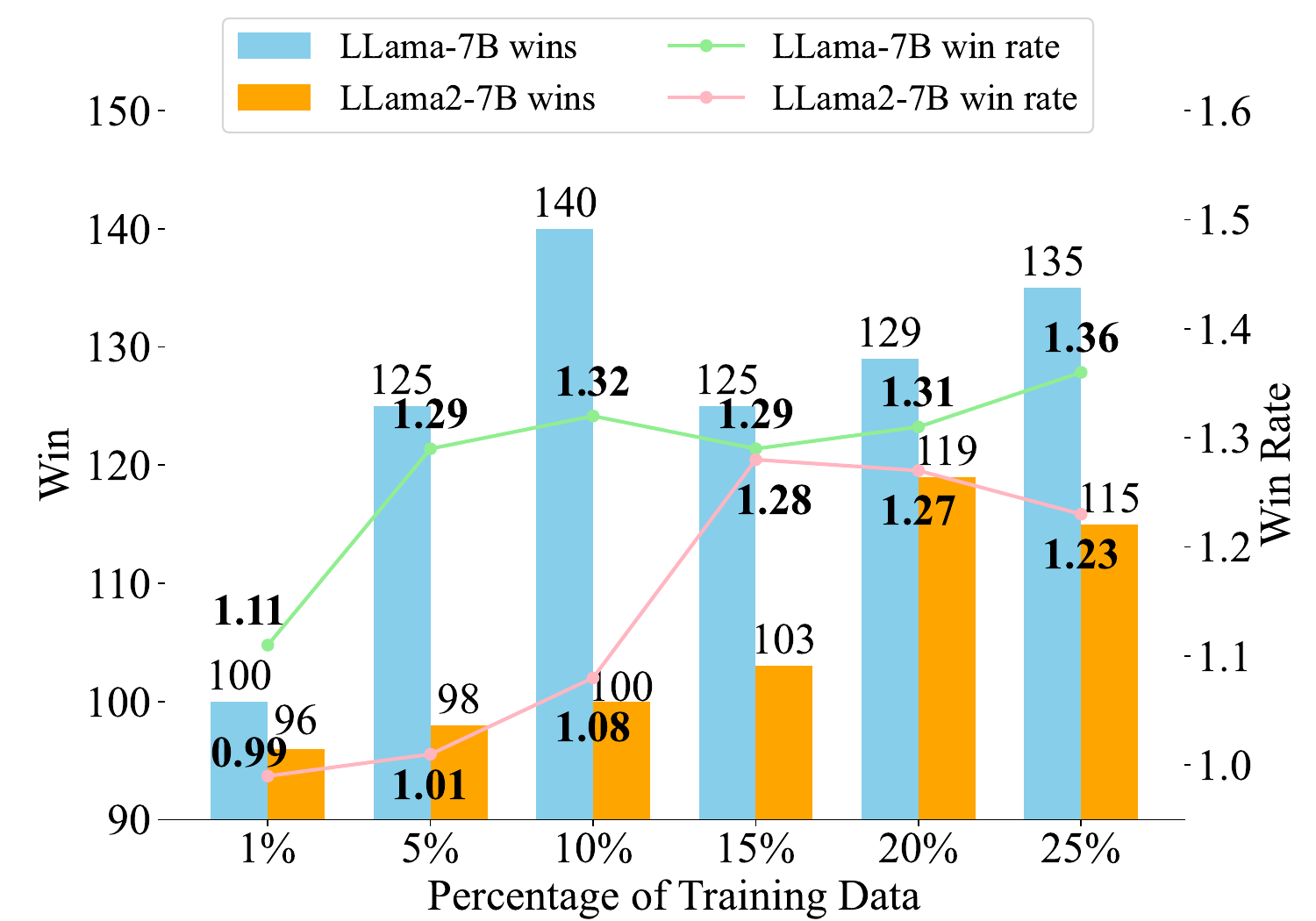}
		\label{fig:5c}
	}\ \ \
        \subfigure[The distribution of the number of win and win rates for the Alpaca instruction dataset on different models under the Lima test set.]{
		\includegraphics[width=0.47\textwidth]{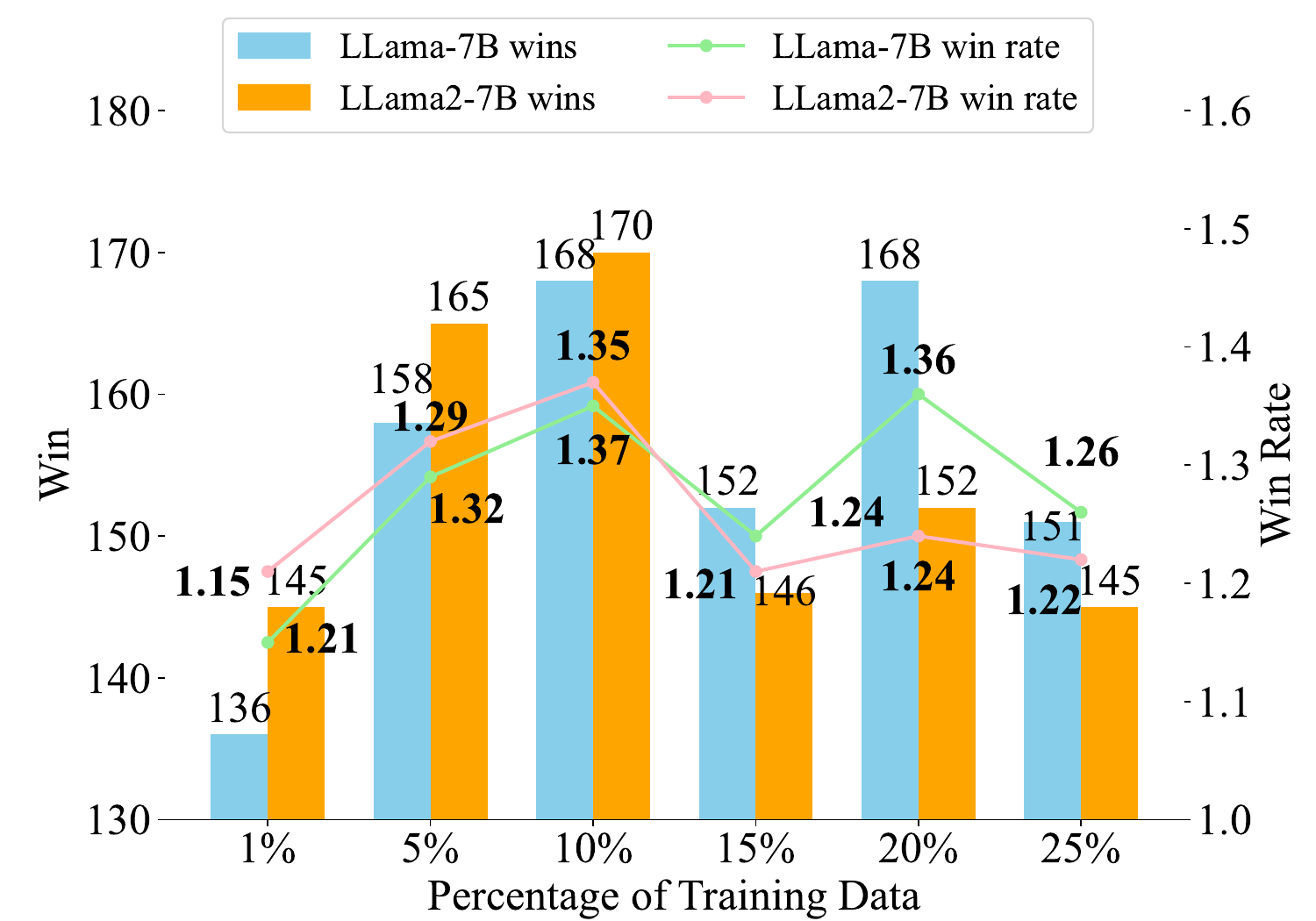}
		\label{fig:5d}
	}\ \ \
	\caption{Comparing our models trained on selected data with full data on the Alpaca dataset. (a), (b), (c), (d) All use GPT4 as the judge.}
\label{fig:5_1}
\end{figure*}

\begin{figure*}[!ht]
	\centering	
        \subfigure[Comparing between our model trained on selected 1\%
data and the official Alpaca model on LLaMA-7b.]{
		\includegraphics[width=0.47\textwidth]{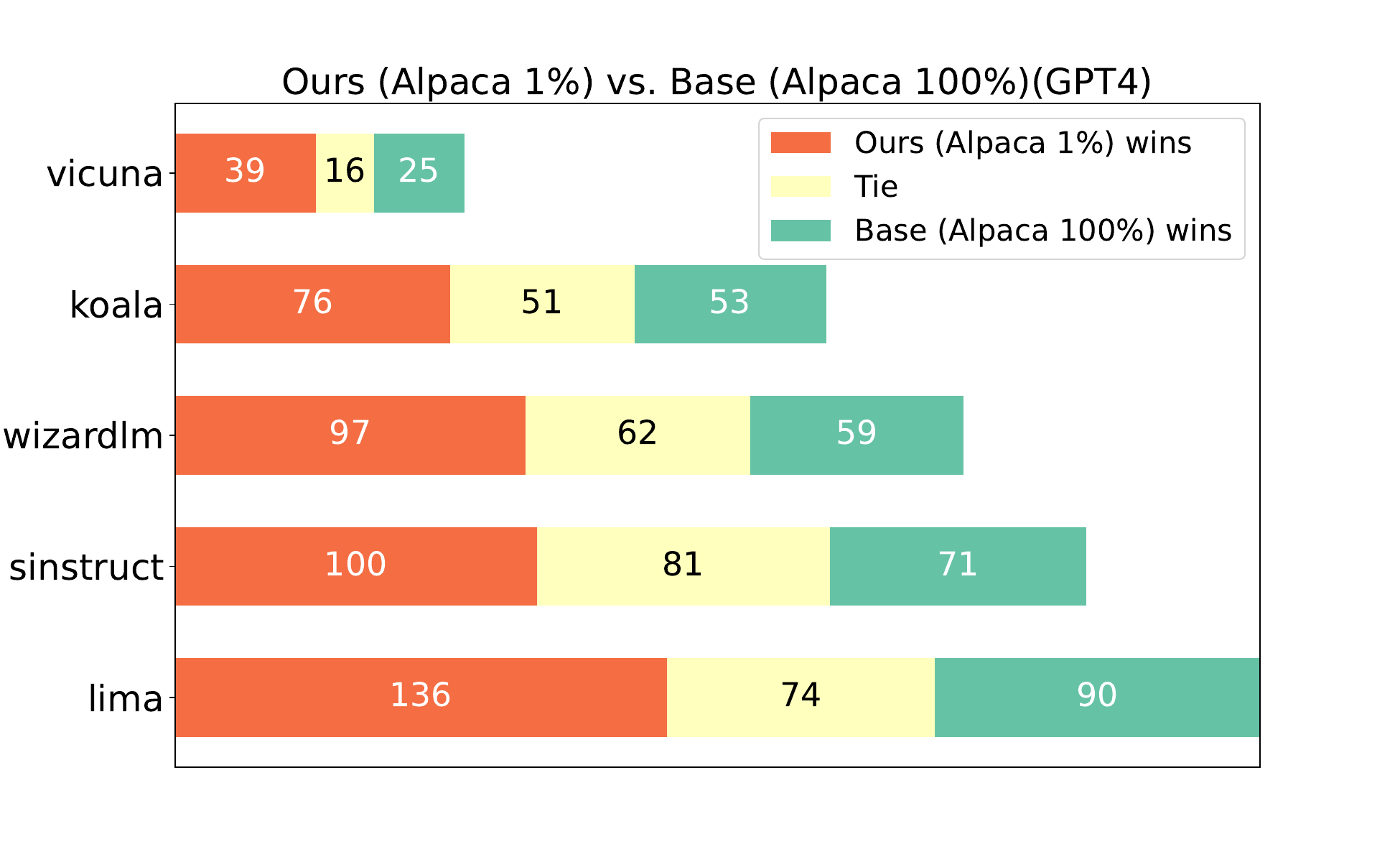}
		\label{fig:before-6a}
	}\ \ \
  	\subfigure[Comparing between our model trained on selected 5\%
data and the official Alpaca model on LLaMA-7b.]{
		\includegraphics[width=0.47\textwidth]{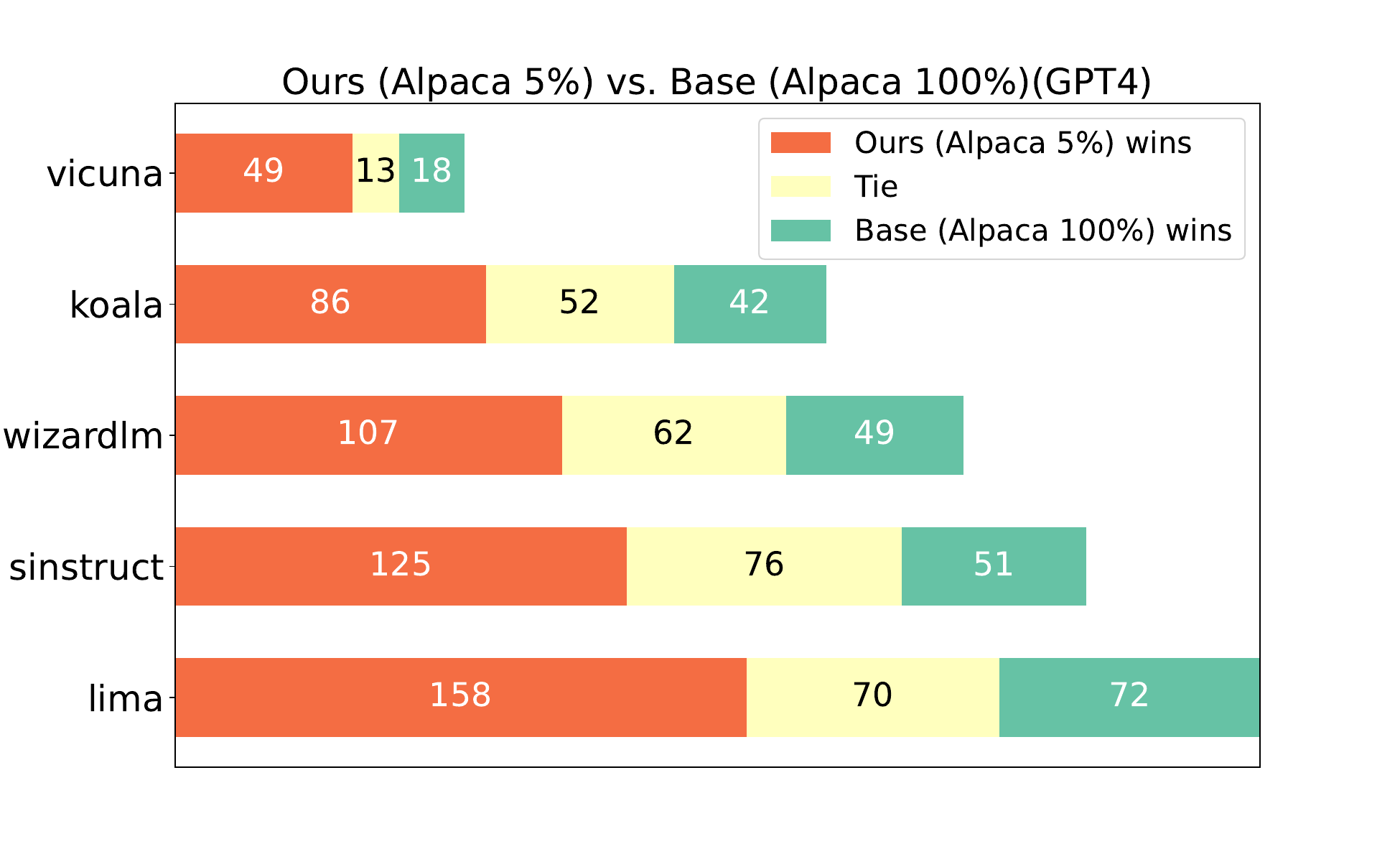}
		\label{fig:6a}
	}\ \ \
        \subfigure[Comparing between our model trained on selected 10\%
data and the official Alpaca model on LLaMA-7b.]{
		\includegraphics[width=0.47\textwidth]{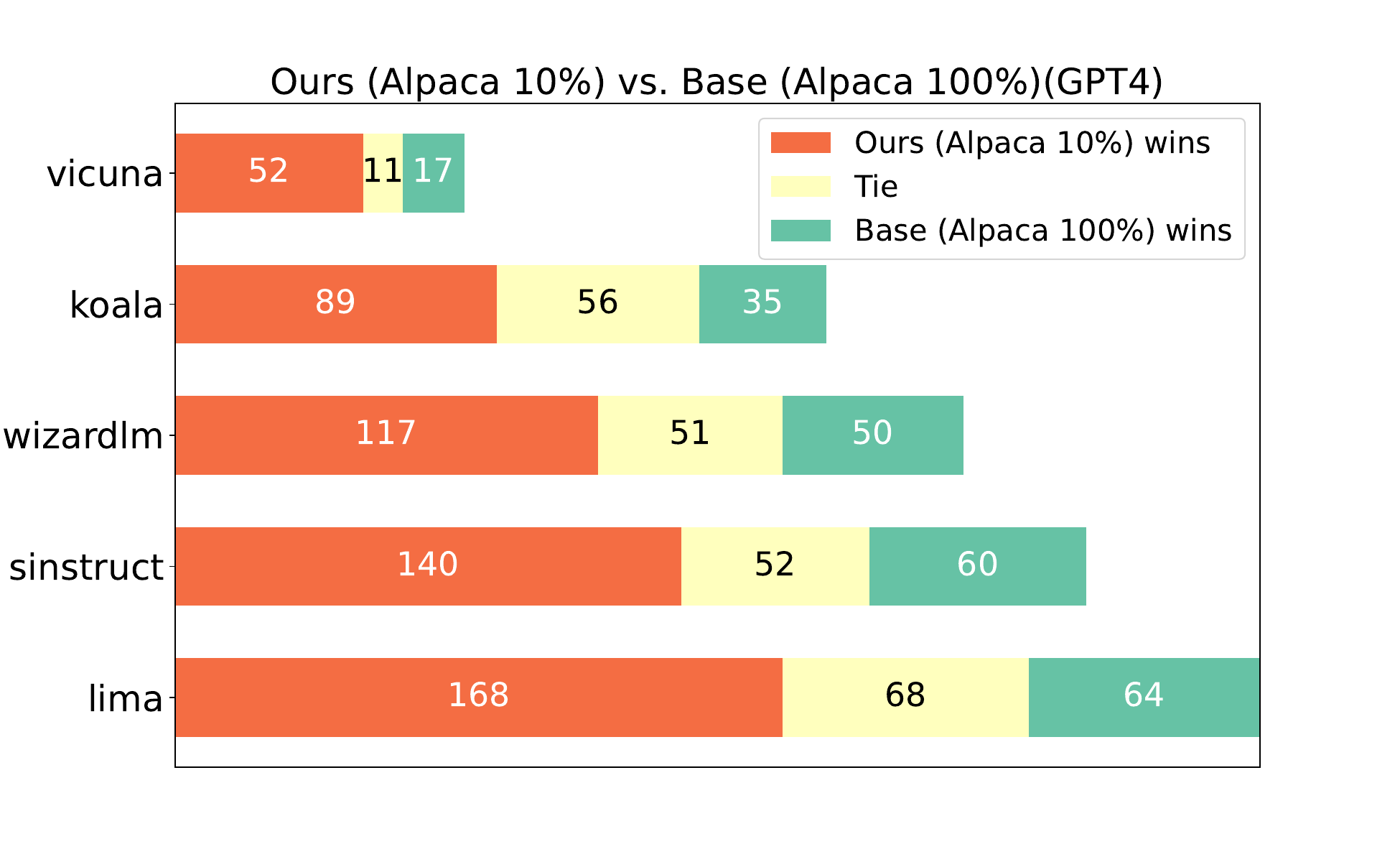}
		\label{fig:6b}
	}\ \ \
        \subfigure[Comparing between our model trained on selected 15\%
data and the official Alpaca model on LLaMA-7b.]{
		\includegraphics[width=0.47\textwidth]{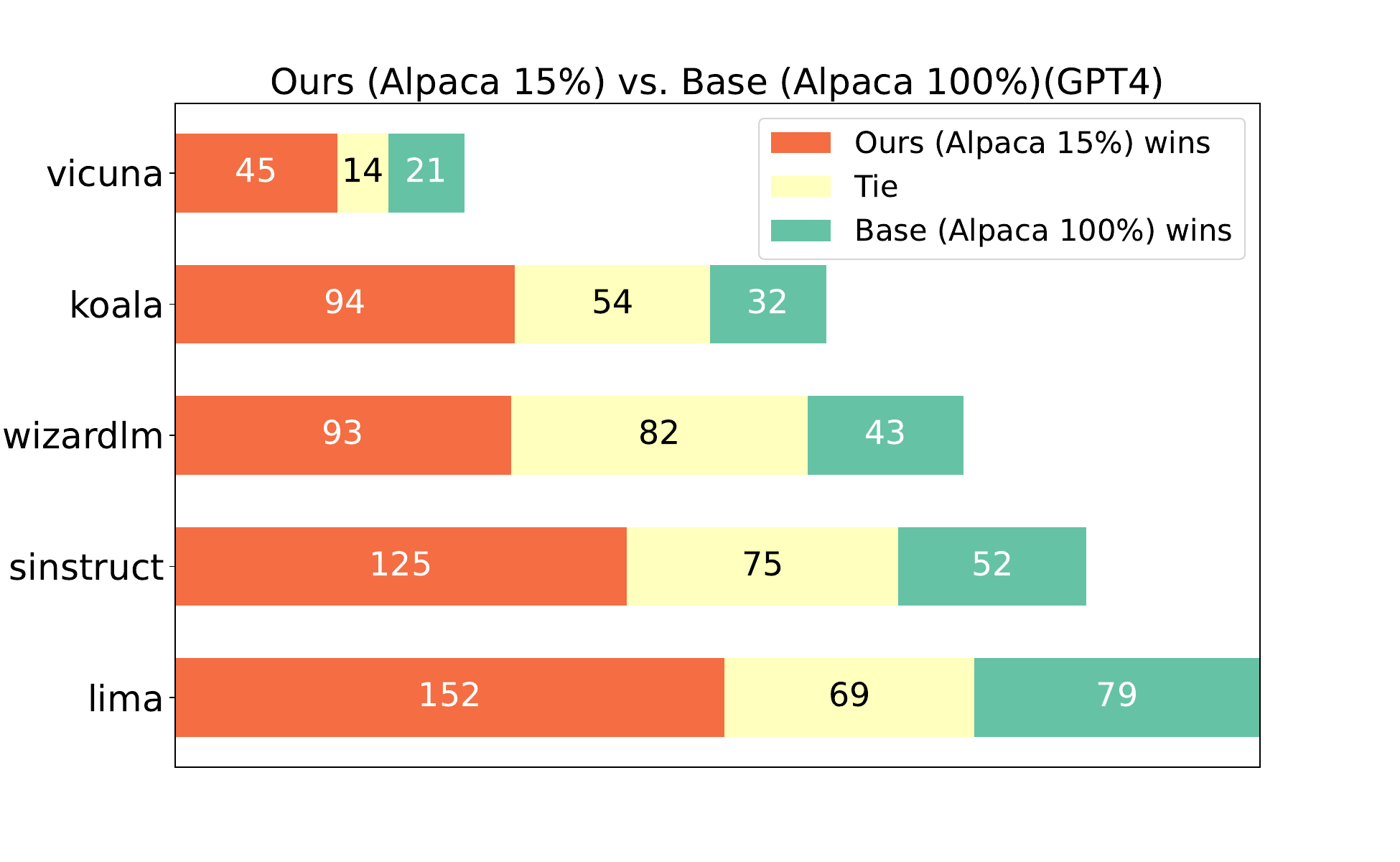}
		\label{fig:6c}
	}\ \ \
        \subfigure[Comparing between our model trained on selected 20\%
data and the official Alpaca model on LLaMA-7b.]{
		\includegraphics[width=0.47\textwidth]{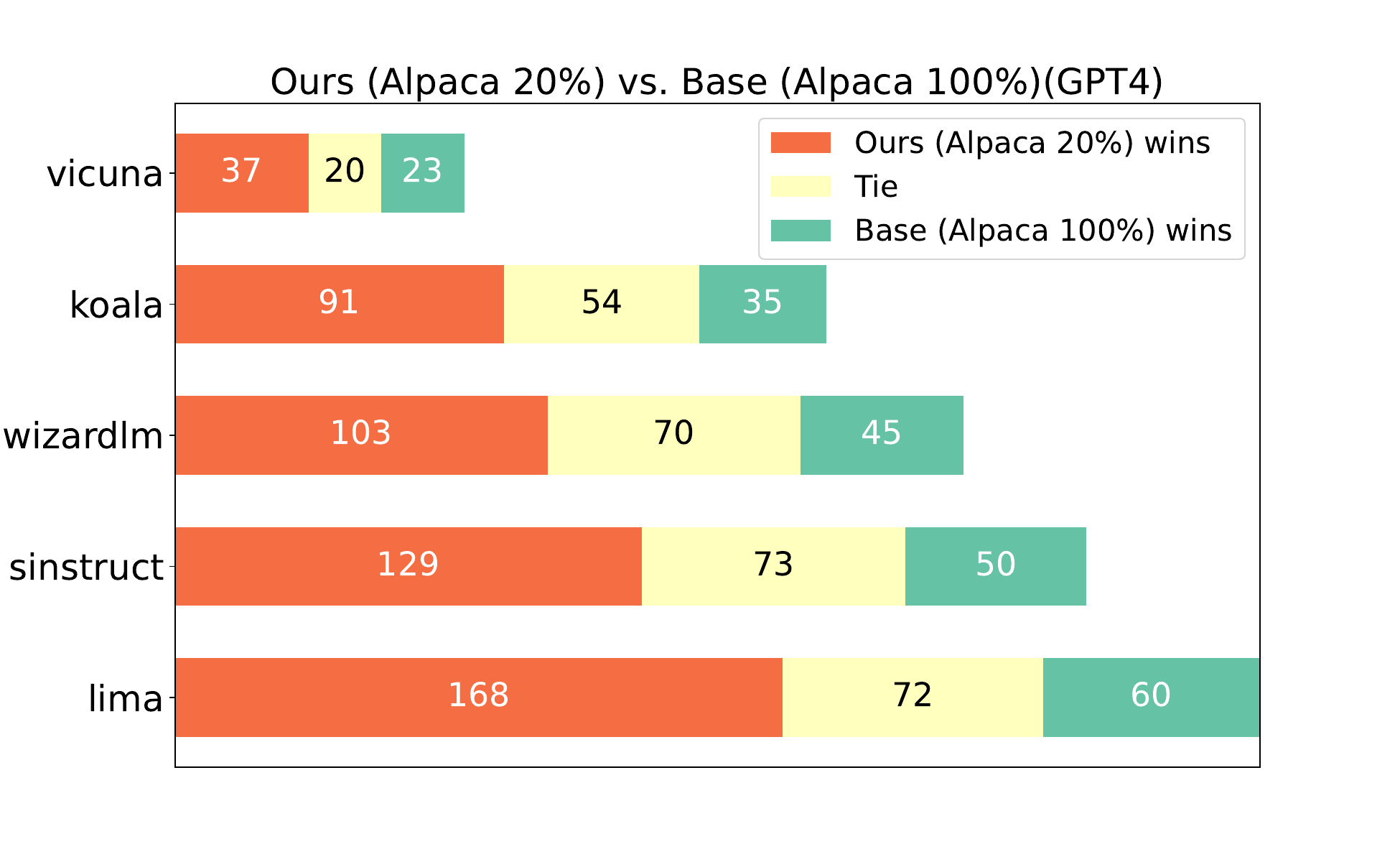}
		\label{fig:6d}
	}\ \ \
        \subfigure[Comparing between our model trained on selected 25\%
data and the official Alpaca model on LLaMA-7b.]{
		\includegraphics[width=0.47\textwidth]{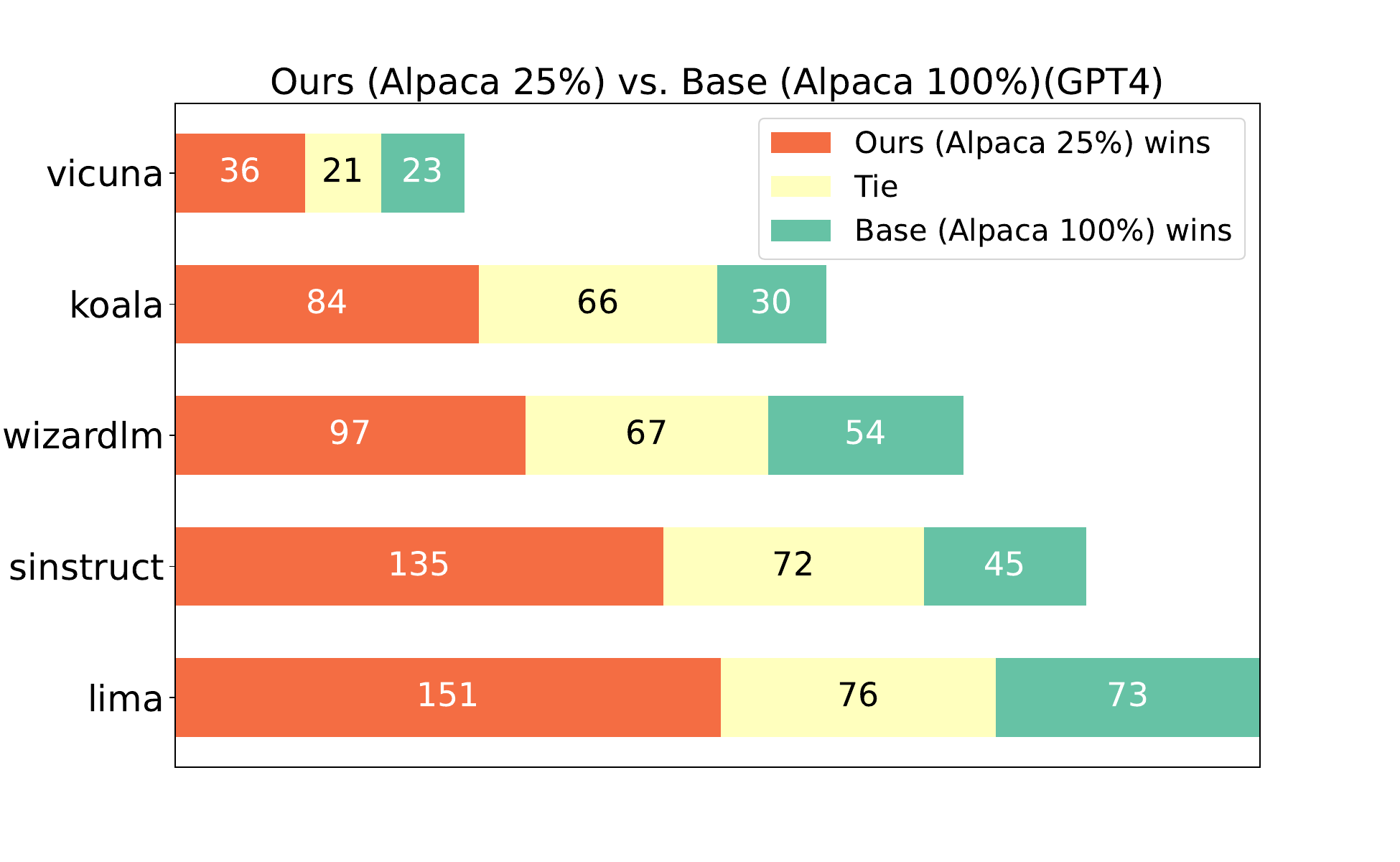}
		\label{fig:6e}
	}\ \ \
\caption{Comparing our models trained on selected
data with full data on LLaMA-7b.}
\label{fig:6_1_7b}
\end{figure*}

\begin{figure*}[!ht]
\centering	
       \subfigure[Comparing between our model trained on selected 1\%
data and the official Alpaca model on LLaMA2-7b.]{
		\includegraphics[width=0.47\textwidth]{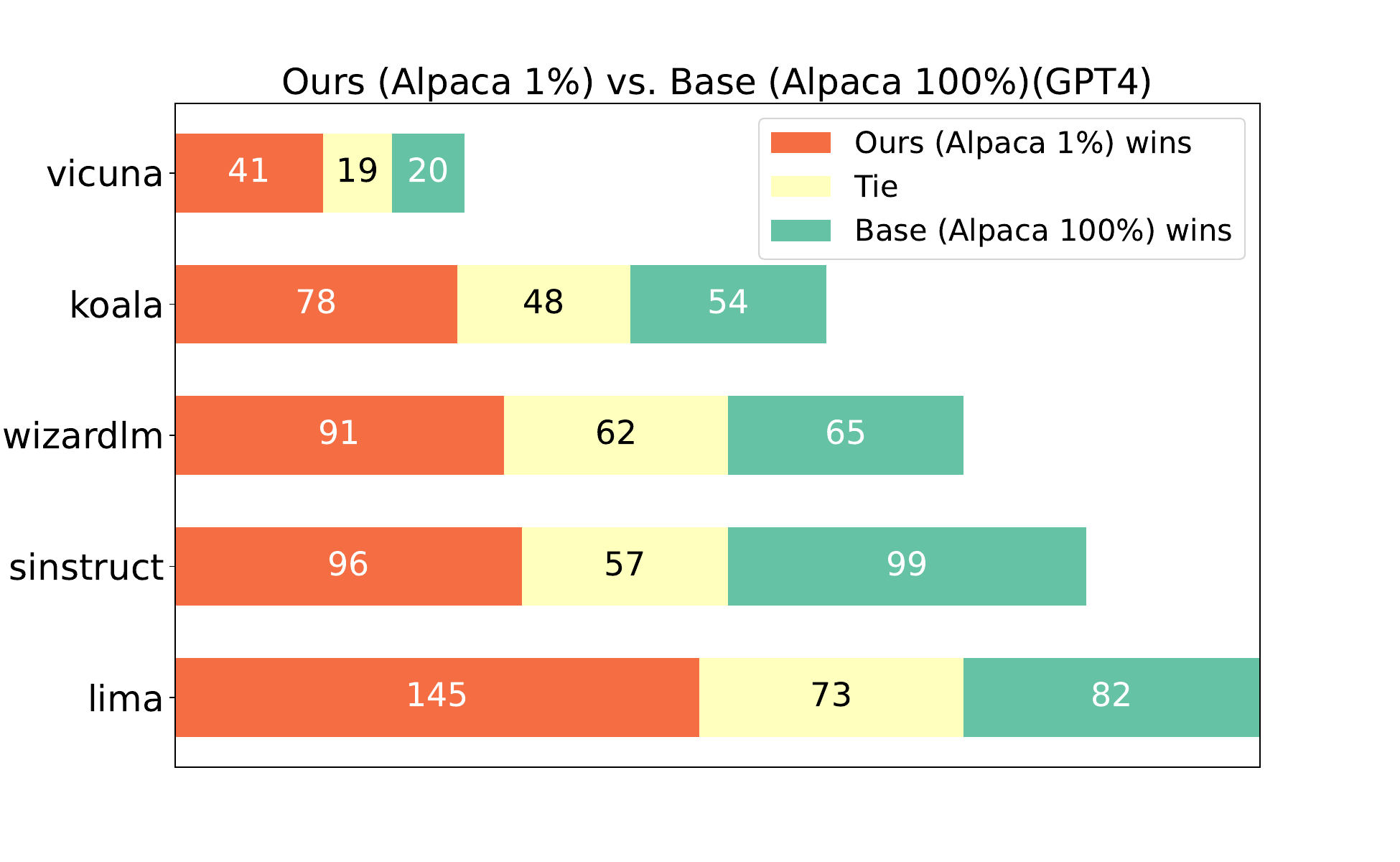}
		\label{fig:before-6f}
	} \ \ \
        \subfigure[Comparing between our model trained on selected 5\%
data and the official Alpaca model on LLaMA2-7b.]{
		\includegraphics[width=0.47\textwidth]{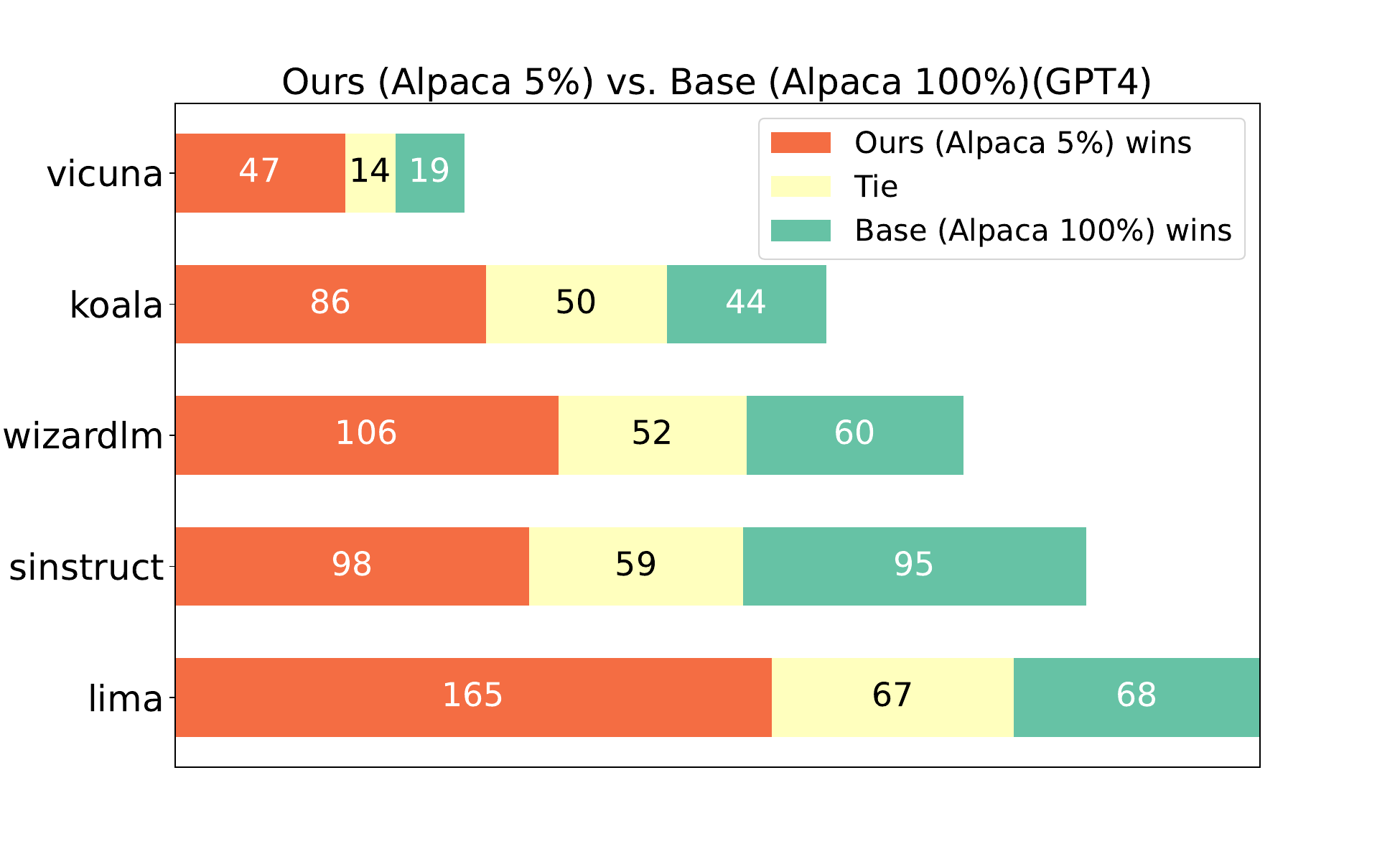}
		\label{fig:6f}
	} \ \ \
        \subfigure[Comparing between our model trained on selected 10\%
data and the official Alpaca model on LLaMA2-7b.]{
		\includegraphics[width=0.47\textwidth]{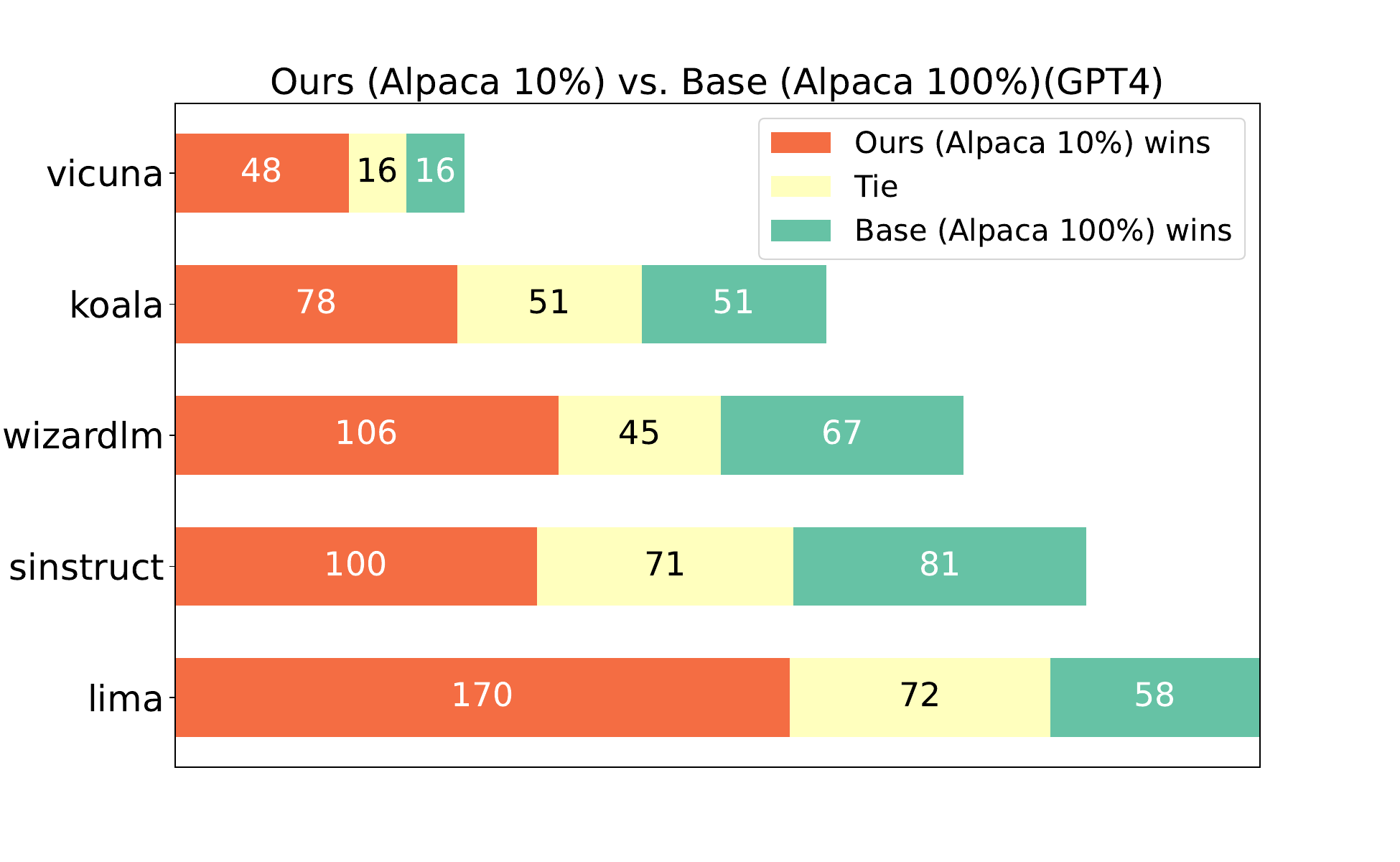}
		\label{fig:6g}
	}\ \ \
        \subfigure[Comparing between our model trained on selected 15\%
data and the official Alpaca model on LLaMA2-7b.]{
		\includegraphics[width=0.47\textwidth]{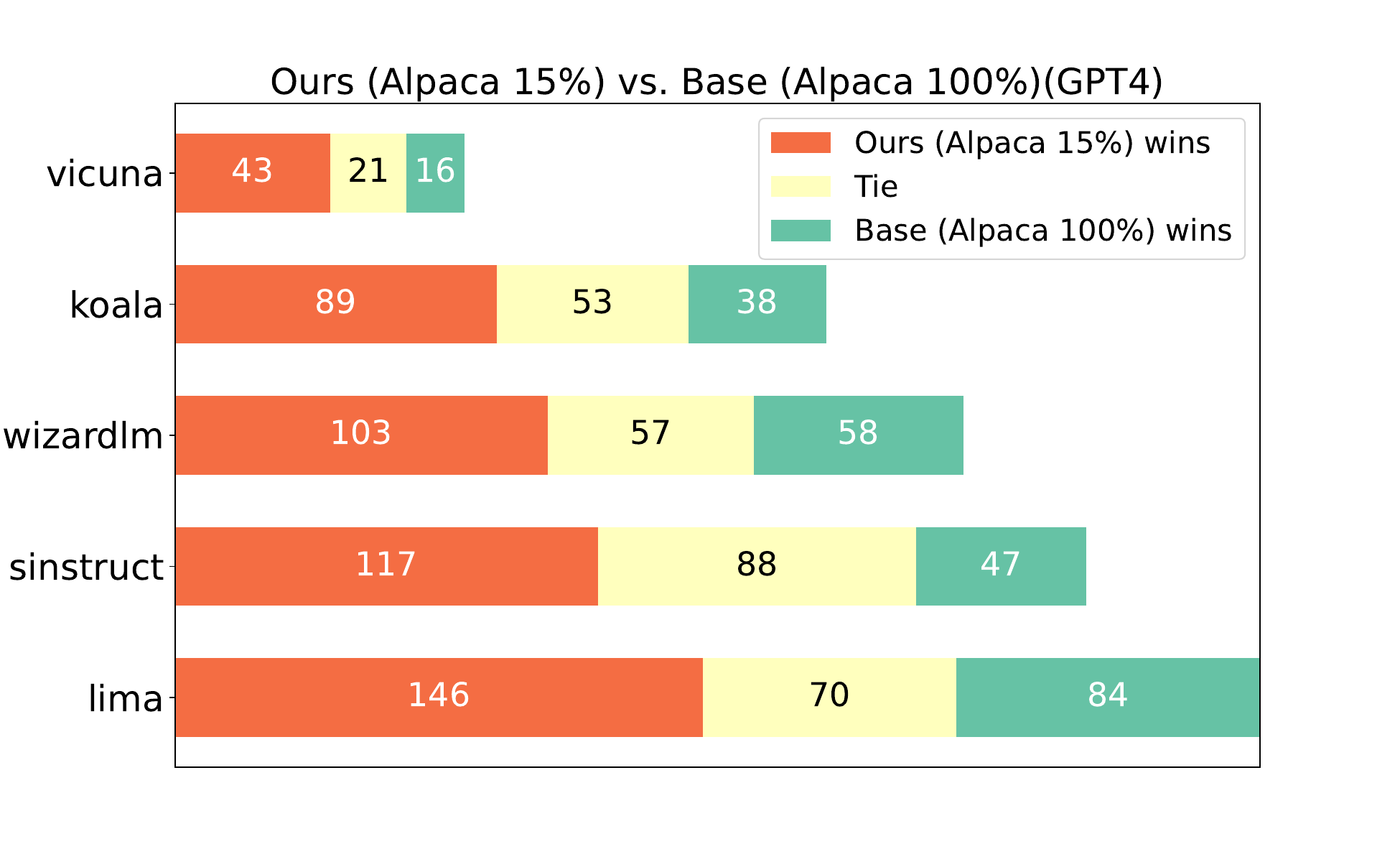}
		\label{fig:6h}
	}\ \ \ 
        \subfigure[Comparing between our model trained on selected 20\%
data and the official Alpaca model on LLaMA2-7b.]{
		\includegraphics[width=0.47\textwidth]{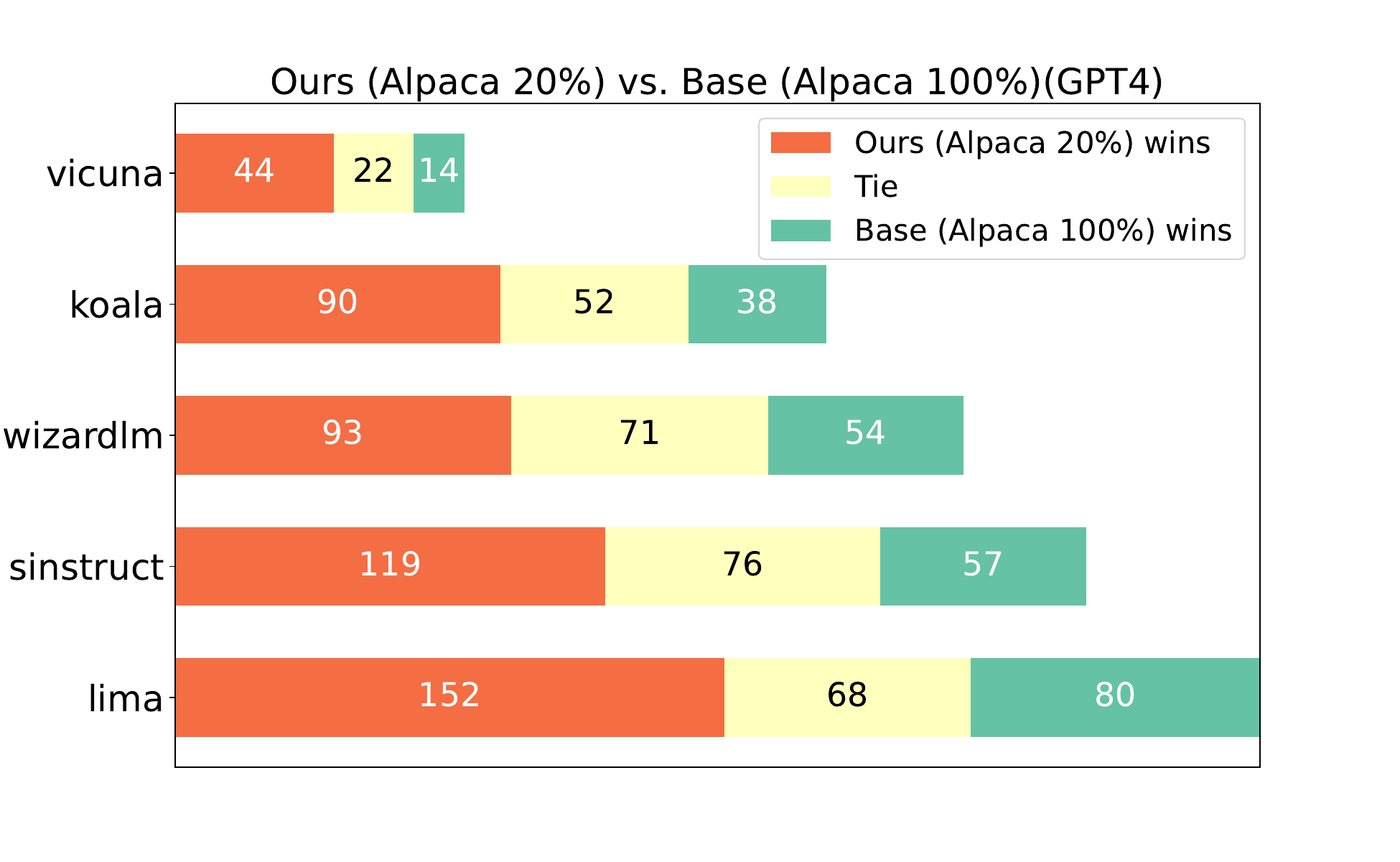}
		\label{fig:6i}
	}\ \ \
    \subfigure[Comparing between our model trained on selected 25\%
data and the official Alpaca model on LLaMA2-7b.]{
		\includegraphics[width=0.47\textwidth]{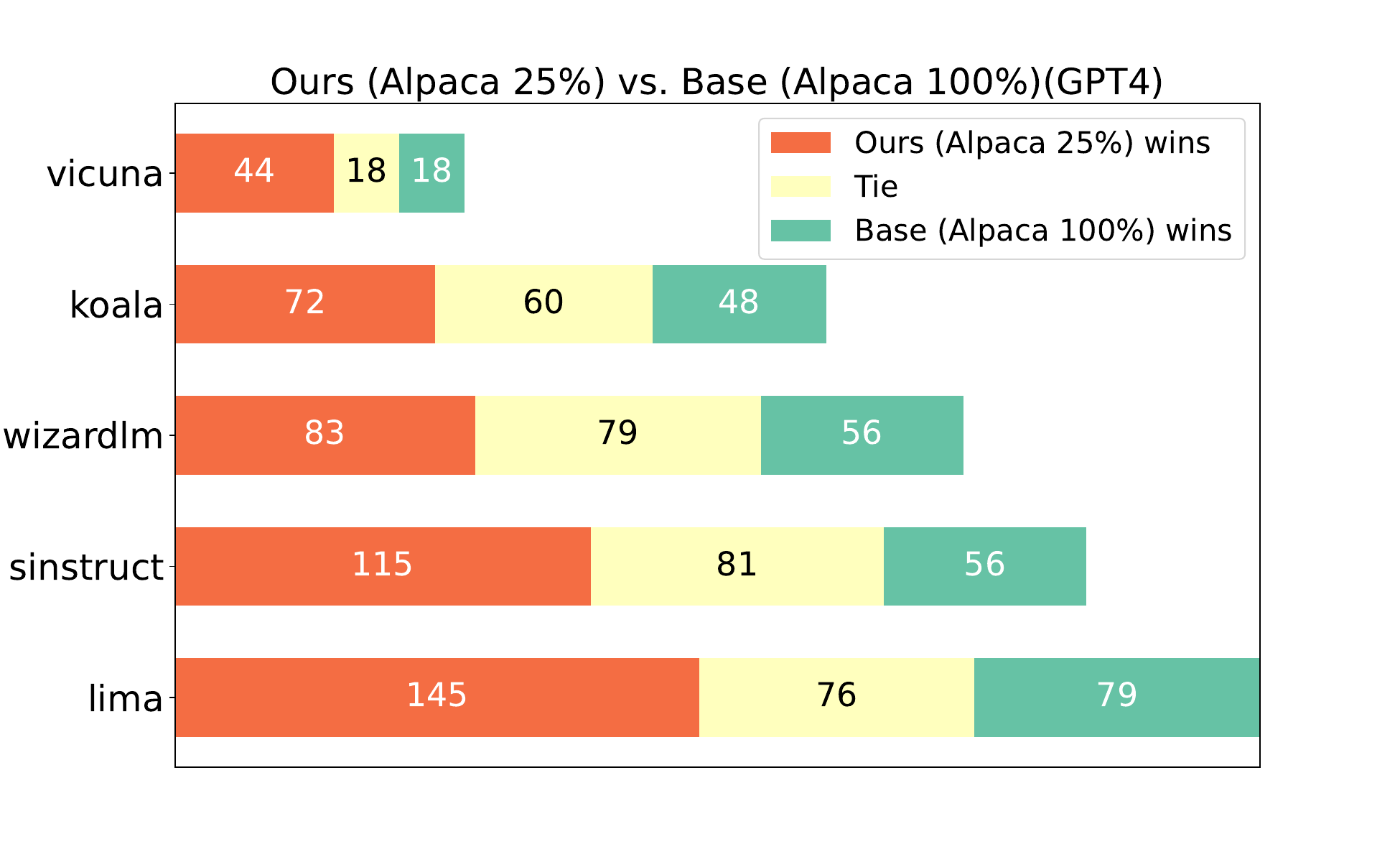}
		\label{fig:6j}
	}\ \ \
	\caption{Comparing our models trained on selected
data with full data on LLaMA2-7b.}
\label{fig:6_1_27b}
\end{figure*}

As illustrated in Figure~\ref{fig:5_1},~\ref{fig:6_1_7b}, and~\ref{fig:6_1_27b}, we utilize GPT-4 as an evaluator and employed Alpaca data subsets selected by MIWV at 1\%, 5\%, 10\%, 15\%, 20\%, and 25\% proportions for instruction tuning on the LLaMA-7b and LLaMA2-7b pretrained language models. We conduct detailed comparisons with the official Alpaca model across five test sets. The experimental results indicate that using only the top 1\% of the selected data can outperform the performance achieved by training on the full dataset. Although our model shows a declining win rate when the data proportion exceeds 20\%, it consistently outperforms the official Alpaca model.

\clearpage

\clearpage

\clearpage

\subsection{Results on the WizardLM Dataset}
\begin{figure*}[!ht]
	\centering	
  	\subfigure[The distribution of the number of win and win rates for the WizardLM instruction dataset on different models under the Vicuna test set.]{
		\includegraphics[width=0.47\textwidth]{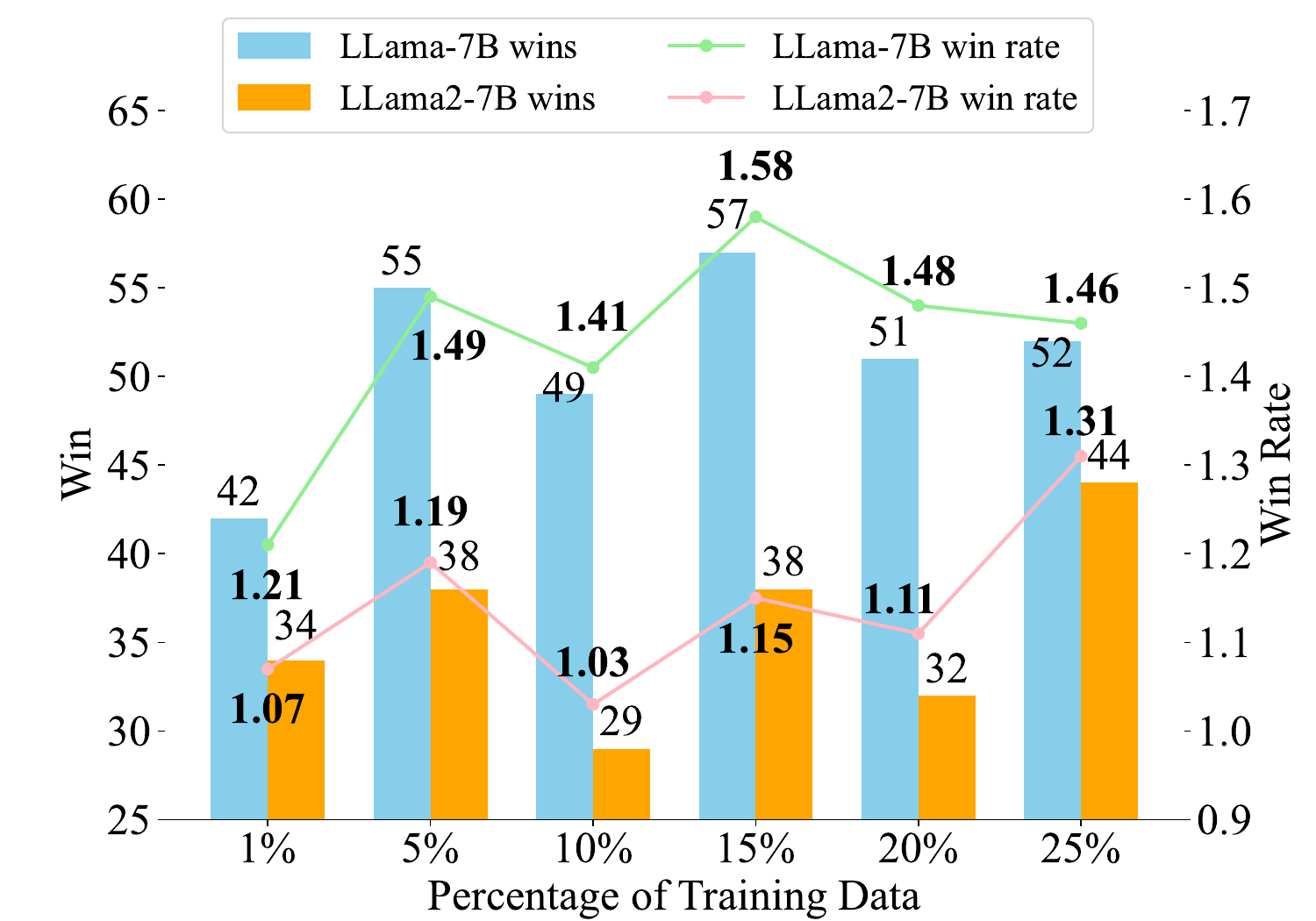}
		\label{fig:7a}
	}\ \ \
        \subfigure[The distribution of the number of win and win rates for the WizardLM instruction dataset on different models under the Koala test set.]{
		\includegraphics[width=0.47\textwidth]{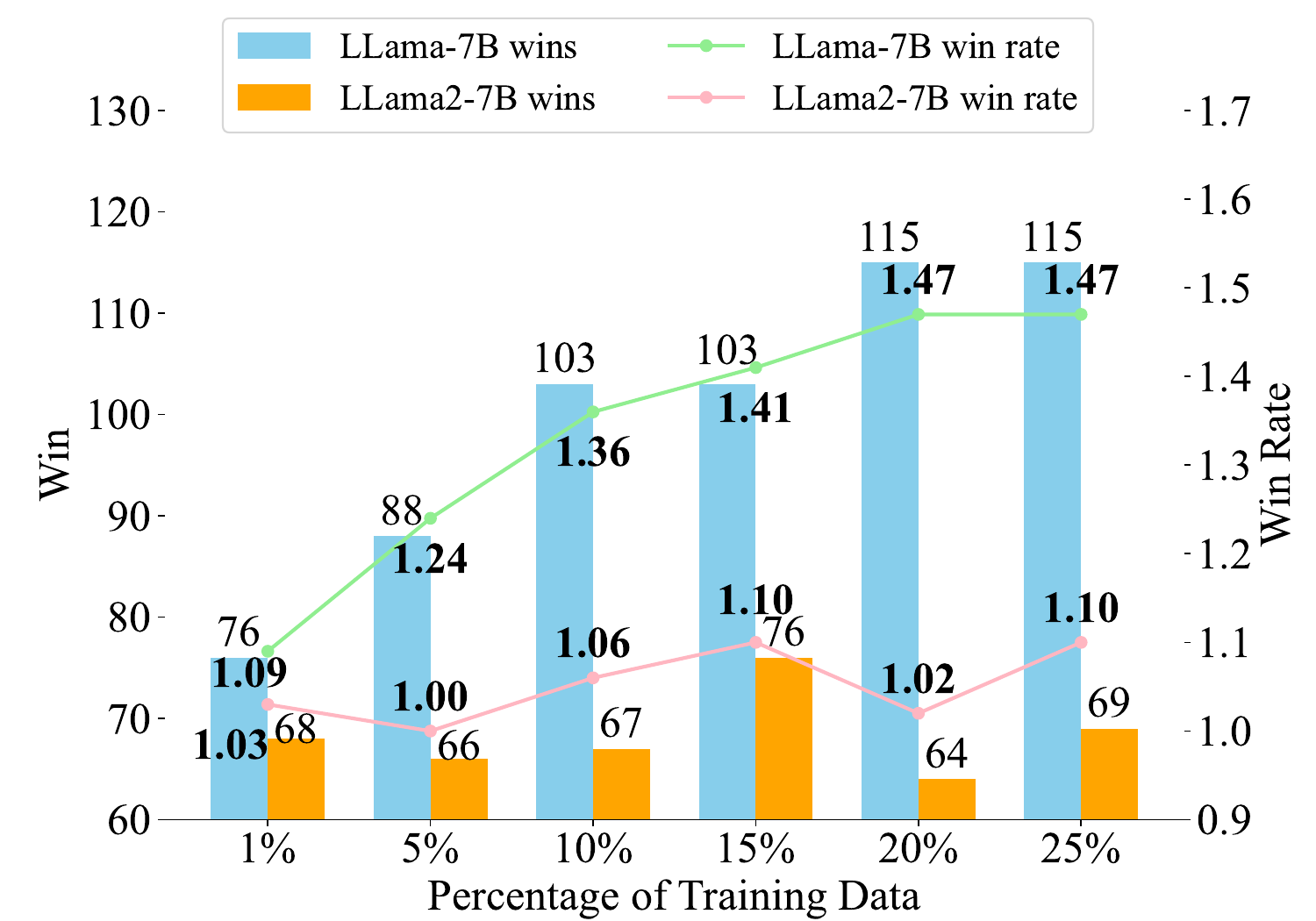}
		\label{fig:7b}
	}\ \ \
        \subfigure[The distribution of the number of win and win rates for the WizardLM instruction dataset on different models under the WizardLM test set.]{
		\includegraphics[width=0.47\textwidth]{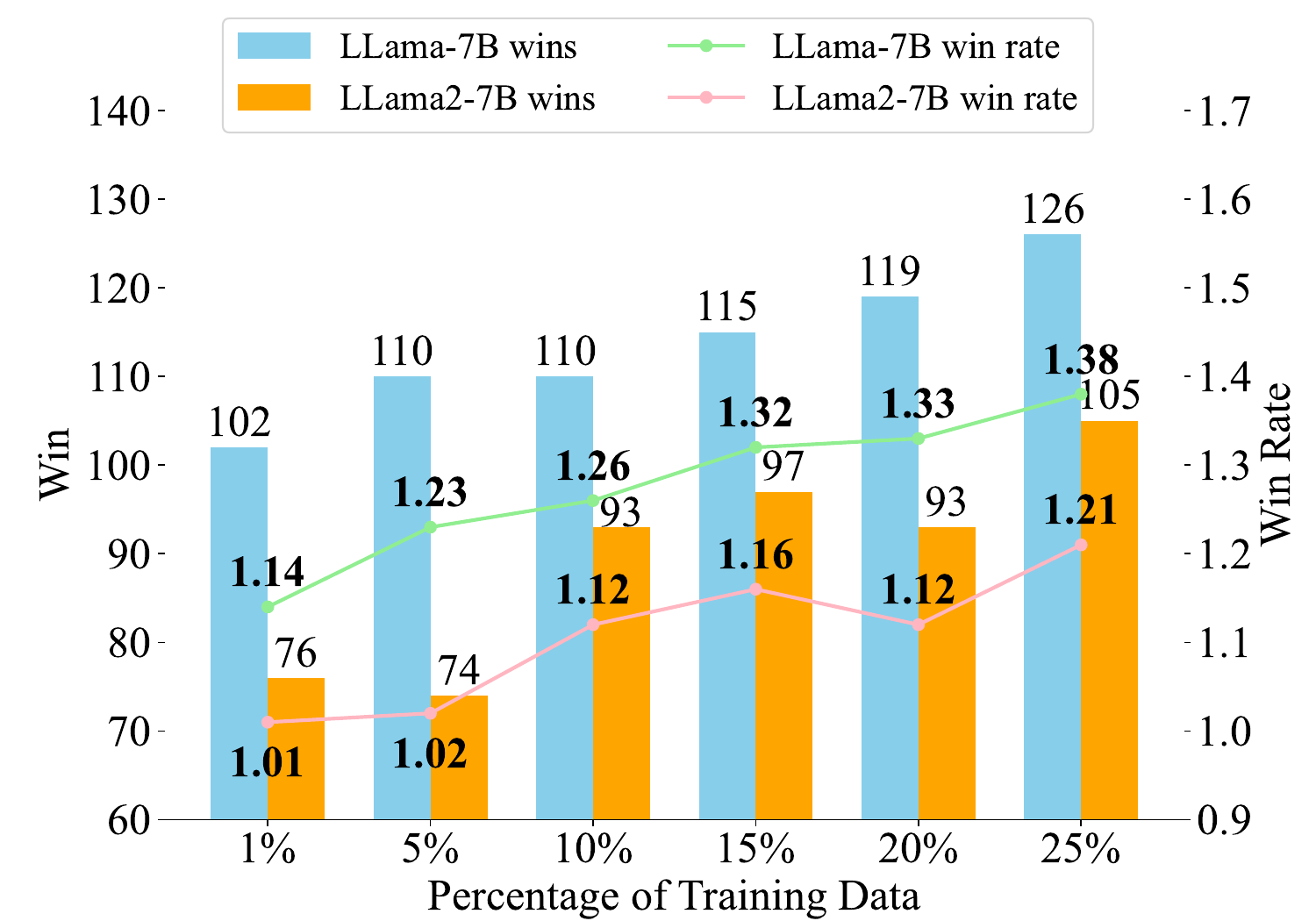}
		\label{fig:7c}
	}\ \ \
        \subfigure[The distribution of the number of win and win rates for the WizardLM instruction dataset on different models under the Lima test set.]{
		\includegraphics[width=0.47\textwidth]{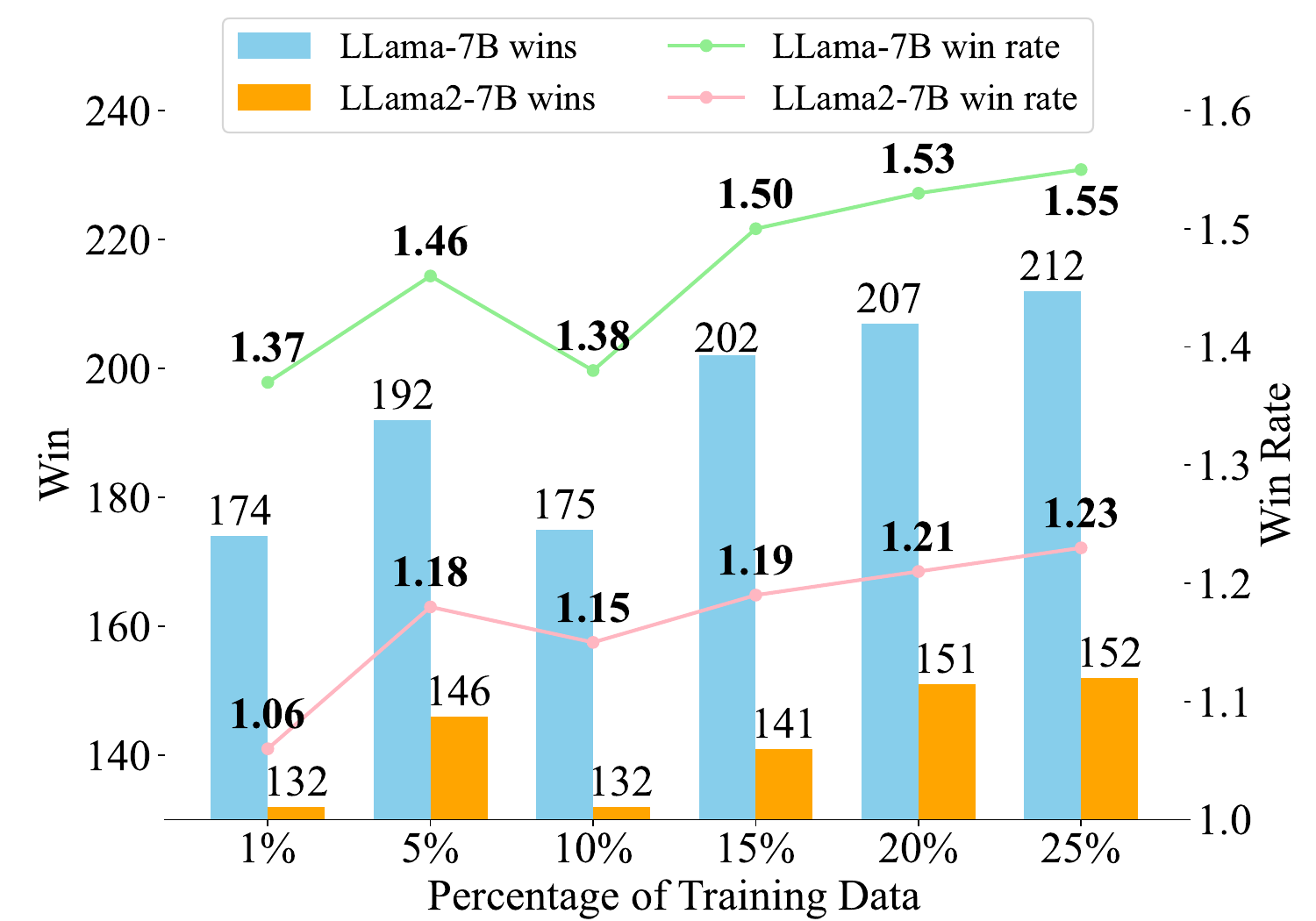}
		\label{fig:7d}
	}\ \ \
	\caption{Comparing our models trained on selected data with full data on the WizardLM dataset. (a), (b), (c), (d) All use GPT4 as the judge.}
\label{fig:7_1}
\end{figure*}

\begin{figure*}[!ht]
    \subfigure[Comparing between our model trained on selected 1\%
data and the WizardLM model trained on 100\% of the data on LLaMA-7b.]{
		\includegraphics[width=0.47\textwidth]{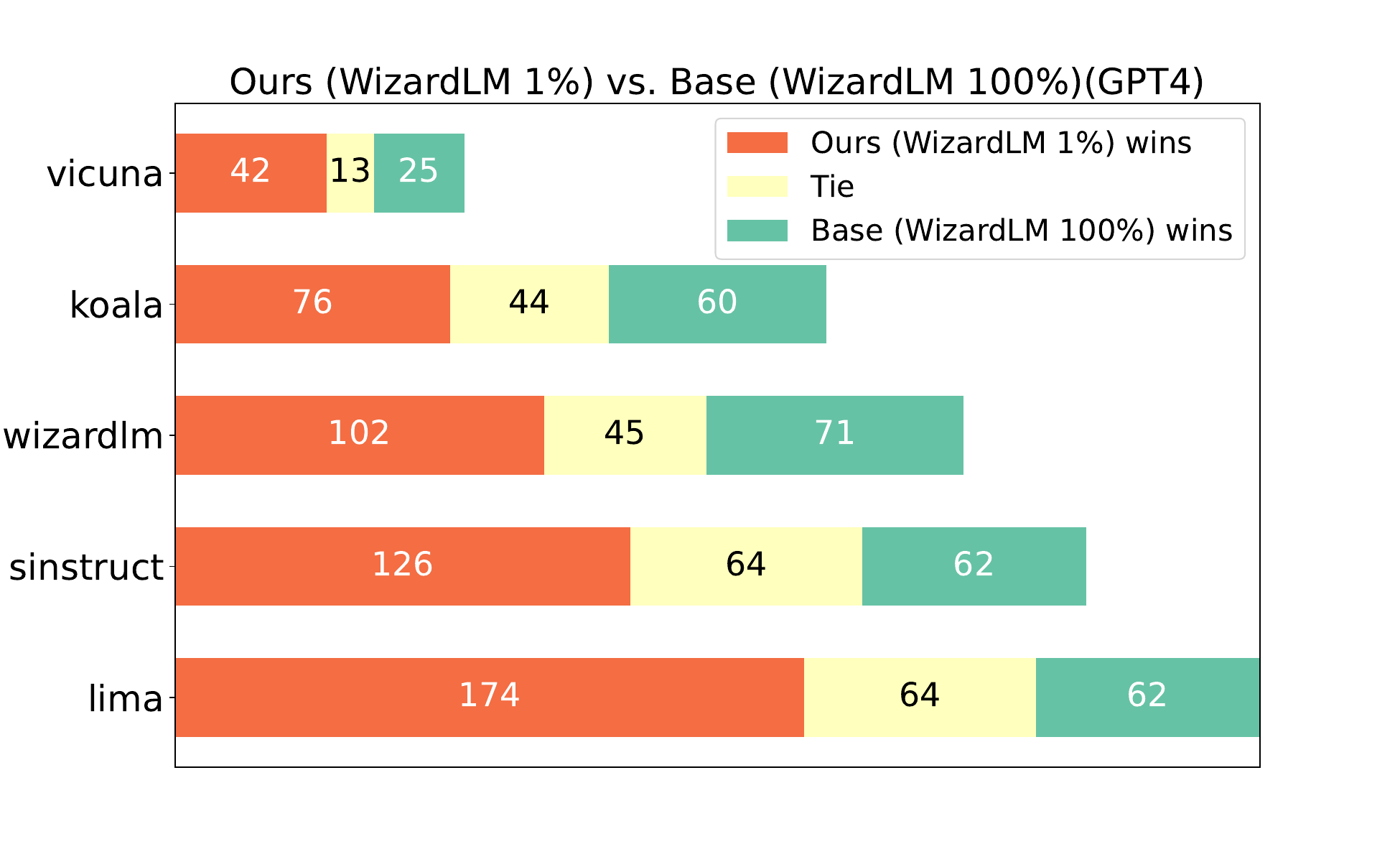}
		\label{fig:8a}
	}\ \ \
  	\subfigure[Comparing between our model trained on selected 5\%
data and the WizardLM model trained on 100\% of the data on LLaMA-7b.]{
		\includegraphics[width=0.47\textwidth]{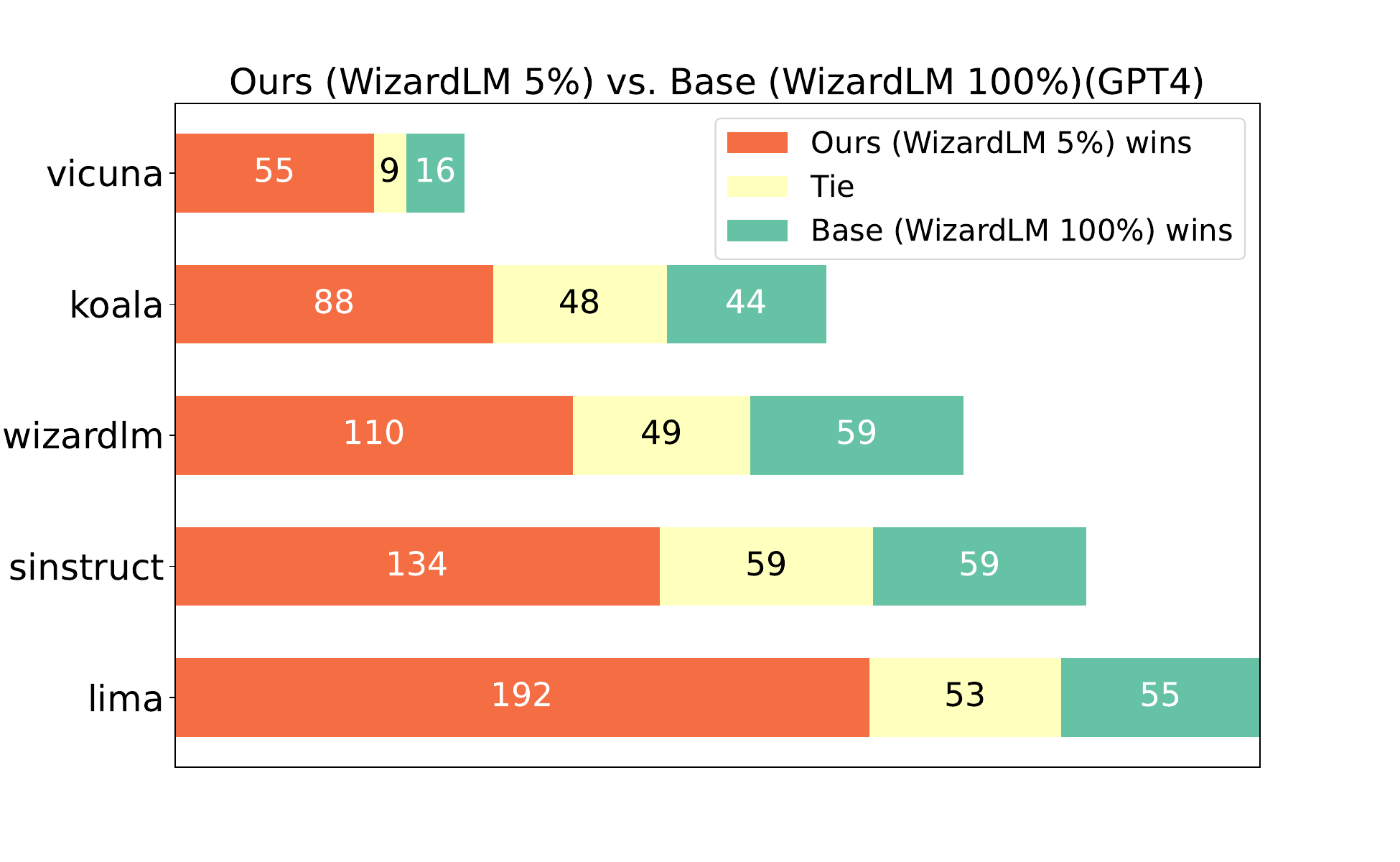}
		\label{fig:8b}
	}\ \ \
        \subfigure[Comparing between our model trained on selected 10\%
data and the WizardLM model trained on 100\% of the data on LLaMA-7b.]{
		\includegraphics[width=0.47\textwidth]{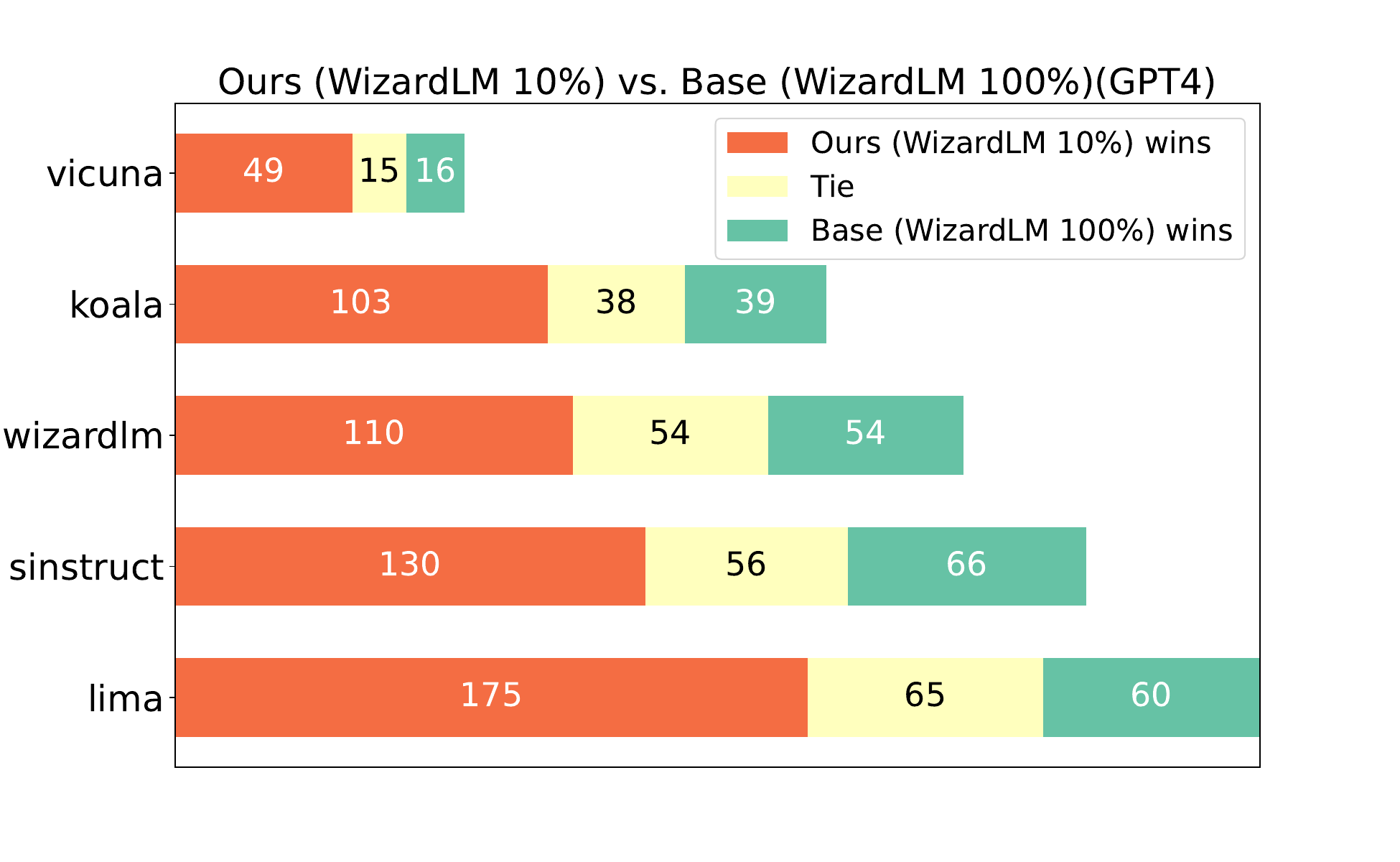}
		\label{fig:8c}
	}\ \ \
        \subfigure[Comparing between our model trained on selected 15\%
data and the Wizardlm model trained on 100\% of the data on LLaMA-7b.]{
		\includegraphics[width=0.47\textwidth]{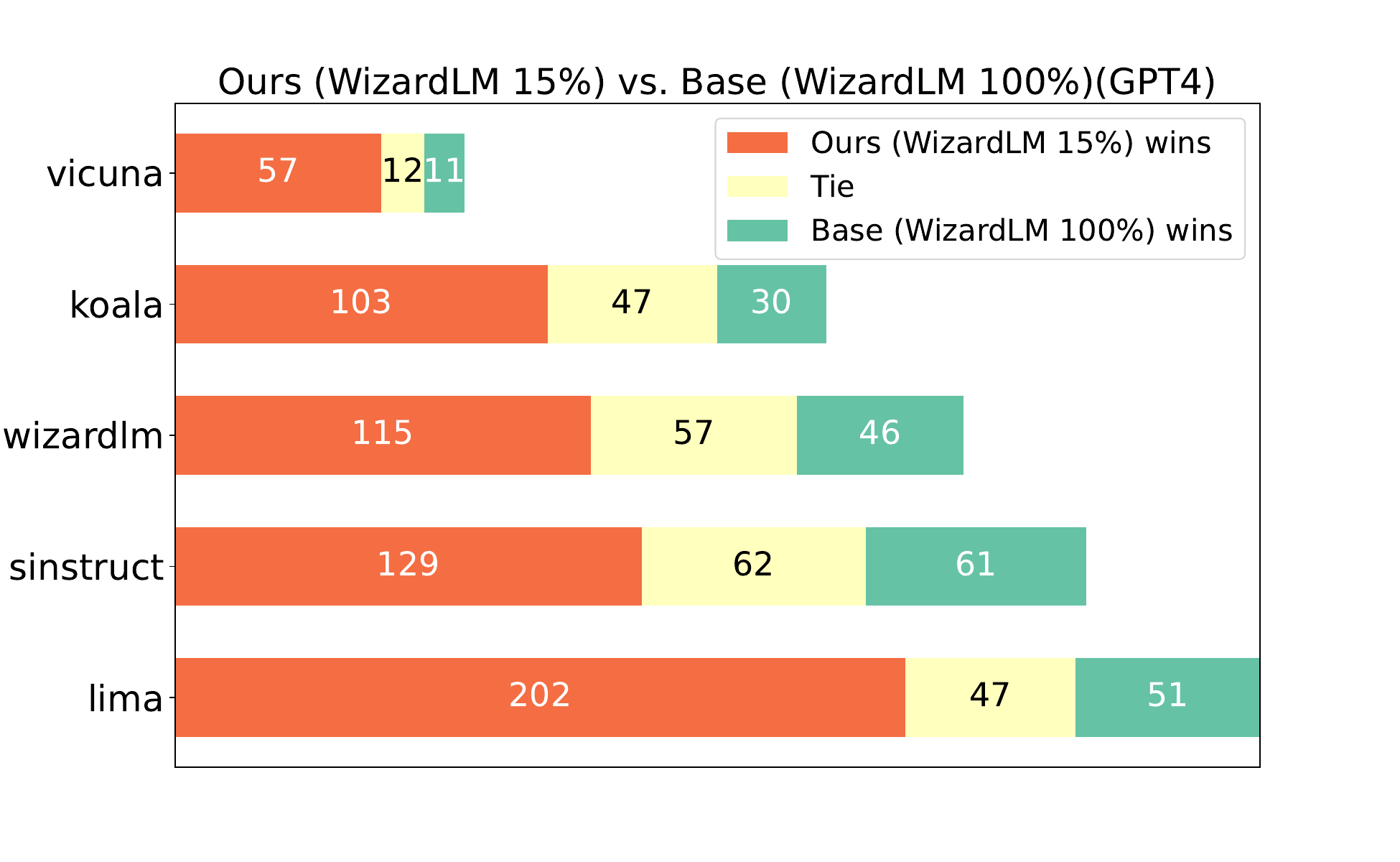}
		\label{fig:8d}
	}\ \ \
        \subfigure[Comparing between our model trained on selected 20\%
data and the WizardLM model trained on 100\% of the data on LLaMA-7b.]{
		\includegraphics[width=0.47\textwidth]{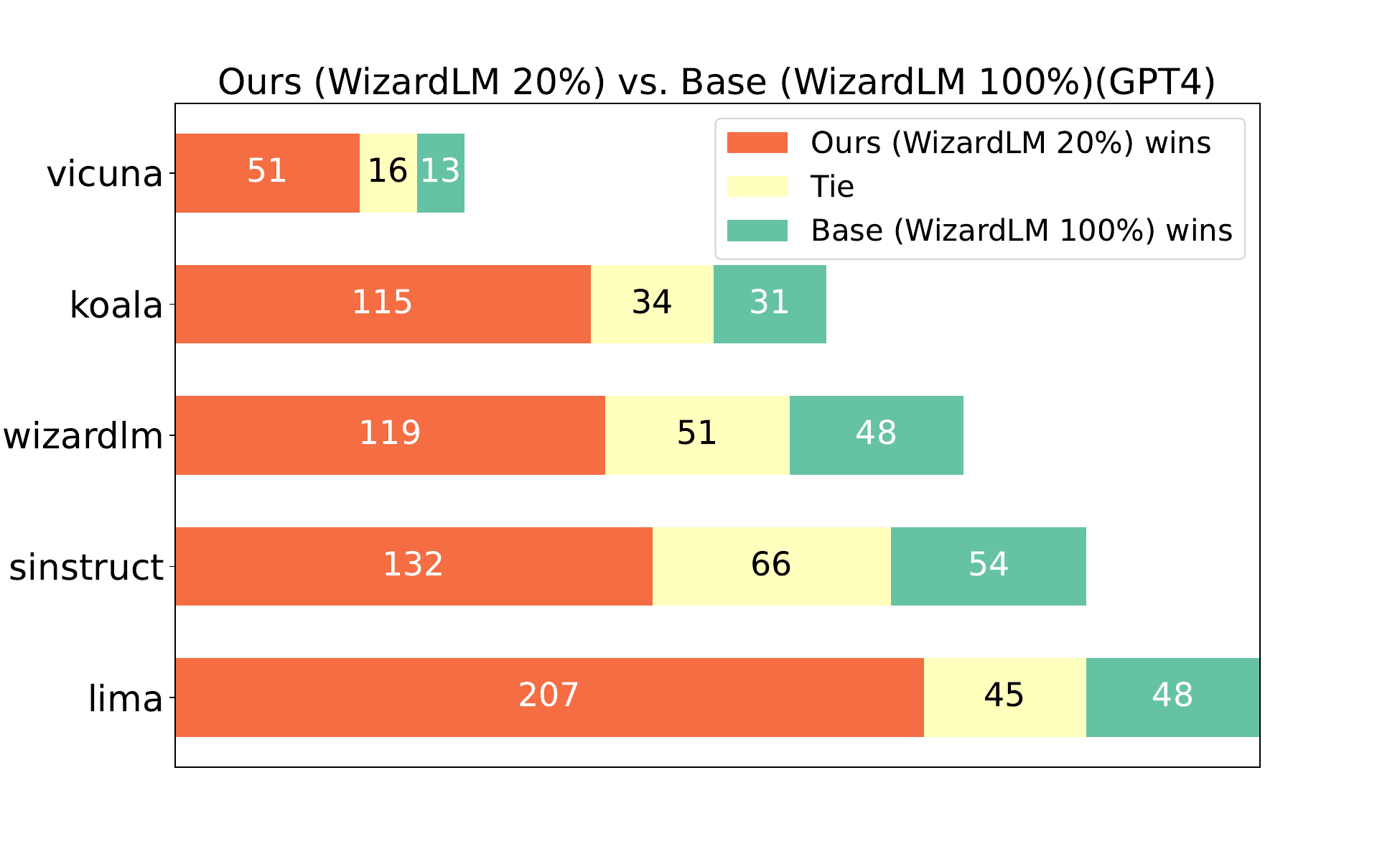}
		\label{fig:8e}
	}\ \ \
        \subfigure[Comparing between our model trained on selected 25\%
data and the WizardLM model trained on 100\% of the data on LLaMA-7b.]{
		\includegraphics[width=0.47\textwidth]{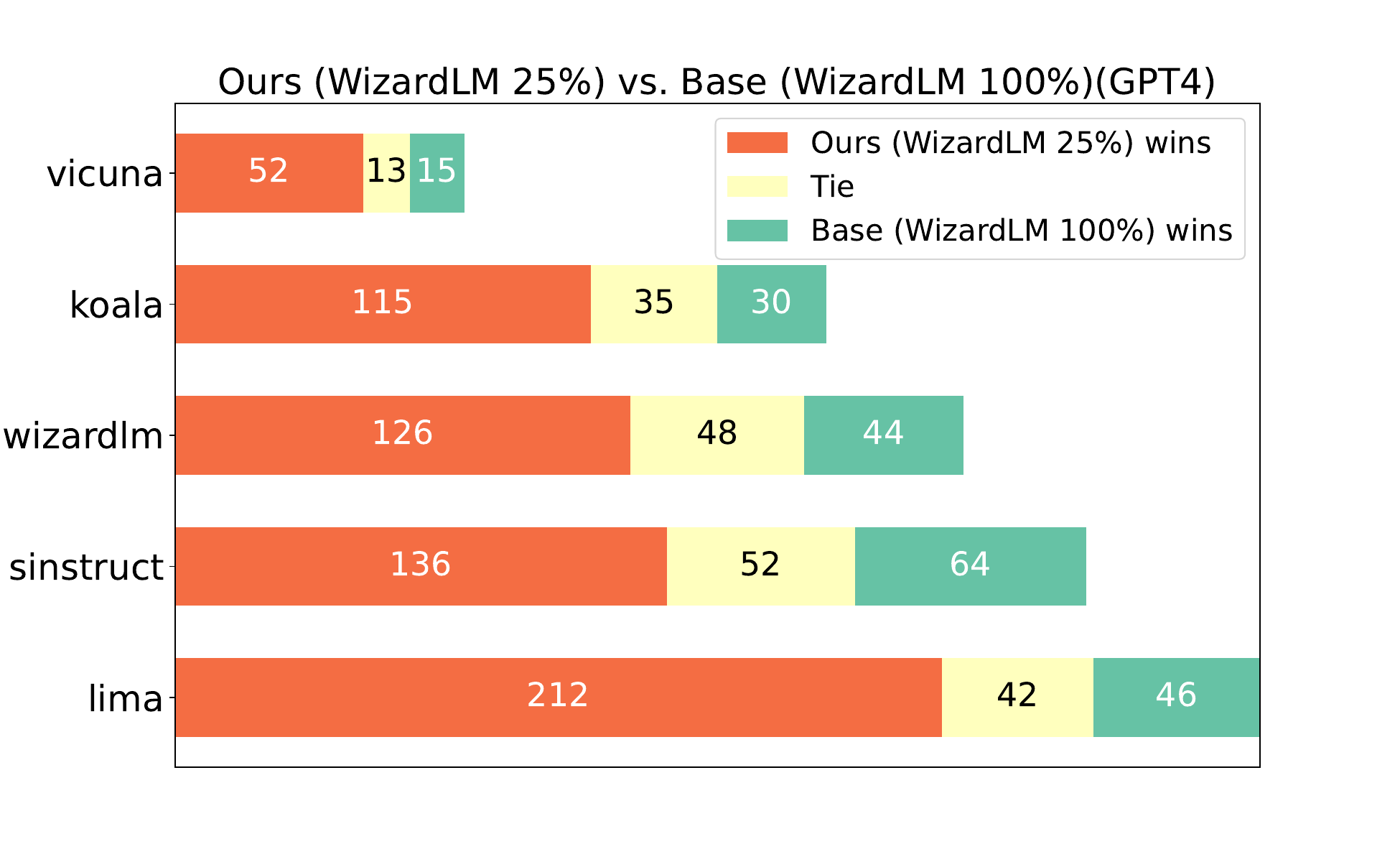}
		\label{fig:8f}
	}\ \ \
\caption{Comparing our models trained on selected
data with full data on LLaMA-7b.}
\label{fig:8_1_7b}  

\end{figure*}

\begin{figure*}[!ht]
        \subfigure[Comparing between our model trained on selected 1\%
        data and the WizardLM model trained on 100\% of the data on LLaMA2-7b.]{
		\includegraphics[width=0.47\textwidth]{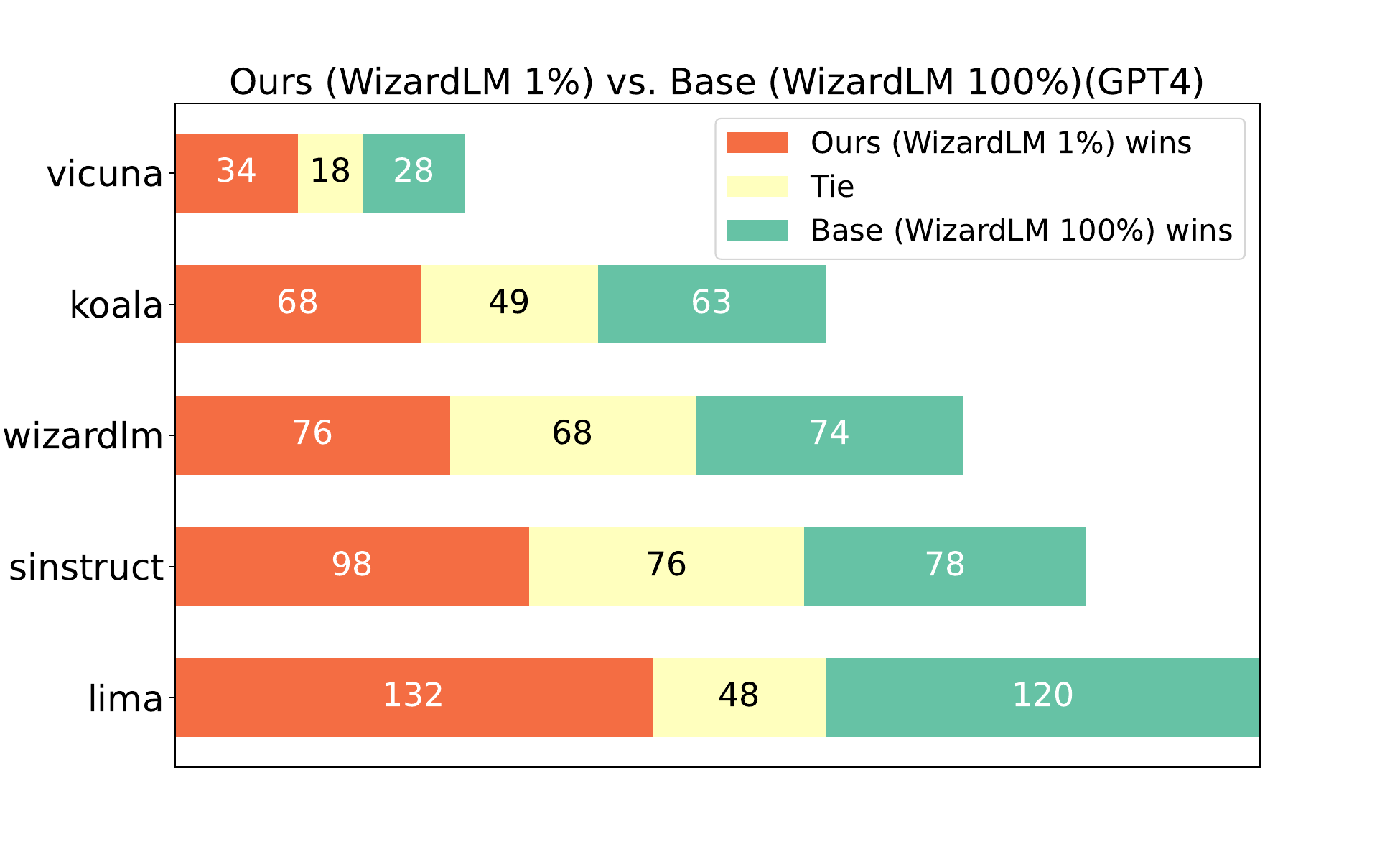}
		\label{fig:before-8f}
	}\ \ \
        \subfigure[Comparing between our model trained on selected 5\%
data and the WizardLM model trained on 100\% of the data on LLaMA2-7b.]{
		\includegraphics[width=0.47\textwidth]{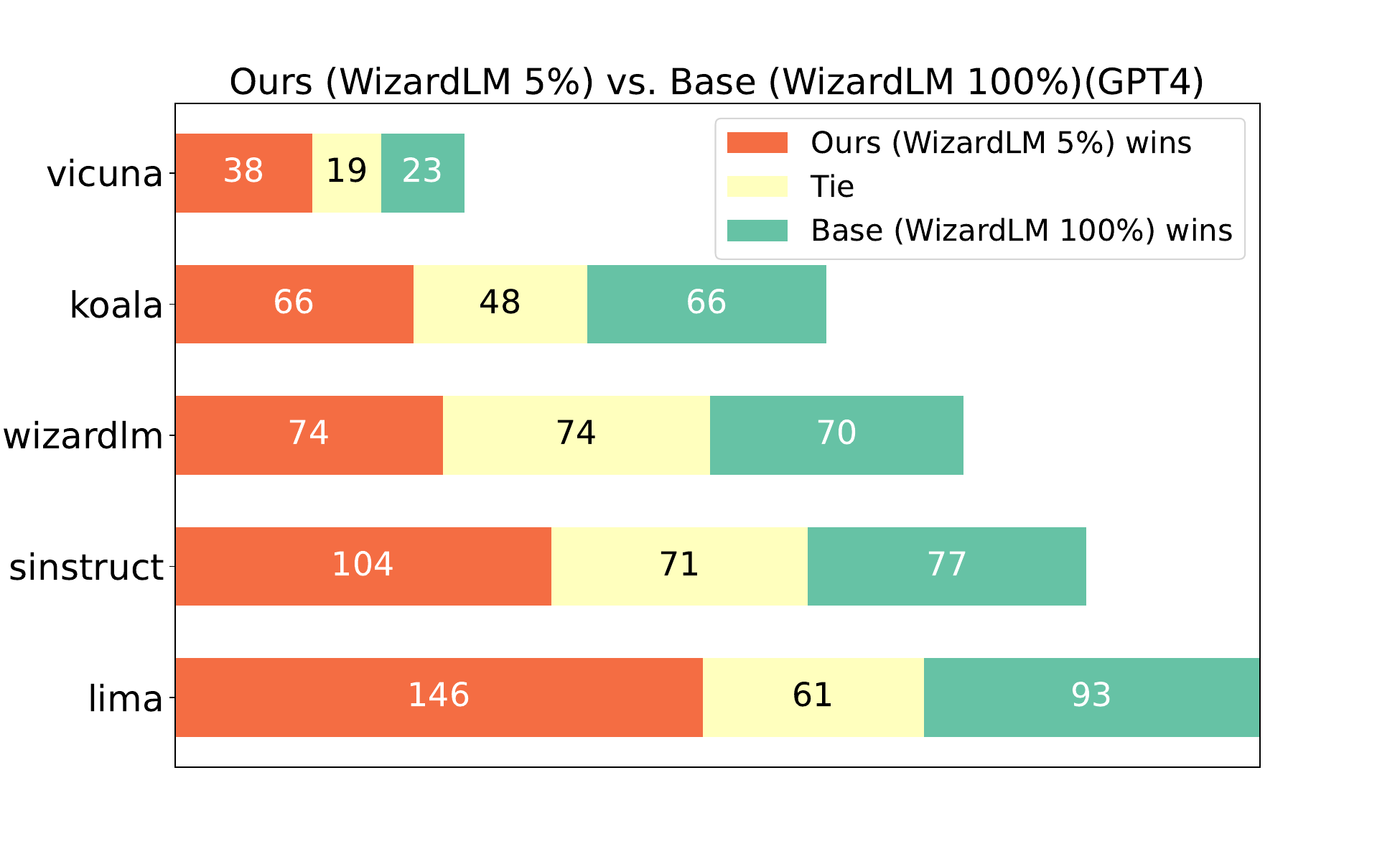}
		\label{fig:8g}
	}\ \ \
        \subfigure[Comparing between our model trained on selected 10\%
data and the WizardLM model trained on 100\% of the data on LLaMA2-7b.]{
		\includegraphics[width=0.47\textwidth]{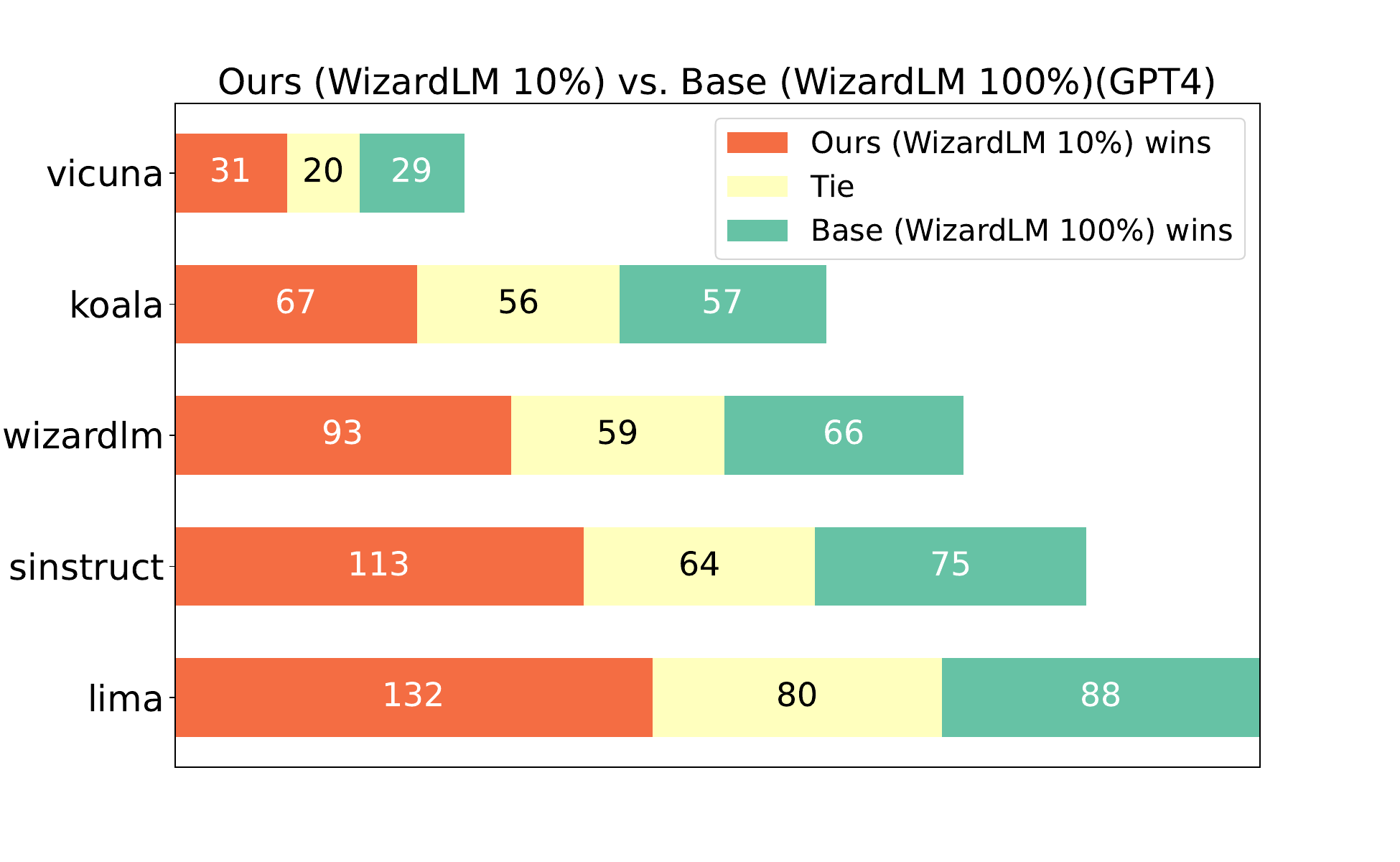}
		\label{fig:8h}
	}\ \ \
        \subfigure[Comparing between our model trained on selected 15\%
data and the WizardLM model trained on 100\% of the data on LLaMA2-7b.]{
		\includegraphics[width=0.47\textwidth]{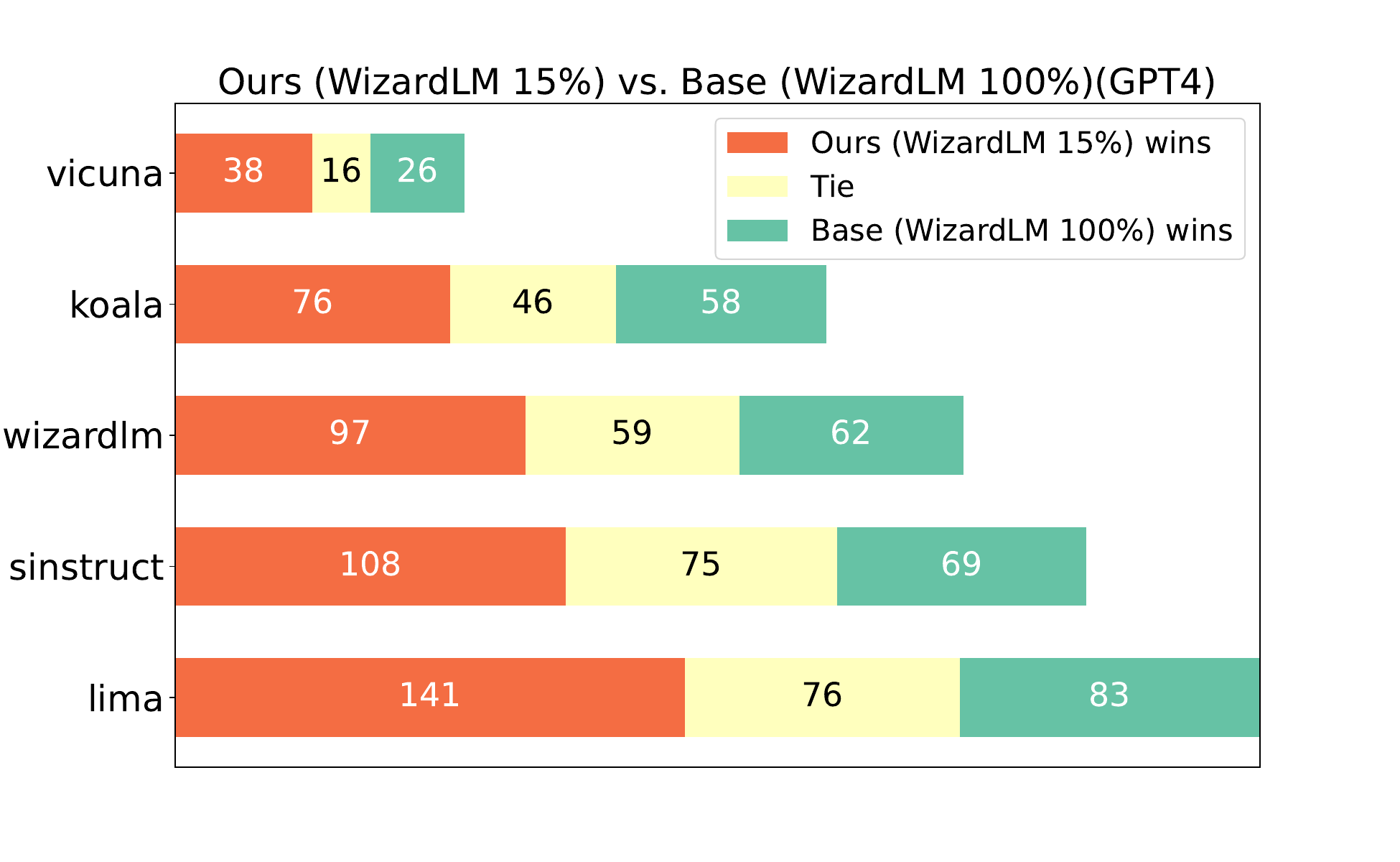}
		\label{fig:8i}
	}\ \ \ 
    
        \subfigure[Comparing between our model trained on selected 20\%
data and the WizardLM model trained on 100\% of the data on LLaMA2-7b.]{
		\includegraphics[width=0.47\textwidth]{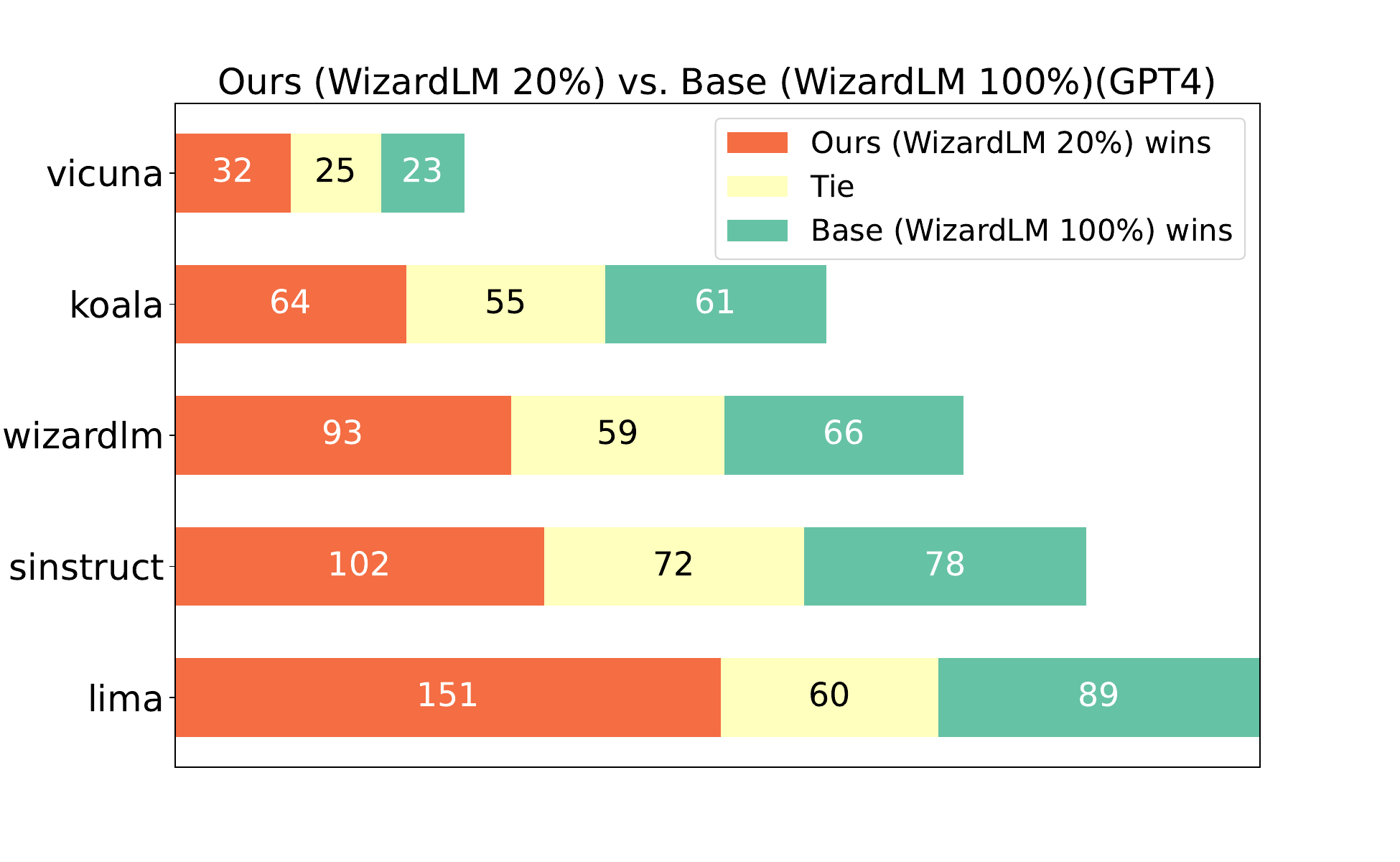}
		\label{fig:8j}
	}\ \ \
    \subfigure[Comparing between our model trained on selected 25\%
data and the WizardLM model trained on 100\% of the data on LLaMA2-7b.]{
		\includegraphics[width=0.47\textwidth]{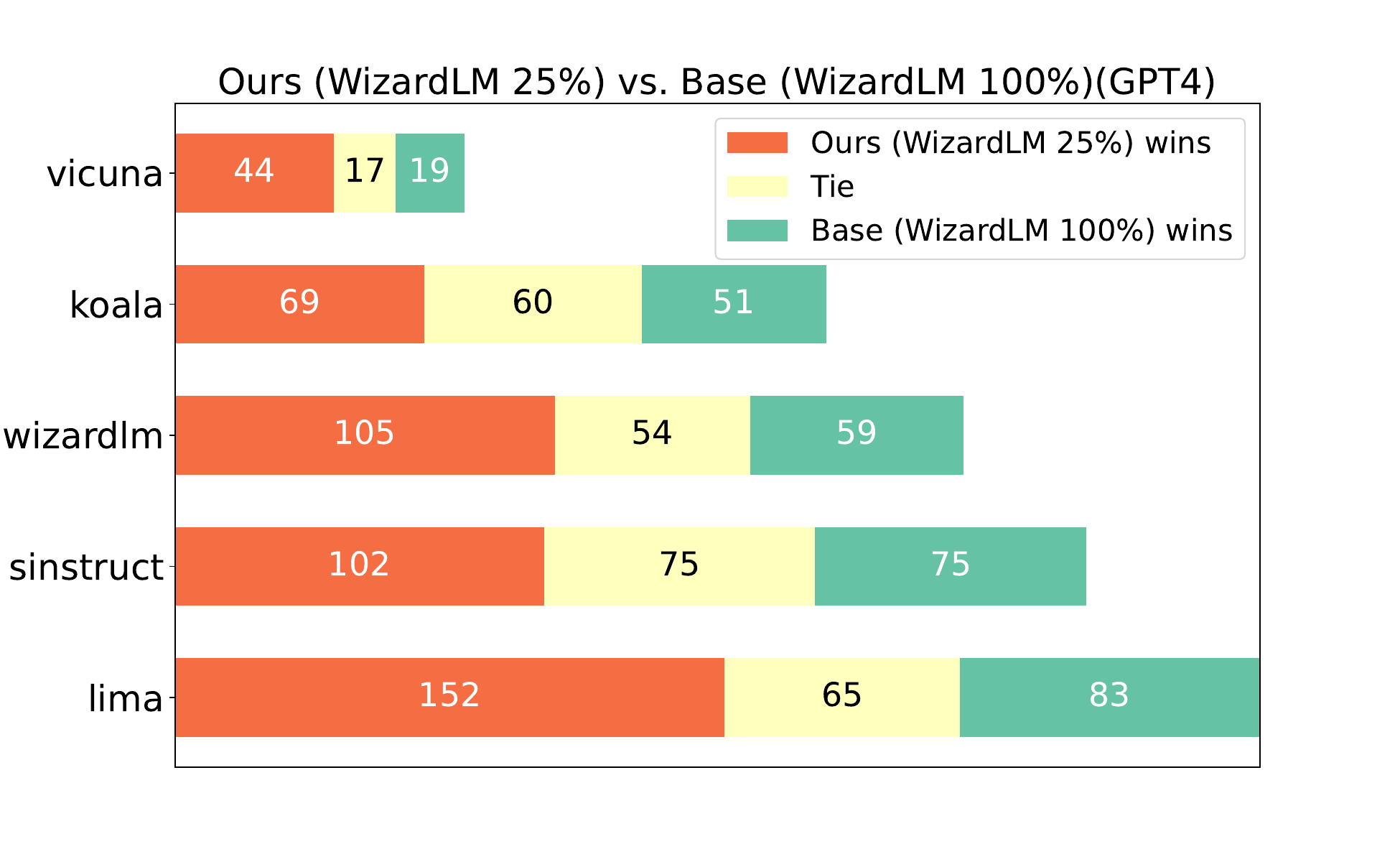}
		\label{fig:8k}
	}\ \ \
\caption{Comparing our models trained on selected
data with full data on LLaMA2-7b.}
\label{fig:8_1_27b}
\end{figure*}

As shown in Figure~\ref{fig:7_1},~\ref{fig:8_1_7b} and~\ref{fig:8_1_27b}, we utilize GPT-4 as an evaluator and employ WizardLM data subsets selected by MIWV at 1\%, 5\%, 10\%, 15\%, 20\%, and 25\% proportions for instruction tuning on the LLaMA-7b and LLaMA2-7b pretrained language models. We conduct detailed comparisons with the WizardLM model trained on 100\% of the data across five test sets. The experimental results indicate that using only the top 1\% of the selected data can outperform the performance obtained by training on the full dataset. Overall, our model achieves the best win rate when using a data proportion of either 15\%, 20\%, or 25\%.

\clearpage

\clearpage

\clearpage

\section{Experiment on NIV2 Dataset}
\label{sec:appendix D}
We conduct experiments on Natural Instructions V2 (NIV2)~\cite{wang2022super}, a leading multi-task NLP dataset. NIV2 is currently one of the largest multi-task learning datasets, encompassing a wide range of task types, including classification and generation~\cite{DBLP:conf/emnlp/KungY0CP23}. These experiments are designed to further validate the effectiveness of our MIWV method.

\noindent\textbf{Train Dataset.} We use the English task division version of NIV2 for experiments, which includes 756 training tasks and 119 test tasks. In order to optimize the use of experimental resources and ensure the reliability of experimental results, we randomly select three sample sets of different sizes from a total of 756 training tasks: 100, 300, and 500 training tasks. For each selected training task, we further randomly select up to 100 instruction samples to construct our training dataset.

\noindent\textbf{Test Dataset.} We refer to the study~\cite{wang2022super} to construct the test dataset, which consists of 20 samples randomly selected from 119 independent test tasks, and finally forms a test dataset containing 2,380 different instructions.

We conduct instruction tuning experiments on T5-11B strictly following the training parameter settings in the work of~\cite{wang2022super}. For the classification and generation tasks involved in the evaluation, we uniformly use the Rouge-L for evaluation. This metric shows a high correlation between the performance in these two types of tasks and the task response accuracy~\cite{wang2022super}. We compare the performance of the two models by comparing the Rouge-L difference between our model and the benchmark model trained using the full data. If the difference is greater than 0, it means that our model outperforms the benchmark model, and a larger difference means a larger lead.

\begin{figure*}[!ht]
	\centering	
  	\subfigure[Experiments with data covering 100 training tasks.]{
		\includegraphics[width=0.5\textwidth]{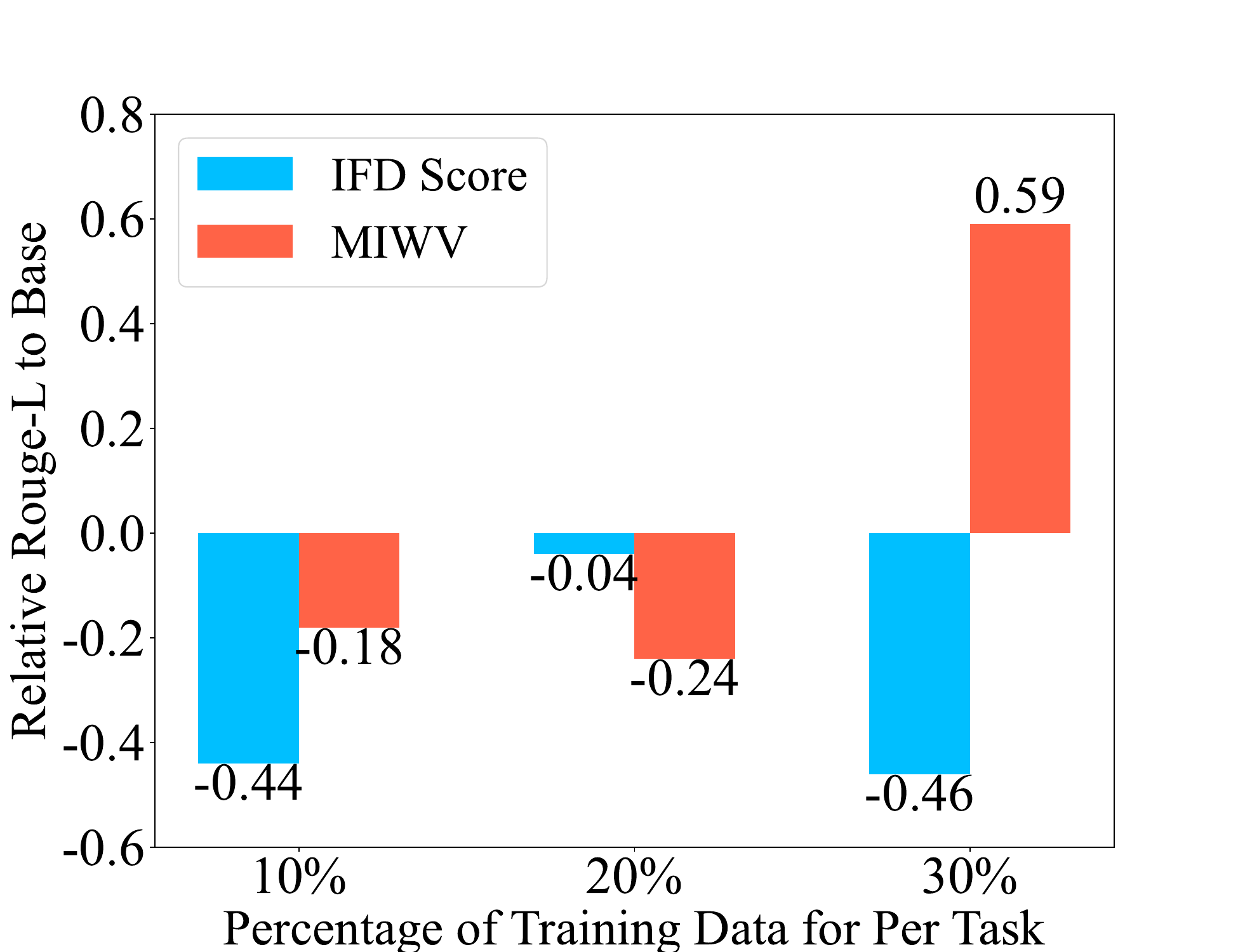}
		\label{fig:3a}
	}\ \ \
        \subfigure[Experiments with data covering 200 training tasks.]{
		\includegraphics[width=0.5\textwidth]{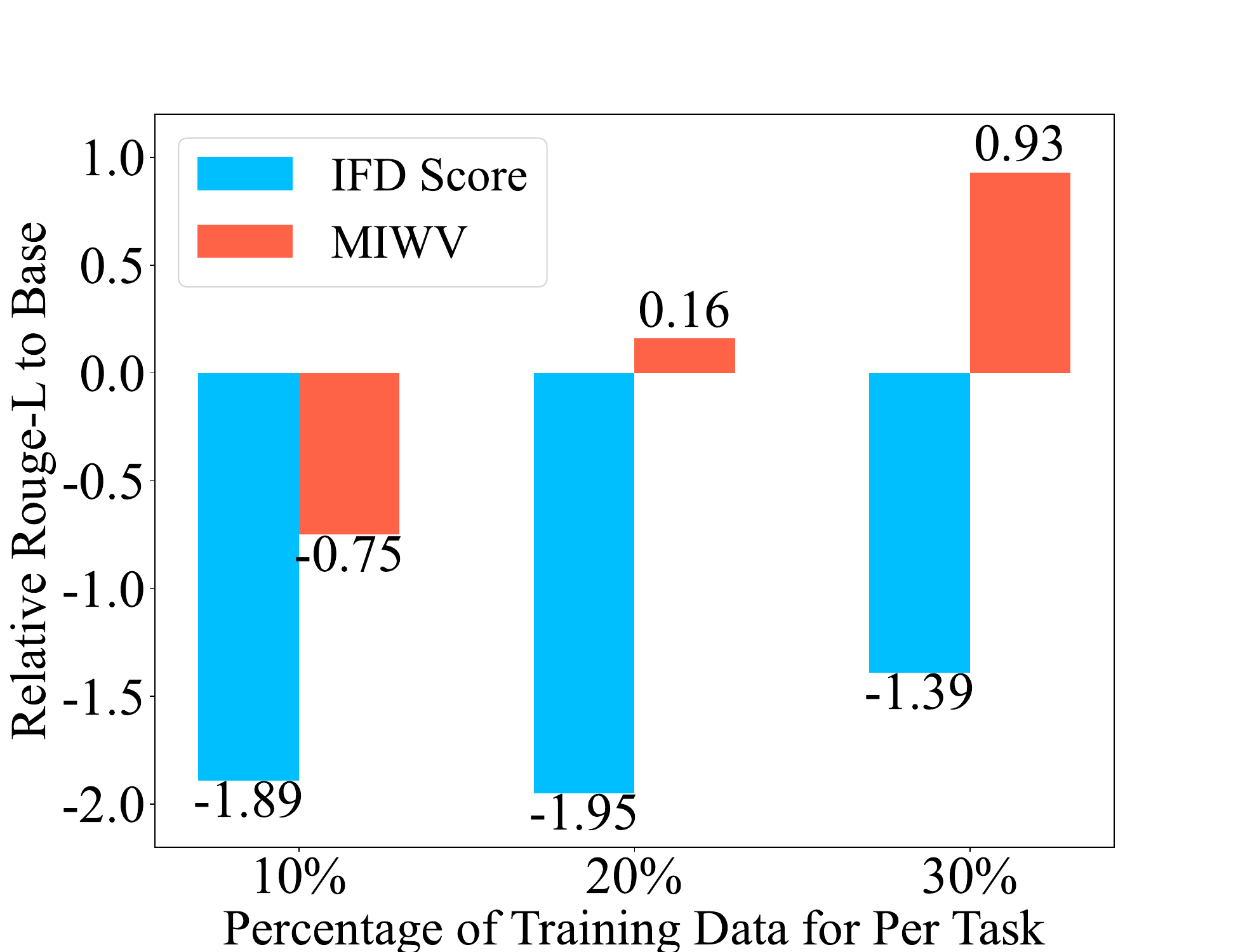}
		\label{fig:3b}
	}\ \ \
        \subfigure[Experiments with data covering 300 training tasks.]{
		\includegraphics[width=0.5\textwidth]{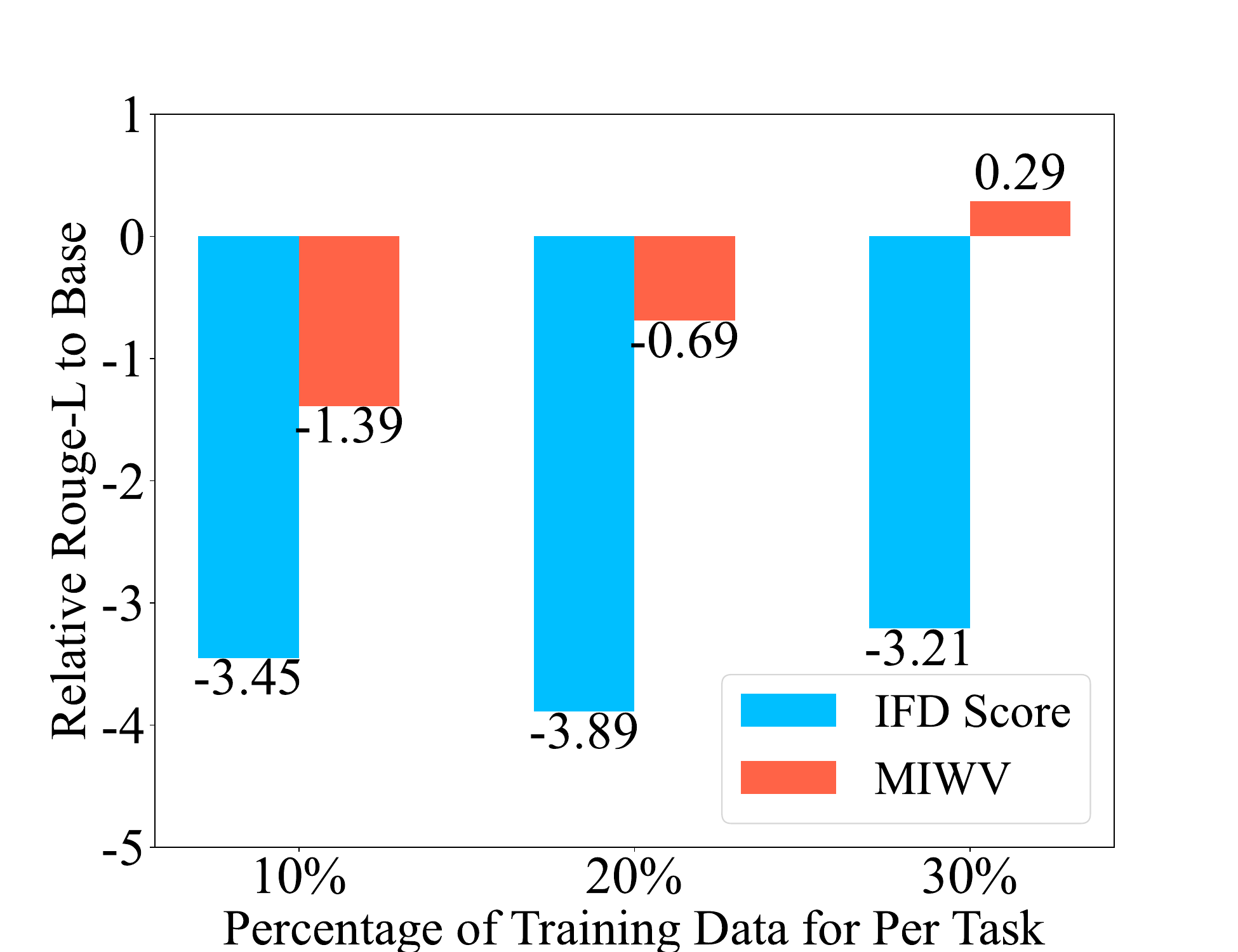}
		\label{fig:3c}
	}\ \ \
	\caption{Comparison between our models and the IFD Score models. (a), (b), (c) All are evaluated using Rouge-L.}
\label{fig:2_1}
\end{figure*}

In Figure~\ref{fig:2_1}, we present a performance comparison between models trained on the complete dataset as a baseline and models trained on the top 10\%, 20\%, and 30\% of high MIWV data from the multi-task dataset. These results are compared with the IFD scores reported in existing studies. Additionally, detailed experimental results are documented in Table~\ref{tab:4}. The results indicate that the model trained on the top 30\% of MIWV data exceeds the performance of the baseline model. Furthermore, when the number of training tasks reaches 200, the model trained on the top 20\% of MIWV data also outperforms the baseline model. It is worth noting that the "IFD Score" does not surpass the performance of the baseline model in any of the experiments conducted.

\clearpage

\begin{table*}[!ht]
\caption{ROUGE-L of our models and the IFD Score models on the NIV2 Dataset, bolded scores indicate performance above the base.}
\label{tab:4}
\centering
\begin{tabular}{l|c|cc|cc|cc}
\toprule
Percentage & 100\% & \multicolumn{2}{c|}{10\%} & \multicolumn{2}{c|}{20\%}       & \multicolumn{2}{c}{30\%} \\ \midrule
Tasks           & Base  & MIWV             & IFD    & MIWV           & IFD            & MIWV            & IFD    \\ \midrule
100             & 47.75 & 47.57   & 47.31  & 47.51          & 47.71 & \textbf{48.34}  & 47.29  \\
300             & 48.48 & 47.73   & 46.59  & 48.64 & 46.53          & \textbf{49.41}  & 47.09  \\
500             & 50.03 & 48.64   & 46.58  & 49.34 & 46.14          & \textbf{50.32}  & 46.82  \\ \bottomrule
\end{tabular}
\end{table*}

\clearpage

\section{Data Characteristics}
\label{sec:appendix E}
The six dimensions are defined as follows:

\noindent\textbf{Complexity:} The instruction addresses multiple layers and intricate aspects of the topic, requiring detailed understanding and sophisticated analysis to fully comprehend.

\noindent\textbf{Clarity:} The instruction is presented in a clear and straightforward manner, ensuring that information is easily understandable and unambiguous for the intended audience.

\noindent\textbf{Scope:} The instruction encompasses all relevant aspects and boundaries related to the topic, ensuring comprehensive coverage and inclusion of every essential elements.

\noindent\textbf{Depth:} The instruction thoroughly explores the subject matter, providing detailed explanations and nuanced insights to facilitate a complete and profound understanding.

\noindent\textbf{Creativity:} The instruction employs innovative and original approaches, encouraging novel thinking and enhancing engagement through unique presentation methods.

\noindent\textbf{Expertise:} The instruction reflects a high level of specialized knowledge and proficiency in the subject area, ensuring accuracy, reliability, and informed perspectives.

\clearpage

\section{Prompt for Evaluation}
As shown in Table~\ref{tab:5}, we provide the detailed prompt used to evaluate the performance of two responses to the same instruction.

\begin{table}[!h]
\caption{The Prompt for Evaluating Response Performance Using GPT-4.}
\centering
\label{tab:5}
\begin{tabular}{p{0.96\columnwidth}}
\toprule
\textbf{The Prompt for Evaluation}                                                                      \\ \midrule
\textbf{System Prompt}                                                                                                                                                                                                              \\
You are a helpful and precise assistant for checking the 
quality of the answer.                                                                              \\
         \\
\textbf{User Prompt}                                                                                                                                                                                                                \\
{[}Question{]}                                                                                                                                          \\
\textit{Question}                                                                                                                                             \\
{[}The Start of Assistant 1’s Answer{]}                                                                                                                                                                                            \\
\textit{Answer1}                                                                                                                                                                                                                     \\
{[}The End of Assistant 1’s Answer{]}                                                                                                                   \\
{[}The Start of Assistant 2’s Answer{]}                                                                                                                        \\
\textit{Answer2}                                                                                                                                                                                     \\
{[}The End of Assistant 2’s Answer{]}                                                                                                       
\\
\\
We would like to request your feedback on the performance of two AI assistants in response to the user question displayed above. \\  

Please rate the helpfulness, relevance, accuracy, and level of detail of their responses. Each assistant receives an overall score on a scale of 1 to 10, where a higher score indicates better overall performance. Please first output a single line containing only two values indicating the scores for Assistant 1 and 2, respectively. The two scores are separated by a space. In the subsequent line, please provide a comprehensive explanation of your evaluation, avoiding any potential bias and ensuring that the order in which the responses were presented does not affect your judgment.  \\
\bottomrule
\end{tabular}
\end{table}

\clearpage

\section{Sample Presentation}
Table~\ref{tab:6} and Table~\ref{tab:7} display examples from the Alpaca dataset selected by MIWV. Specifically, Table~\ref{tab:6} showcases the top 5\% of samples selected by MIWV, while Table~\ref{tab:7} presents the least 5\% of samples selected by MIWV.

\begin{table*}[]
\caption{Examples sampled from the Alpaca dataset with top 5\% of data selected by MIWV.}
\centering
\label{tab:6}
\begin{tabular}{p{0.96\textwidth}}
\toprule
\textbf{Top 5\% Examples}                                                                                                                                                             \\ \midrule
\textbf{[Instruction]}                                                                                                                                                                                                                                     \\
Design a survey to assess customer satisfaction with an online retail company.                                                                                                                                                                                                                                           \\
\textbf{[Input]}                                                                                                                                                                                                                                                                                                                \\
Company: Amazon.                                                                                                                                                                                                                                    \\
\textbf{[Response]}                                                                                                                                                               \\
Design a survey to assess customer satisfaction with Amazon:                                                                                                                                                                                      \\
The survey should include a combination of Likert-scale and open-ended questions in order to gauge customers' overall satisfaction with Amazon. Questions should address the website's user experience, selection of items, customer service, delivery speed, and return policy. Additionally, it should also include questions regarding any areas for improvement that customers would like to see. Furthermore, the survey should provide customers with an opportunity to explain why they feel satisfied or dissatisfied with the company. \\ \midrule
\textbf{[Instruction]}                                                                                                                                                                                                                                 \\
Classify the following musical genres:                                                                                                                                                                                                            \\
\textbf{[Input]}                                                                                                                                                                                                                                         \\
- Punk                                                                                                                                                                                                                                              \\
- Jazz                                                                                                                                                                       \\
- Heavy metal                                                                                                                                                              \\
\textbf{[Response]}                                                                                                                                                                                                                                     \\
- Punk: Genre of rock music that developed in the mid-1970s and typically lyrics with hard-edged melodies and singing styles, stripped-down instrumentation, and often political, anti-establishment lyrics.                                                                                                                                                                                   \\
- Jazz: A genre of music that originated in the late 19th and early 20th centuries, characterized by blue notes, improvisation, swing, call and response, polyrhythms, and syncopation.                                                                                                                                   \\
- Heavy metal: A genre of rock music that developed in the late 1960s and early 1970s, characterized by aggressive, driving rhythms and amplified, distorted guitars, bass, drums, and vocals.                                                                                                                               \\ \midrule
\textbf{[Instruction]}                                                                                                                                                                                                                              \\
Determine what type of conflict this story has.                                                                                                                             \\
\textbf{[Input]}                                                                                                                                                                \\
Damien and Tucker had been competing in a game of chess for hours. Both were determined to win, and the tension between the two friends was palpable.                                                                                              \\
\textbf{[Response]}                                                                                                                                                            \\
This story has a Man vs. Man conflict, which is the conflict between two or more people. In this case, the conflict is between Damien and Tucker as they both strive to win the chess game.                                           
\\ \midrule
\textbf{[Instruction]}                                                                                                                                           \\
Explain how tree pruning works.                                                                                                                          \\
\textbf{[Input]}                                                \\                                                                                                                                                                      None. \\                                             
\textbf{[Response]}                                                                                                                                \\
Tree pruning is the practice of selectively removing branches, buds, or roots to enhance or maintain the shape or structure of a tree. It is performed to reduce the risk of injury or damage caused by weak or hazardous branches, to enhance the quality of flowers and fruits, to improve the overall appearance of the tree, to reduce height, and to reduce shading from a neighbor’s property. It is important to ensure the tree is pruned correctly and that only healthy and viable branches are removed. Improper pruning can cause damage to the tree or even death.   
\\ \midrule
\textbf{[Instruction]}                                                                                                                                                         \\
Given an array of strings, remove any strings that contain the letter ``a''.                                                                                                                                                                                                                                       \\
\textbf{[Input]}                                                                                                                          \\
{[}``apple'', ``banana'', ``orange'', ``grape''{]}                                                                                                                                                                        \\
\textbf{[Response]}                                                                                                                \\
{[}``banana'', ``orange'', ``grape''{]} \\
\bottomrule
\end{tabular}
\end{table*}

\begin{table*}[]
\caption{Examples sampled from the Alpaca dataset with least 5\% of data selected by MIWV.}
\centering
\label{tab:7}
\begin{tabular}{l}
\toprule
\textbf{Least 5\% Examples}                                                                                   \\ \midrule
\textbf{[Instruction]}                                                                                    \\
Find the frequency of the word ``the'' in a given text.                                                 \\
\textbf{[Input]}                                                                                          \\
The quick brown fox jumps over the lazy dog.                                                          \\
\textbf{[Response]}                                                                                       \\
2.                                                                                                    \\ \midrule
\textbf{[Instruction]}                                                                                    \\
Identify if the sentence contains a simile.                                                          \\
\textbf{[Input]}                                                                                          \\
It was like a maze of winding streets.                                                               \\
\textbf{[Response]}                                                                                       \\
Yes.                                                                                                  \\ \midrule
\textbf{[Instruction]}                                                                                    \\
Identify whether the following phrase is a true or false statement ``The universe is ever expanding.'' \\
\textbf{[Input]}                                                                                          \\
None.   \\

\textbf{[Response]}                                                                                       \\
True.                                                                                                \\ \midrule
\textbf{[Instruction]}                                                                                    \\
Classify this statement as a fact or opinion                                                         \\
\textbf{[Input]}                                                                                          \\
Germany is the largest country in Europe                                                             \\
\textbf{[Response]}                                                                                       \\
Fact.           
   \\ \midrule
\textbf{[Instruction]}                                                                                    \\
Select one of the following items. Output the selected item.                                                         \\
\textbf{[Input]}                                                                                          \\
Items: Football, Basketball, Volleyball.                                                             \\
\textbf{[Response]}                                                                                       \\
Basketball.
\\ \bottomrule
\end{tabular}
\end{table*}

\end{document}